\documentclass{article}

\usepackage{arxiv}

\usepackage[utf8]{inputenc} 
\usepackage[T1]{fontenc}    
\usepackage{url}            
\usepackage{booktabs}       

\usepackage{amsmath,amssymb,amsfonts,latexsym,amsthm,bm,breqn,dsfont,gensymb}
\usepackage{latexsym,breqn}
\usepackage{nicefrac}       
\usepackage{microtype}      
\usepackage{natbib}

\usepackage{graphicx,subfigure}

\usepackage{algorithm}
\usepackage[noend]{algpseudocode}
\makeatletter
\def\BState{\State\hskip-\ALG@thistlm}

\usepackage{ctable,multicol,multirow}

\usepackage{color}
\definecolor{green}{rgb}{0, 0.5, 0}
\usepackage[pdftex,
			bookmarks, bookmarksnumbered=true, bookmarksopen=true,
			colorlinks, citecolor=blue, filecolor=black, linkcolor=black, urlcolor=green,
			pdfauthor={López-Lopera, Andrés Felipe}, pdftitle={PhDEMSE},
			pdftoolbar=true, pdfmenubar=true, pdffitwindow = true
			]{hyperref}

\newcommand{\partialdiff}[2]{\frac{\partial #1}{\partial #2}}
\newcommand{\cov}[1]{\operatorname{cov} \left\{ #1 \right\}}

\newcommand{\erf}[1]{\operatorname{erf} \left\{ #1 \right\}}
\newcommand{\erfc}[1]{\operatorname{erfc} \left\{ #1 \right\}}
\newcommand{\realset}[1]{\mathds{R}^{#1}}
\newcommand{\normrnd}[2]{\mathcal{N}\left({#1,#2}\right)}

\newcommand{\Bu}{\textbf{u}}
\newcommand{\Bzero}{\boldsymbol{0}}
\newcommand{\By}{\textbf{y}}
\newcommand{\Btheta}{\boldsymbol{\theta}}
\newcommand{\BK}{\textbf{K}}

\newcommand*\rotv{\rotatebox{90}}

\title{Physically-Inspired Gaussian Process Models for Post-Transcriptional Regulation in Drosophila}
\date{}
\author{
  Andr\'es F. L\'opez-Lopera\thanks{Part of this work was completed during an internship of A. F. L\'opez-Lopera at PROWLER.io.} \\
  Mines Saint-\'Etienne\\
  42000 Saint-\'Etienne, France \\
  \texttt{andres-felipe.lopez@emse.fr} \\
  \And
  Nicolas Durrande\\
  PROWLER.io \\
  Cambridge, CB2 1LA, UK \\
  \texttt{nicolas@prowler.io} \\
  \And
  Mauricio A. \'Alvarez \\
  University of Sheffield \\
  Sheffield, S1 4DP, UK \\
  \texttt{mauricio.alvarez@sheffield.ac.uk} \\  
}

\begin{document}
\maketitle

\begin{abstract}
			The regulatory process of Drosophila is thoroughly studied for understanding a great variety of biological principles. While pattern-forming gene networks are analysed in the transcription step, post-transcriptional events (e.g. translation, protein processing) play an important role in establishing protein expression patterns and levels. Since the post-transcriptional regulation of Drosophila depends on spatiotemporal interactions between mRNAs and gap proteins, proper physically-inspired stochastic models are required to study the link between both quantities. Previous research attempts have shown that using Gaussian processes (GPs) and differential equations lead to promising predictions when analysing regulatory networks. Here we aim at further investigating two types of physically-inspired GP models based on a reaction-diffusion equation where the main difference lies in where the prior is placed. While one of them has been studied previously using protein data only, the other is novel and yields a simple approach requiring only the differentiation of kernel functions. In contrast to other stochastic frameworks, discretising the spatial space is not required here. Both GP models are tested under different conditions depending on the availability of gap gene mRNA expression data. Finally, their performances are assessed on a high-resolution dataset describing the blastoderm stage of the early embryo of Drosophila melanogaster
\end{abstract}

		
		\section{Introduction}
		\label{sec:introduction}
		Regulatory process modelling in molecular mechanisms has taken great attention due to its significant role in systems biology \citep{Alon2006SystemsBiology,Barrera2016SurveyTFs}. The gene regulation consists mainly of two steps. First, DNA sequences are encoded in mRNA molecules according to the need of proteins' production in cells (transcription step). Then, the mRNA is used as a template for protein synthesis (translation step) \citep{Alon2006SystemsBiology,Forgacs2005BiologyEmbryo}. Both transcription and translation steps are crucial to understand a great variety of biological phenomena such as the molecular mechanisms of cell survival and protein production  \citep{Alon2006SystemsBiology,Barrera2016SurveyTFs,Dilao2010mRNADros}. While pattern-forming gene networks can be analysed in the former step, post-transcriptional events (e.g. RNA splicing, translation, protein processing) play an important role in establishing protein expression patterns and levels \citep{Alon2006SystemsBiology,Becker2013DrosMel,Dilao2010mRNADros}. This paper is focused on the post-transcriptional regulation at the mRNA level, more precisely, on modelling the gap gene network dynamics of Drosophila.
		
		The genus Drosophila is one of the most studied examples of regulatory processes \citep{Becker2013DrosMel,Rogers2014SurveyDrosophila}. The regulation in Drosophila is commonly analysed to understand how protein concentrations are expressed and their influence on biological systems \citep{Rogers2014SurveyDrosophila,Hsiao2016GRN,Dilao2010mRNADros}. It is also used to investigate the interaction between the environment and biological processes \citep{Rogers2014SurveyDrosophila}. For example, in recent work, it has been discovered that there are specific molecular mechanisms that control the circadian rhythm in living species (research awarded with the 2017 Nobel prize in Physiology and Medicine) \citep{Yagita2018LetterNobel}. Therefore, because of the versatility of Drosophila, its regulatory network is commonly used as a standard model.
		
		Due to the importance of post-transcriptional regulation in Drosophila, accurate approaches for capturing the existing link between mRNAs and proteins are required \citep{Becker2013DrosMel,Dilao2010mRNADros,Wilson2010GqpGenes}. In the last decades, several frameworks have been proposed encouraging the modelling of regulatory processes as stochastic processes \citep{Barenco2006p53,Becker2013DrosMel,Lipniacki2006StochGeneExpression,Dalessi2012ModelingMorphogen,Erban2007PracticalGuide}. More precisely, they have suggested the use of Gaussian processes (GPs) and differential equations \citep{Lawrence2006modelling,Gao2008GP,Alvarez2011LLFM,Vasquez2014LFMDrosMel}. For the case of post-transcriptional regulation, GP-based approaches assume that both mRNA and protein concentrations are Gaussian-distributed \citep{Liu2012GPBicoid,Alvarez2011LLFM}. Then, due to the linearity of the differential equation used to link those quantities, mechanistic parameters can be encoded as parameters of GP covariance functions \citep{Lawrence2006modelling,Alvarez2011LLFM}. According to experiments on both synthetic and real-world data in \citep{Lawrence2006modelling,Alvarez2011LLFM,Vasquez2014LFMDrosMel,Gao2008GP}, physically-inspired GPs have yielded competitive and promising results for modelling regulatory processes.
		
		Different classes of physically-inspired approaches have been proposed to model the regulation of the early embryo of Drosophila melanogaster \citep{Dalessi2012ModelingMorphogen,Liu2012GPBicoid,Alvarez2011LLFM,Vasquez2014LFMDrosMel}. In \citep{Dalessi2012ModelingMorphogen}, a Green's function method is introduced to model the Bicoid gradient establishment in Drosophila's embryos assuming the mRNA to be deterministic. Their approach presents two main drawbacks. First, it provides only deterministic solutions which do not allow expressing uncertainties on mRNA distributions. Second, their model is tested only on synthetic data due to issues on experimental settings. In contrast to \citep{Dalessi2012ModelingMorphogen}, a stochastic framework based on GPs is proposed in \citep{Liu2012GPBicoid}, allowing the expression of uncertainty on inferred Bicoid concentrations. Although their results are promising, the discretisation of the spatial space is required; losing quality of resolution in predictions. More recently, in \citep{Alvarez2011LLFM,Vasquez2014LFMDrosMel}, an alternative physically-inspired GP framework using Green's functions is proposed to model continuous mRNA concentrations. There, it is assumed that the mRNA acting in the regulatory network was unknown and had to be estimated. Their assumption stands due to the common lack of (commonly expensive) mRNA data and the availability of protein concentration data. They place a GP prior over the mRNA and built up the resulting GP over the protein. Their framework requires explicitly solving the associated differential equation, followed by solving multiple integrals involving kernel functions which is not always feasible \citep{Alvarez2011LLFM,Guarnizo2018KernelApprox}. Assuming that closed-form solutions are available, then the resulting GP over the protein can be established, and the mRNA concentration can be inferred conditionally to the protein concentration data. The model proposed in \citep{Alvarez2011LLFM,Vasquez2014LFMDrosMel} still presents some limitations. First, due to the lack of mRNA data, their approach could not be thoroughly tested for the inference of mRNA patterns. Second, in order to obtain closed-form expressions, their work is limited to a class of kernels.
		
		An alternative approach is to assume the GP prior over the protein rather than over the mRNA, building up the resulting GP framework. This leads to a GP model where the solution of the differential equation is not required, but the differentiation of kernel functions is. For further discussions, we refer to GP-mRNA and GP-Protein to the physically-inspired GP with prior over the mRNA or protein concentrations, respectively. Therefore, our main contributions are threefold. First, we introduce the GP-Protein model as a novel alternative that does not require solving differential equations. Second, we further investigate both GP-mRNA and GP-Protein models, assessing their performances when data from both mRNA and protein concentrations are available. Three situations are analysed depending on the data availability: whether from the mRNA, from the protein or from both quantities. Third, we test both physically-inspired GP models on a high-resolution dataset describing the blastoderm stage of the early embryo of Drosophila \citep{Becker2013DrosMel}.
		
		This paper is organized as follows. In Section \ref{sec:drosophila}, we briefly describe the gap gene network associated with segmentation in early Drosophila organism development. In Section \ref{sec:hGP}, we establish both physically-inspired GP models based on a diffusion equation. For the GP-mRNA model, we refer to the formulation from \citep{Alvarez2011LLFM}, but we write the main equations here for readability and further discussion. For the GP-Protein model, its formulation is completely detailed in Appendix \ref{app:SISOfGPGene}. We also assess both GP models on synthetic examples under different conditions depending on the data availability. In Section \ref{sec:results}, we test the models using the Drosophila's database from \citep{Becker2013DrosMel}. In Section \ref{sec:conclusion}, we summarise the conclusions, as well as potential future work.

		\section{Gap gene network of Drosophila}
		\label{sec:drosophila}
		This work focuses on the role of post-transcriptional regulation within the context of gap gene networks associated with segmentation during the blastoderm stage of early Drosophila development. There, a set of molecules known as morphogens are responsible for embryo segmentation \citep{Jaeger2012DrosBlastodermPattern,Dalessi2012ModelingMorphogen,Vasquez2014LFMDrosMel}. Morphogens propagate spatially and establish maternal gradients along the anterior-posterior (A-P) axis of the embryo, describing a reaction-diffusion process \citep{Jaeger2012DrosBlastodermPattern,Meinhardt2015EmbryonicBody}. Then, maternal gradients interact with specific trunk gap genes (e.g. \textit{hunchback}--$hb$, \textit{caudal}--$cd$, \textit{Kr\"uppel}--$kr$, \textit{knirps}--$kni$ and \textit{giant}--$gt$), and this gap gene network of interactions constitutes the segmentation of the Drosophila \citep{Becker2013DrosMel,Surkova2013QuantitativeDynamics,Alvarez2011LLFM}. 
		
		The reaction-diffusion process during early Drosophila embryo development is usually modelled through linear partial differential equations (PDEs) \citep{Becker2013DrosMel,Alvarez2011LLFM,Vasquez2014LFMDrosMel}. For readability, and according to the structure of the dataset used in Section \ref{subsec:dataset}, we focus on the gap gene network dynamics,
		\begin{align}
		\partialdiff{y(x,t)}{t} = S  u(x,t) - \lambda y(x,t) + D \partialdiff{^2 y (x,t)}{x^2},
		\label{eq:PDE}
		\end{align} 
		where the relative gap protein concentration $y(x,t)$, at location $x$ and instant $t$, is driven by the mRNA $u(x,t)$. Here, the translation rate constant $S$ represents the rate of protein production from the mRNA, and parameters $\lambda$ and $D$ are the decay and diffusion rate constants.
		
		\section{Physically-inspired Gaussian processes for post-transcriptional regulation}
		\label{sec:hGP}
		Physically-inspired Gaussian process (GP) models are stochastic approaches where linear differential equations are encoded into kernel functions \citep{Lawrence2006modelling,Alvarez2011LLFM}. From the data-driven point of view, they can be established without specifying all the physical interactions involved in mechanistic processes. From the physically-inspired models' point of view, they provide accurate predictions even in regions where data are not available. Since they can account for different types of differential equations, physically-inspired GPs have been applied successfully in several fields such as human motion capture and robotics \citep{Alvarez2011LLFM,Agudelo2017LFMHMM,Guarnizo2018KernelApprox}, neuroscience \citep{Alvarado2014LFMWaveEq}, and molecular biology and genetics \citep{Lawrence2006modelling,Alvarez2011LLFM,Vasquez2014LFMDrosMel,LopezLopera2017SDSIM,Croix2018ArXiv,Gao2008GP}. In \citep{Alvarez2011LLFM,Vasquez2014LFMDrosMel,Croix2018ArXiv}, they have been applied to model the early embryo development of Drosophila melanogaster.
		
		Using the reaction-diffusion model in \eqref{eq:PDE} as mechanistic model, we can then assume a zero-mean GP prior with covariance function $k$ either over $u$ or $y$. Since \eqref{eq:PDE} involves only linear operations, the Gaussianity holds for both sides no matter where the GP prior is placed. Let $\Bu$ and $\By$ be Gaussian vectors containing evaluations of the GPs $u$ and $y$ (respectively) at a given set of couples $(x_i,t_i)_{i=1}^{N}$. Then, the joint process, at $(x_i,t_i)_{i=1}^{N}$, follows a multivariate Gaussian distribution,
		\begin{align}
		\begin{bmatrix} \Bu \\ \By \end{bmatrix} \sim \normrnd{\begin{bmatrix} \Bzero \\ \Bzero \end{bmatrix}}{\begin{bmatrix} \BK_{\Bu,\Bu} & \BK_{\By,\Bu}^{\top}  \\  \BK_{\By,\Bu} & \BK_{\By,\By} \end{bmatrix}},
		\label{eq:jointGP}
		\end{align}
		with covariance matrices $(\BK_{\Bu,\Bu})_{i,j} = k_{u,u}(x_i,t_i,x_j,t_j)$, $(\BK_{\By,\Bu})_{i,j} = k_{y,u}(x_i,t_i,x_j,t_j)$, $(\BK_{\By,\By})_{i,j} = k_{y,y}(x_i,t_i,x_j,t_j)$ for $i,j = 1, \cdots, N$, with kernel structure $k_{z,z'}$ given by $k_{z,z'}(x_i,t_i,x_j,t_j) = \cov{z(x_i,t_i), z'(x_j,t_j)}$. One can note that the physically-inspired GP model is fully established if we encode the mechanistic model from \eqref{eq:PDE} into the covariance matrix of the joint process in \eqref{eq:jointGP}. 
		
		Next, we study two choices of physically-inspired GP models depending on whether the GP prior is placed over the mRNA (GP-mRNA) or the gap protein (GP-Protein).
		
		\subsection{GP-mRNA model}
		\label{subsec:hGPmRNA}
		Since mRNA data are not always available, GP prior assumptions are commonly placed over mRNA profiles \citep{Alvarez2011LLFM,Vasquez2014LFMDrosMel}. This approach requires writing the output process $y$ in terms of the driving-force $u$, with the explicit solution of the PDE in \eqref{eq:PDE}. Notice that the complexity of this solution depends on the initial and boundary conditions \citep{Polyanin2001handbook,Stakgold2011GreenFunctions,Abramowitz1965Handbook}. Here we assume homogeneous conditions, i.e. $y(x,t=0) = 0$ and $y (x=0,t) = y (x=l,t) = 0$ for a diffusion evolution in $x \in [0, l]$ with $l \in \realset{+}$. These assumptions are made according to the structure of the dataset used in Section \ref{sec:results}. Hence, the solution of \eqref{eq:PDE} is given by  \citep{Polyanin2001handbook,Stakgold2011GreenFunctions}
		\begin{dmath}
			y (x,t) = S \int_{0}^{t} \int_{0}^{l} u(\xi,\tau) G(x,\xi, t-\tau) d\xi d\tau,
			\label{eq:hGPmRNAEq}
		\end{dmath}
		where the Green's function $G(x,\xi, t)$ is defined as
		\begin{displaymath}
		G(x,\xi, t) = 
		c(t) \sum_{n=1}^{\infty} \sin(\omega_n x) \sin(\omega_n \xi) \exp\{-D \omega_n^2\},
		\end{displaymath}
		with $c(t) = \frac{2}{l} \exp\{-\lambda t\}$ and $\omega_n = \frac{n \pi}{l}$. In practice, the Green's function is commonly truncated, and the accuracy of the solution in \eqref{eq:hGPmRNAEq} depends on the number of terms used in the approximation. 
		
		Then, a GP prior can be placed over the mRNA $u$. In order to obtain analytical expressions in further steps, we use a zero-mean GP prior with covariance function $k_{u,u}$ given by the product of two squared exponential (SE) kernel functions, i.e.
		\begin{align}
		k_{u,u}(x,t,x',t')
		= \sigma^2 k(x,x') k(t,t') = \sigma^2 \exp \bigg\{-\frac{(x-x')^2}{\theta_x^2} \bigg\} \exp \bigg\{- \frac{(t-t')^2}{\theta_t^2} \bigg\},
		\label{eq:hGPmRNAKuu} 
		\end{align}
		where $\theta_x$ and $\theta_t$ are the length-scale parameters.
		
		Now, we aim at computing the covariance function for the output $k_{y,y}$, and the cross-covariance function between the output and the driving-force $k_{y,u}$.
		
		\subsubsection{Covariance function for the output}
		Since the PDE in \eqref{eq:PDE} is linear, the output process $y$ is also a GP with covariance function $k_{y,y}(x,t,x',t') = \cov{y(x,t),y(x',t')}$ given by
		\begin{align*}
		k_{y,y}(x,t,x',t')
		= \sigma^2 S^2 \int_{0}^{t} \int_{0}^{t'} \int_{0}^{l} \int_{0}^{l}  \widehat{G}(x,\xi,t,\tau,x',\xi',t',\tau') 
		\times  k(\xi,\xi')k(\tau,\tau') d\xi' d\xi d\tau' d\tau,
		\end{align*}
		with $\widehat{G}(x,\xi,t,\tau,x',\xi',t',\tau') = G(x,\xi, t-\tau) G(x',\xi', t'-\tau')$.
		
		After solving the multiple integrals, one can show that the covariance function $k_{y,y}$ is given by \citep{Alvarez2011LLFM}
		\begin{align}
		k_{y,y}(x,t,x',t')
		= \frac{4 \sigma^2 S^2}{l^2}  \sum_{\forall n} \sum_{\forall m} \sin(\omega_n x) \sin(\omega_m x') K (t,t',n,m) C (n,m), \label{eq:hGPmRNAKyy}
		\end{align}
		where
		\begin{align*}
		K (t,t',n,m) = \frac{\theta_t \sqrt{\pi}}{2} [ h(\beta_m,t',t) + h(\beta_n,t,t') ],
		\end{align*}
		and
		\begin{dmath*}
			h(\beta_m,t',t)
			= \frac{e^{ \left(\frac{\beta_m \theta_t }{2}\right)^2} }{\beta_m+\beta_n} \left[ e^{-\beta_m(t'-t)} \mathcal{H}  (\beta_m,t,t') - e^{-(\beta_m t' + \beta_n t)} \mathcal{H}  (\beta_m,0,t') \right],
		\end{dmath*}
		with $\beta_n = \lambda + D \omega_n^2$, $\beta_m = \lambda + D \omega_m^2$, and $\mathcal{H}(\zeta,v,\varphi) = \erf{\frac{\varphi-v}{\theta} - \frac{\theta \zeta}{2}} + \erf{\frac{v}{\theta} + \frac{\theta \zeta}{2}}$. The operator $\operatorname{erf}$ denotes the error function \citep{Polyanin2001handbook}.
		
		The term $C(n,m)$ is defined according to \citep{Alvarez2011LLFM}. When $n \ne m$ such that $m$ and $n$ are both even or both odd numbers, $C(n,m)$ follows
		\begin{displaymath}
		C(n,m)
		= \frac{\theta_x l}{\sqrt{\pi}(m^2-n^2)} \left\{n \mathcal{I}\left[ \mathcal{W}_{\theta_x}(m) \right] - m \mathcal{I}\left[\mathcal{W}_{\theta_x}(n) \right] \right\},
		\end{displaymath}
		where $\mathcal{I}$ is the operator returning the imaginary part of the argument, and
		\begin{displaymath}
		\mathcal{W}_{\theta_x}(m) = w(jz_1^{\gamma_m})-e^{-\left(\frac{l}{\theta_x}\right)^2}e^{-\gamma_ml}w(jz_2^{\gamma_m}),
		\end{displaymath}
		with $\gamma_n = j \omega_n$, $\gamma_m = j \omega_m$, $z_1^{\gamma_n}=\frac{\theta_x\gamma_n}{2}$ and $z_2^{\gamma_n}=\frac{l}{\theta_x}+\frac{\theta_x\gamma_n}{2}$. We note that $w(z) = \exp\{-z^2\} \erfc{-jz}$ is known as the Faddeeva function \citep{Poppe1990Faddeva}, with operator $\operatorname{erfc}$ denoting the complementary error function. Otherwise, if $n \ne m$ but $m$ and $n$ are not both even or both odd numbers, then $C(n,m)=0$. If $n = m$, then
		\begin{dmath*}
			C(n,n) = 
			\frac{\theta_x\sqrt{\pi}\,l}{2}\left\{\mathcal{R}\left[ \mathcal{W}_{\theta_x}(n) \right] -\mathcal{I}\left[ \mathcal{W}_{\theta_x}(n) \right]\left[\frac{\theta_x^2n\pi}{2l^2}+\frac{1}{n\pi}\right]\right\} +\frac{\theta^2_x}{2}\left[e^{-(\frac{l}{\theta_x})^2}\cos(n\pi)-1\right],
		\end{dmath*}
		where $\mathcal{R}$ is the operator returning the real part of the argument. 
		
		\subsubsection{Covariance function between the output and the driving-force}
		The cross-covariance function $k_{y,u}(x,t,x',t') = \cov{y(x,t),u(x',t')}$ between the output $y$ and the driving-force $u$ is given by
		\begin{align*}
		k_{y,u}(x,t,x',t')
		= \sigma^2 S \int_{0}^{t} \int_{0}^{l}  G(x,\xi, t-\tau) k(\xi,x')k(\tau,t') d\xi d\tau.
		\end{align*}
		After solving the double integral, one can show that \citep{Alvarez2011LLFM}
		\begin{align}
		k_{y,u}(x,t,x',t') = \frac{2\sigma^2 S}{l} \sum_{\forall n}\sin\left(\omega_n x\right)\widetilde{K}(t,t',n)\widetilde{C}(x',n),
		\label{eq:hGPmRNAKyu}
		\end{align}
		where
		\begin{displaymath}
		\widetilde{K}(t,t',n)=\frac{\theta_t\sqrt{\pi}}{2} e^{\left(\frac{\beta_n\theta_t}{2}\right)^2 } e^{ -\beta_n(t-t')} \mathcal{H}(\beta_n,t',t),
		\end{displaymath}
		\begin{displaymath}	
		\widetilde{C}(x',n)=\frac{\theta_x\sqrt{\pi}}{2}\mathcal{I}\left[\widetilde{\mathcal{W}}_{\theta_x}(x',n)\right],
		\end{displaymath}
		\begin{displaymath}
		\widetilde{\mathcal{W}}_{\theta_x}(x',n)
		= e^{-\left(\frac{x'-l}{\theta_x}\right)^2}e^{\gamma_nl} w(jz_2^{\gamma_n,x'})-e^{-\left(\frac{x'}{\theta_x}\right)^2}w(jz_1^{\gamma_n,x'}),
		\end{displaymath}
		with $z_1^{\gamma_n,x'}=\frac{x'}{\theta_x}+\frac{\theta_x \gamma_n}{2}$ and $z_2^{\gamma_n,x'}=\frac{x'-l}{\theta_x}
		+\frac{\theta_x\gamma_n}{2}$. Finally, the process in \eqref{eq:jointGP} can be computed using \eqref{eq:hGPmRNAKuu}, \eqref{eq:hGPmRNAKyy} and \eqref{eq:hGPmRNAKyu}. 
		
		One must note that the stability of the GP-mRNA model, besides depending on the number of terms of the Green's function, it also depends on the computation of the $\operatorname{erf}$ and Faddeeva functions (see expressions \eqref{eq:hGPmRNAKyu} and \eqref{eq:hGPmRNAKyy}). Since those functions do not have closed-form expressions, they have to be computed numerically \citep[see, e.g.,][]{Poppe1990Faddeva,Weideman1994ComplexErf}.
		
		\subsubsection{Toy example: inference of simulated data}
		\label{subsubsec:hGPmRNAtoy}
\begin{figure}
			\centering
			\subfigure[mRNA]{\includegraphics[width=0.325\columnwidth]{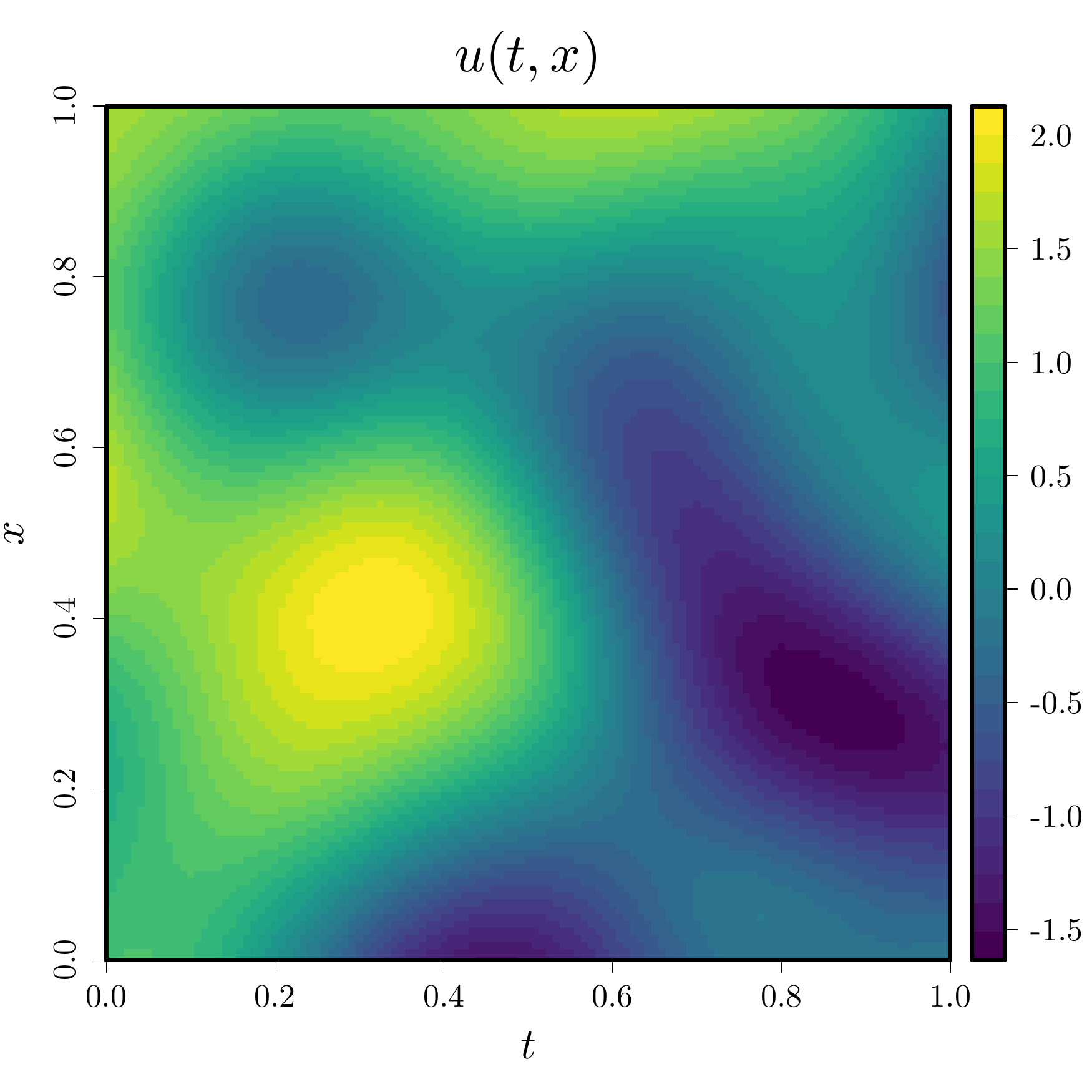}}
			\subfigure[Protein]{\includegraphics[width=0.325\columnwidth]{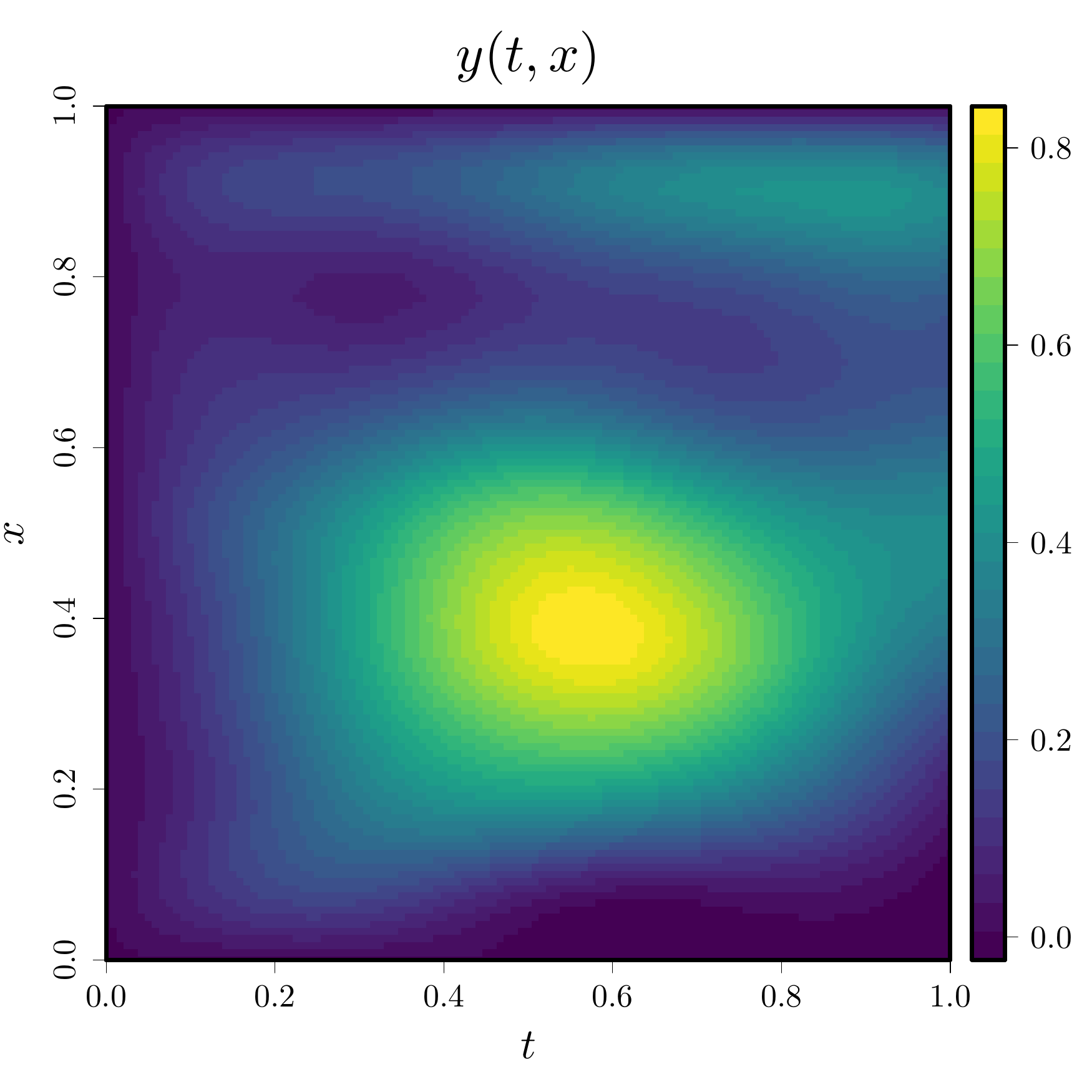}}
			\caption{Synthetic example generated by the GP-mRNA model.}
			\label{fig:hGPmRNAsample}
		\end{figure}

		To illustrate the properties of the GP-mRNA model, we generate synthetic data by sampling from the joint GP using the kernels functions \eqref{eq:hGPmRNAKuu}, \eqref{eq:hGPmRNAKyy} and \eqref{eq:hGPmRNAKyu}. We consider the domain $(x,t) \in [0,1]^2$, and we use ten components of the Green's function. We fix as covariance parameters $\sigma^2 = 1$, $\theta_x = \theta_t = 0.3$, and as mechanistic parameters $S = 1$, $\lambda = 0.1$, and $D = 0.01$. Samples are generated using a $41 \times 41$ equispaced grid on $[0,1]^2$. Figure \ref{fig:hGPmRNAsample} shows the generated mRNA and protein profiles. One can observe that the homogeneous conditions are ensured in the protein profile. Also notice that the GP-mRNA model does not guarantee that the mRNA and gap proteins are strictly positive quantities. As for standard GP models, GP-mRNA cannot account for positivity but non-linear transformations of GPs can be applied for ensuring it everywhere \citep[e.g. exponential of GPs ][]{Vanhatalo2007SparseLogGPs}. However, those transformations do not yield analytical solutions of the resulting joint GP as we provided in Section \ref{subsec:hGPmRNA}. Therefore, the synthetic example proposed here is for illustrative purposes only.
		
		We aim at testing the performance of GP-mRNA under three conditions. In the first two cases, we establish a joint GP using data only from the mRNA or the protein, and we then estimate both quantities in the whole domain $[0,1]^2$. We repeat the same procedure using conditioning data from both the mRNA and protein rather than only from one of them. We use the $Q^2 = 1 - \operatorname{SMSE}$ criterion, where SMSE is the standardised mean squared error \citep{Rasmussen2005GP}, to evaluate the quality of predictions over the points that were not used for training the GP model (test data). For noise-free observations, the $Q^2$ criterion is equal to one if the predictive mean of the resulting process is equal to the test data and lower than one otherwise. To evaluate the quality of the predictive variances, we use a criterion based on the coverage accuracy (CA) of the confidence intervals. For one standard deviation intervals, predictive variances should cover around $68\%$ of the test points \citep{Meyer1970IntroProbaStats}. Departure from $\operatorname{CA}_{\pm \sigma} = 0.68$ may indicate that the predictive variances are either underestimated (i.e. $\operatorname{CA}_{\pm \sigma} < 0.68$) or overestimated (i.e. $\operatorname{CA}_{\pm \sigma} > 0.68$). For a further discussion on assessing the quality of predictions using GP-mRNA, we assume that both covariance and mechanistic parameters are known and are equal to the ones used to generate the synthetic data.
		\begin{figure}
			\centering
			\subfigure[$Q^2 = 0.674$, $\operatorname{CA}_{\pm\sigma} = 0.527$]{\includegraphics[width=0.325\columnwidth]{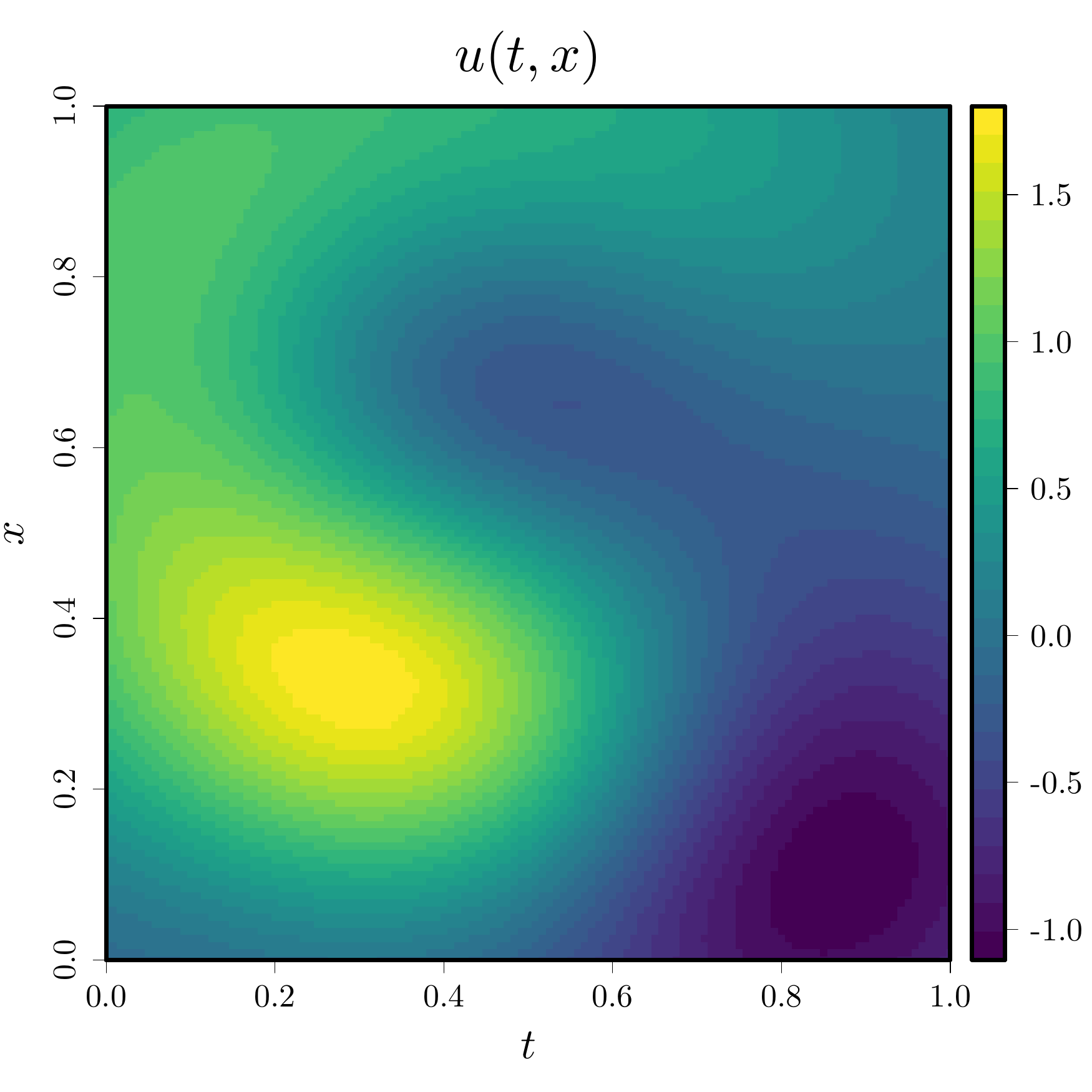}}
			\subfigure[$Q^2 = 0.520$, $\operatorname{CA}_{\pm\sigma} = 0.536$]{\includegraphics[width=0.325\columnwidth]{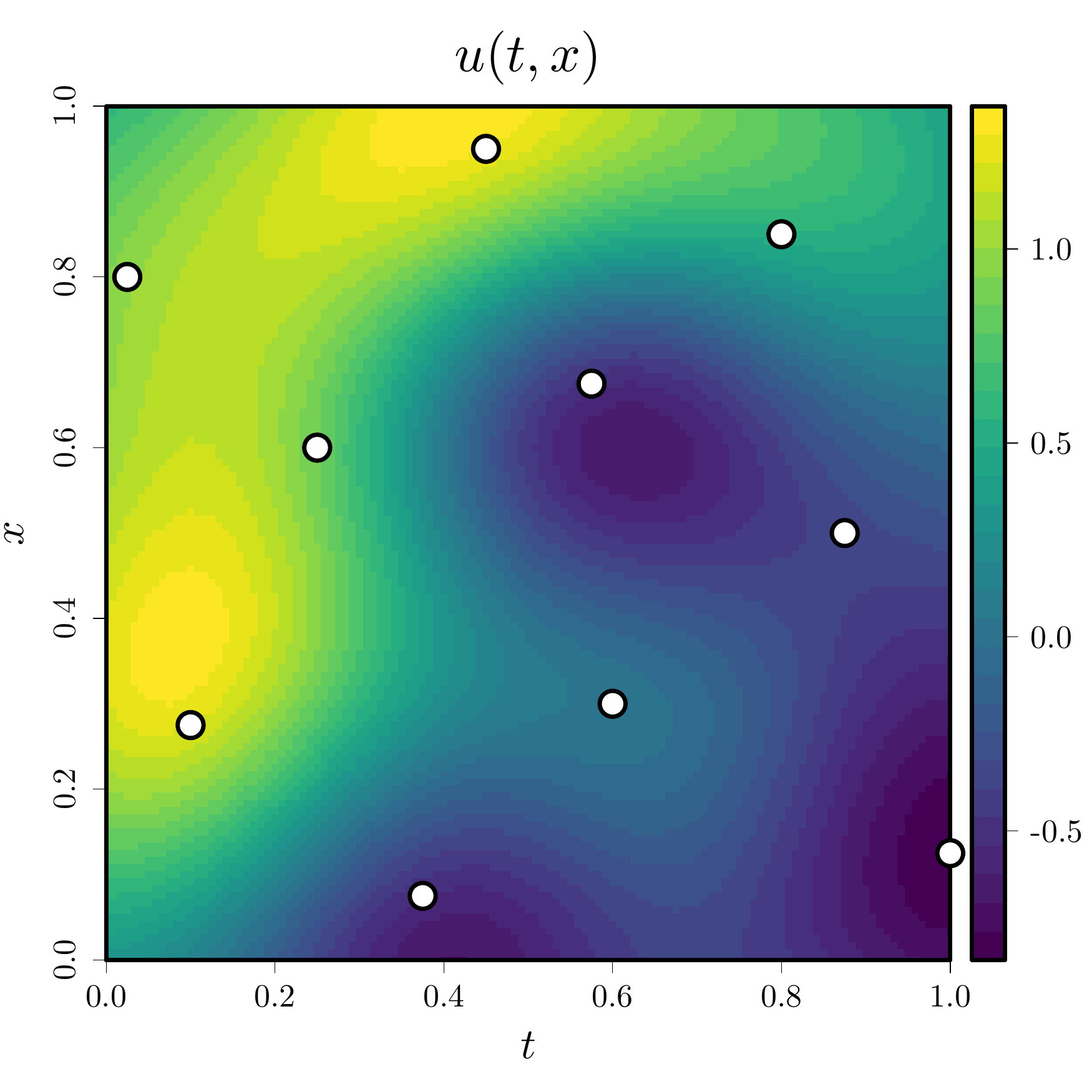}}
			\subfigure[$Q^2 = 0.872$, $\operatorname{CA}_{\pm\sigma} = 0.521$]{\includegraphics[width=0.325\columnwidth]{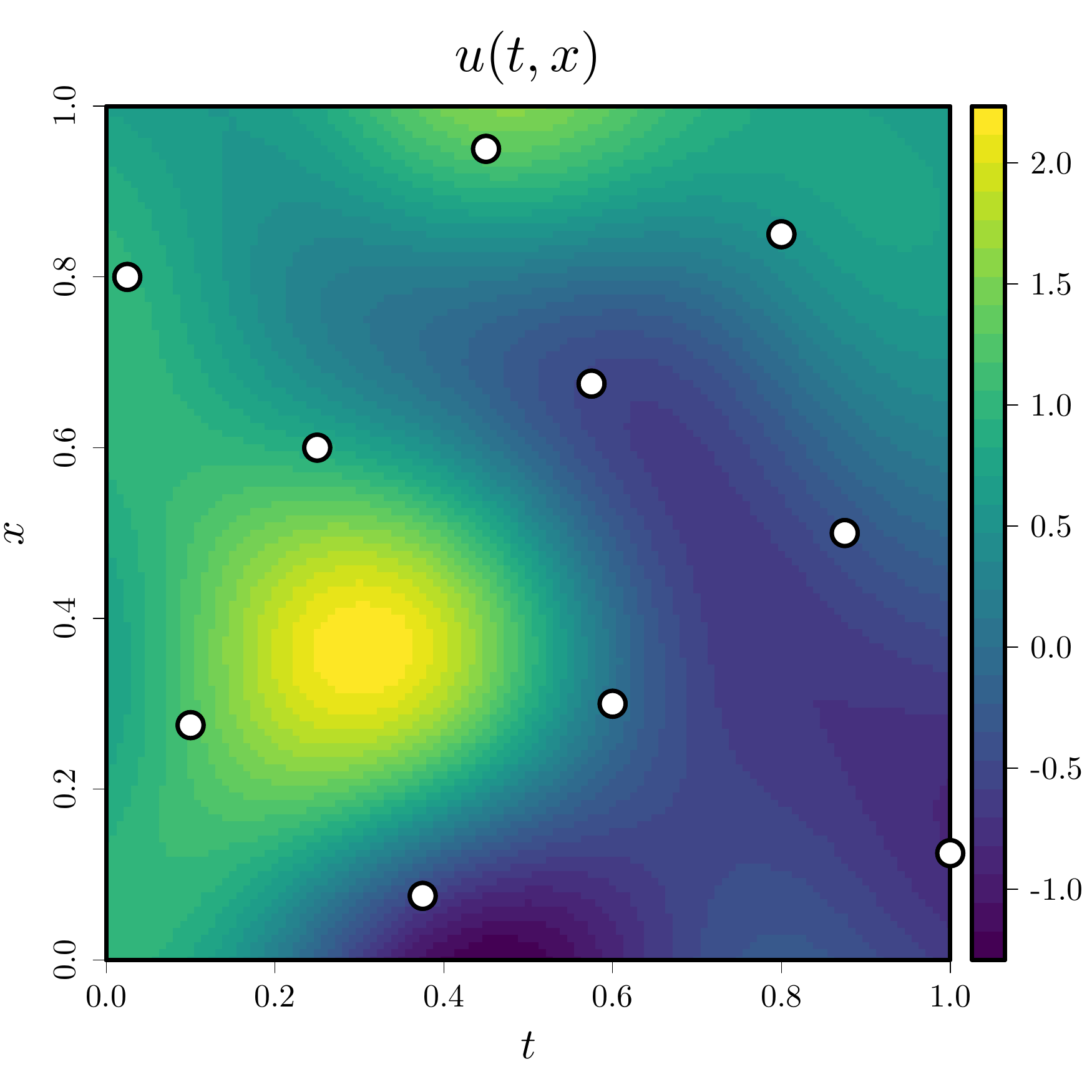}}			
			
			\subfigure[$Q^2 = 0.842$, $\operatorname{CA}_{\pm\sigma} = 0.416$]{\includegraphics[width=0.325\columnwidth]{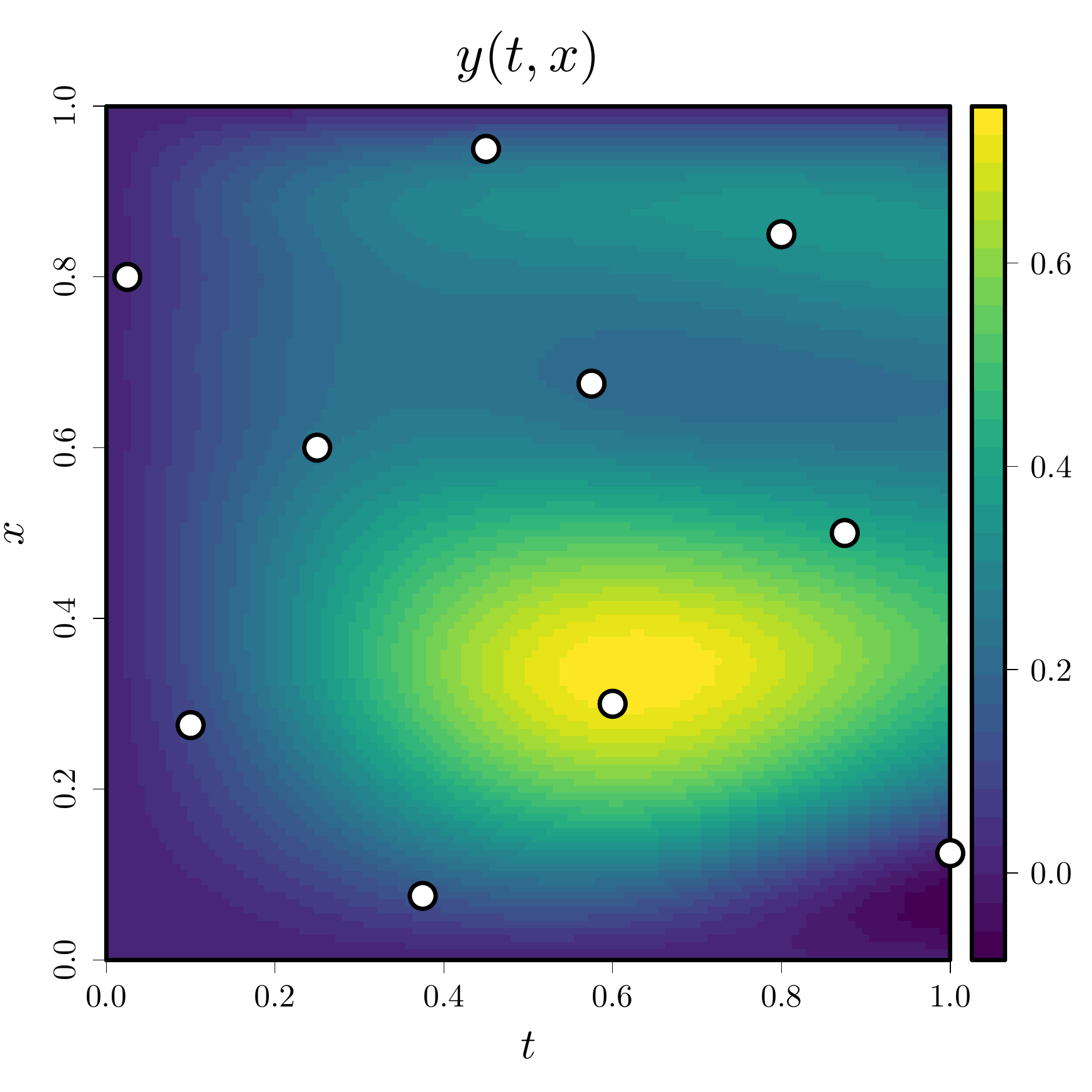}}
			\subfigure[$Q^2 = 0.276$, $\operatorname{CA}_{\pm\sigma} = 0.400$]{\includegraphics[width=0.325\columnwidth]{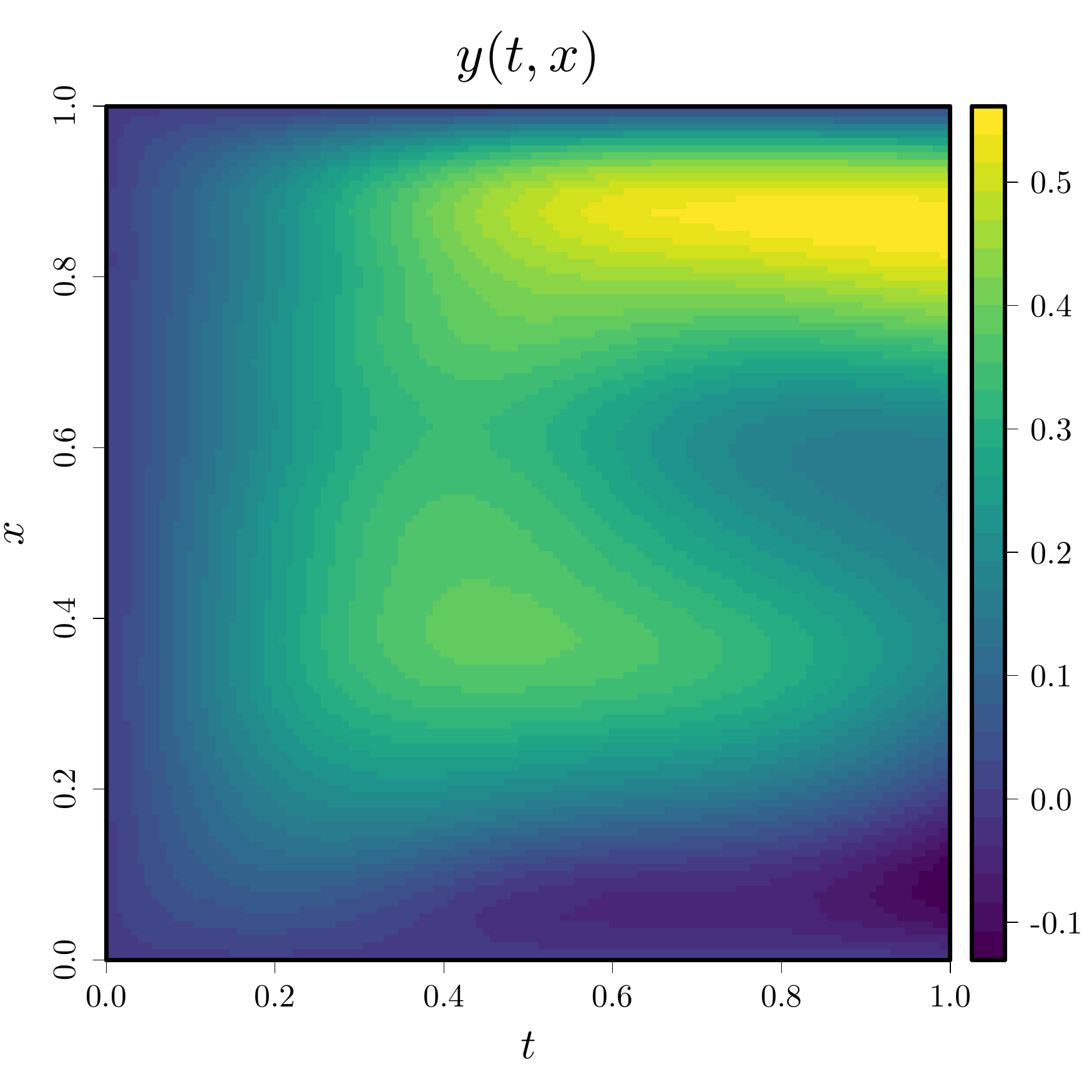}}
			\subfigure[$Q^2 = 0.950$, $\operatorname{CA}_{\pm\sigma} = 0.417$]{\includegraphics[width=0.325\columnwidth]{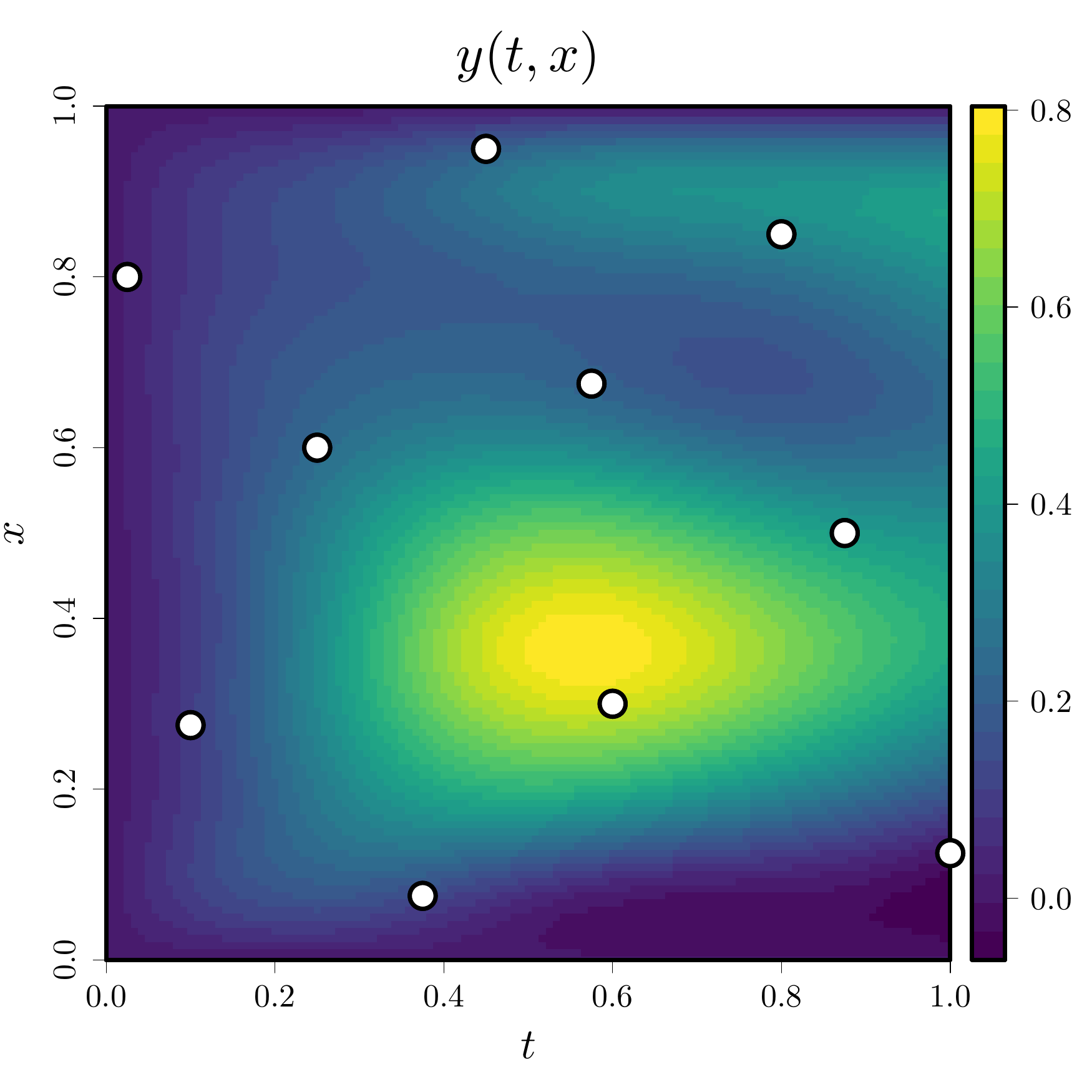}}
			\caption{GP-mRNA prediction results using conditioning data either only from the mRNA (left), or from the protein (centre), or from both of them (right). Conditioning points (white dots) were chosen using a maximin LHD with 10 points, and the quality of predictions is assessed using the $Q^2$ and $\operatorname{CA}_{\pm\sigma}$ criteria.}
			\label{fig:hGPmRNAinf1}
		\end{figure}
		
		Figure \ref{fig:hGPmRNAinf1} shows the performance of the GP-mRNA model using a maximin Latin hypercube design (LHD) at ten locations.\footnote{A maximin LHD is a space-filling design consisting in the iterative maximisation of the distance between two closest design points from a random LHD. In this paper, we used the simulated annealing routine \texttt{maximinSA\_LHS} from the R package \texttt{DiceDesign} \citep{Dupuy2015DiceDesign}.} For the case when only conditioning data from the protein concentration are used, we can observe that GP-mRNA presents accurate performances to reconstruct both the mRNA and protein profiles providing $Q^2$ results above $0.67$. Since conditioning data belongs to the protein profile where the mechanistic parameters were encoded, this information is taken into account in the inference of the mRNA. On the other hand, when only conditioning mRNA data are used, predictions over both quantities are poor. There, the influence of the PDE over the conditional process seems to be weak due to conditioning data belonging to the GP prior. Finally, one can observe that predictions are considerably improved when data from both the protein and mRNA are used. In that case, we obtained improvements of the $Q^2$ criterion obtaining values above $0.85$. Although the predictive variances were commonly underestimated, resulting $\operatorname{CA}_{\pm\sigma}$ values are not far from the expected $68\%$ (with departures of $15$-$28\%$).
		
\begin{figure}
			\centering
			\subfigure[$Q^2 = 0.988$, $\operatorname{CA}_{\pm\sigma} = 0.732$]{\includegraphics[width=0.325\columnwidth]{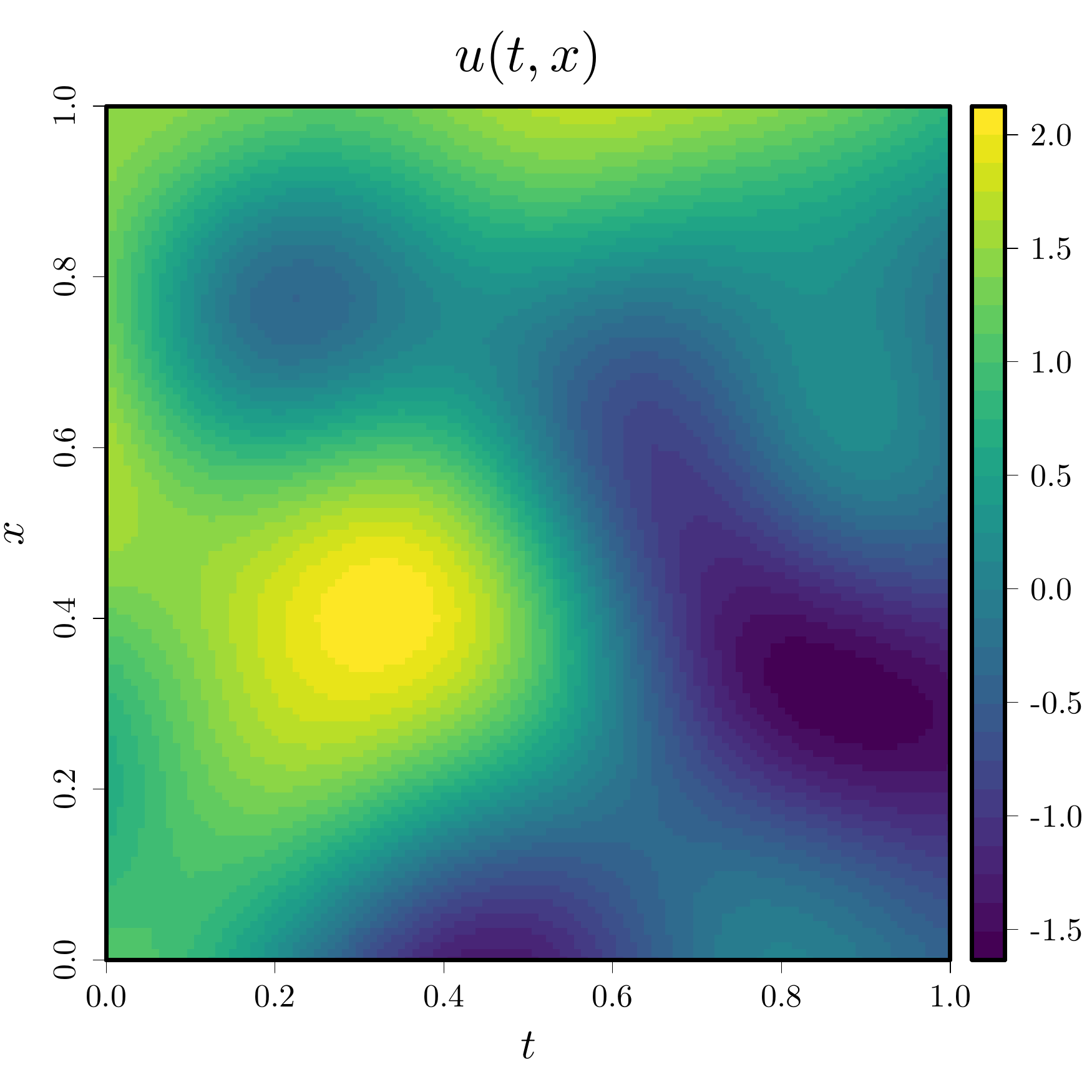}}
			\subfigure[$Q^2 = 0.994$, $\operatorname{CA}_{\pm\sigma} = 0.679$]{\includegraphics[width=0.325\columnwidth]{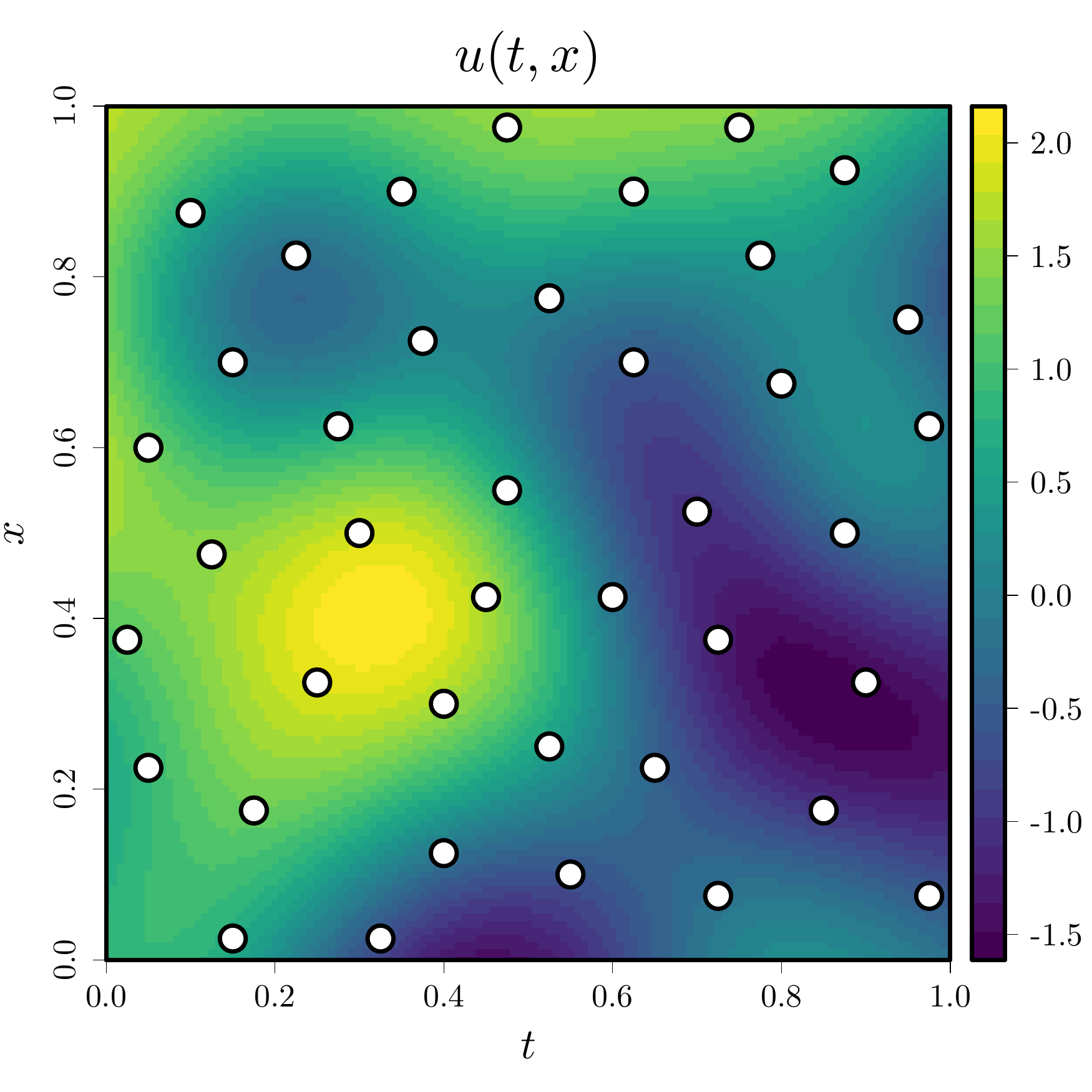}}
			\subfigure[$Q^2 = 1.000$, $\operatorname{CA}_{\pm\sigma} = 0.519$]{\includegraphics[width=0.325\columnwidth]{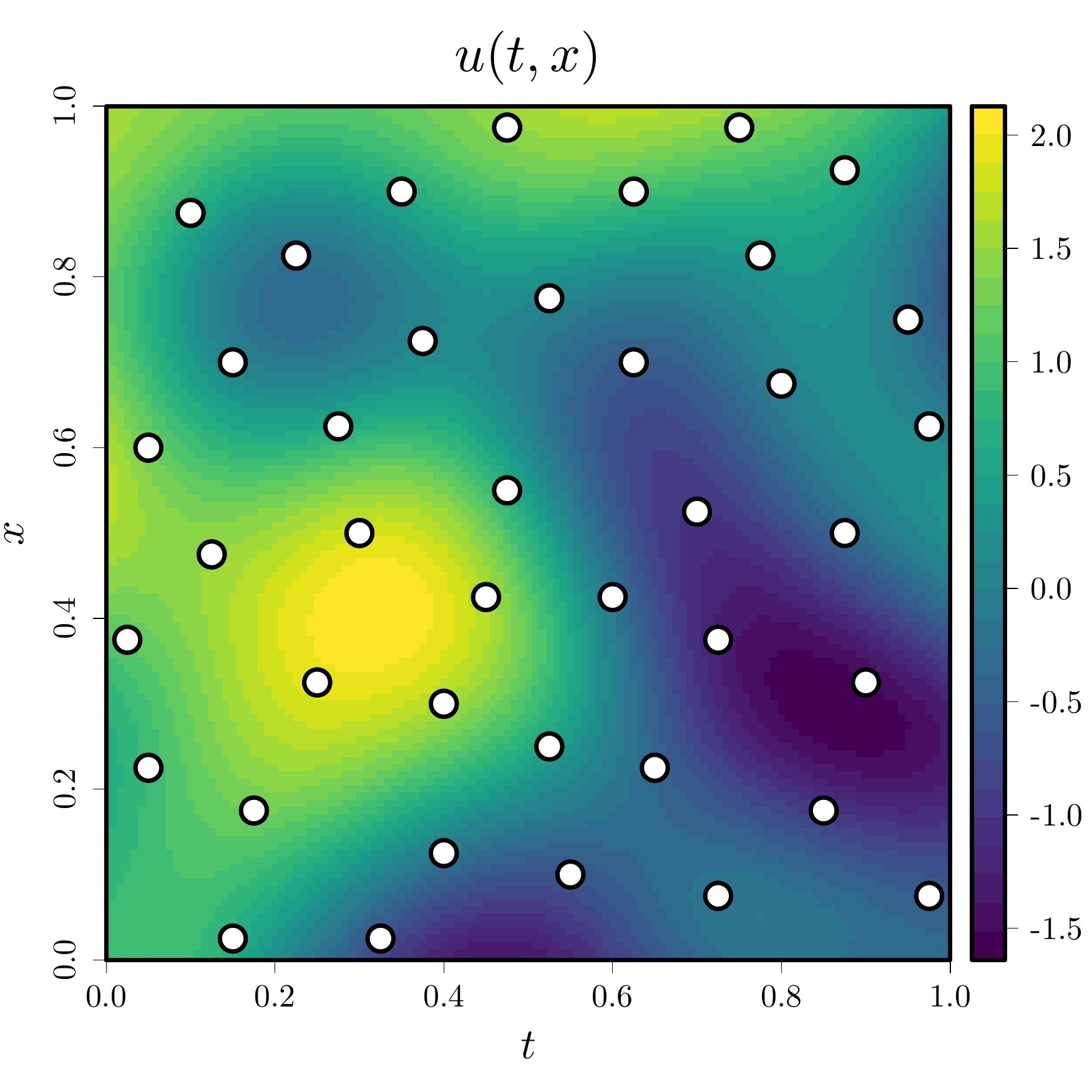}}			
			
			\subfigure[$Q^2 = 0.999$, $\operatorname{CA}_{\pm\sigma} = 0.739$]{\includegraphics[width=0.325\columnwidth]{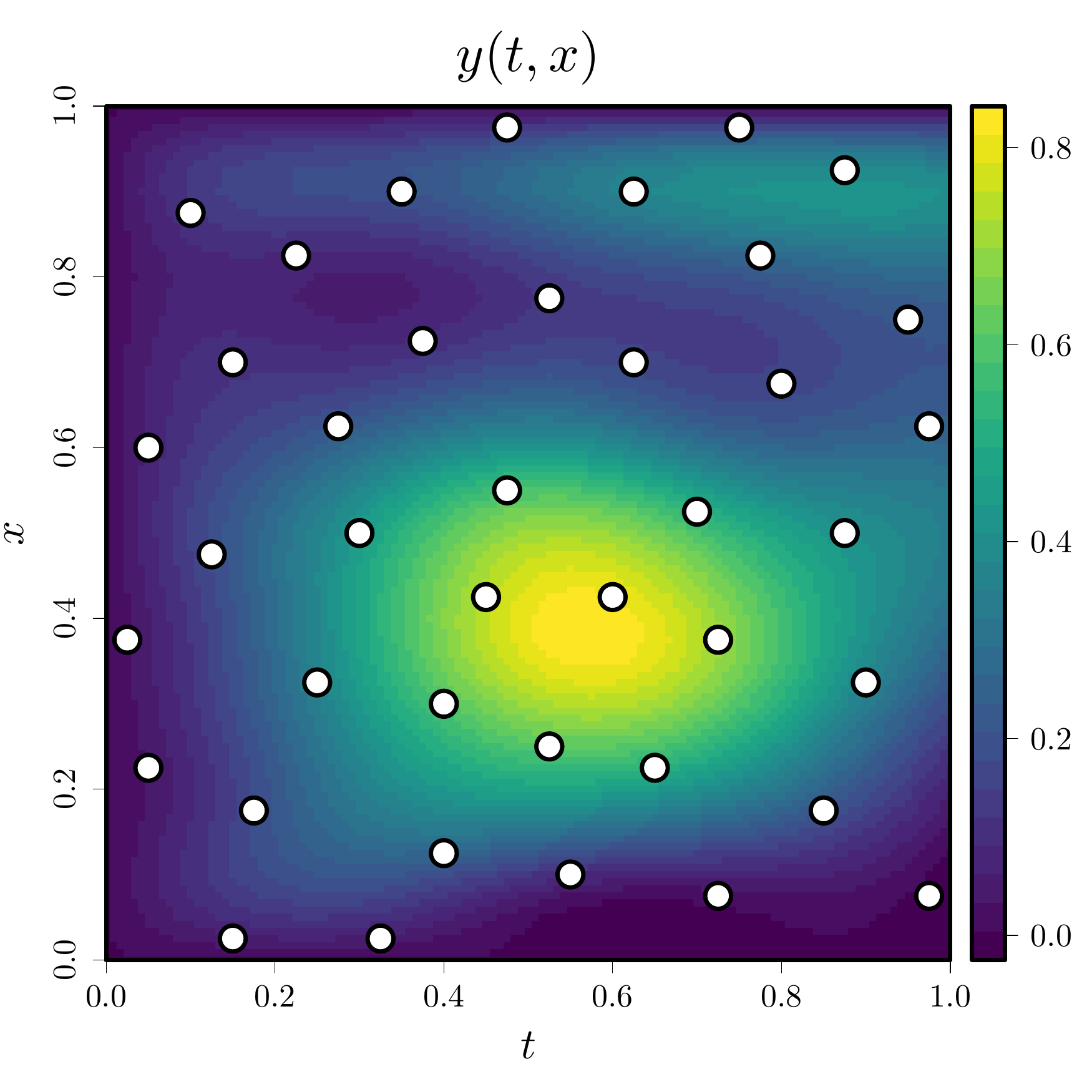}}			
			\subfigure[$Q^2 = 1.000$, $\operatorname{CA}_{\pm\sigma} = 0.821$]{\includegraphics[width=0.325\columnwidth]{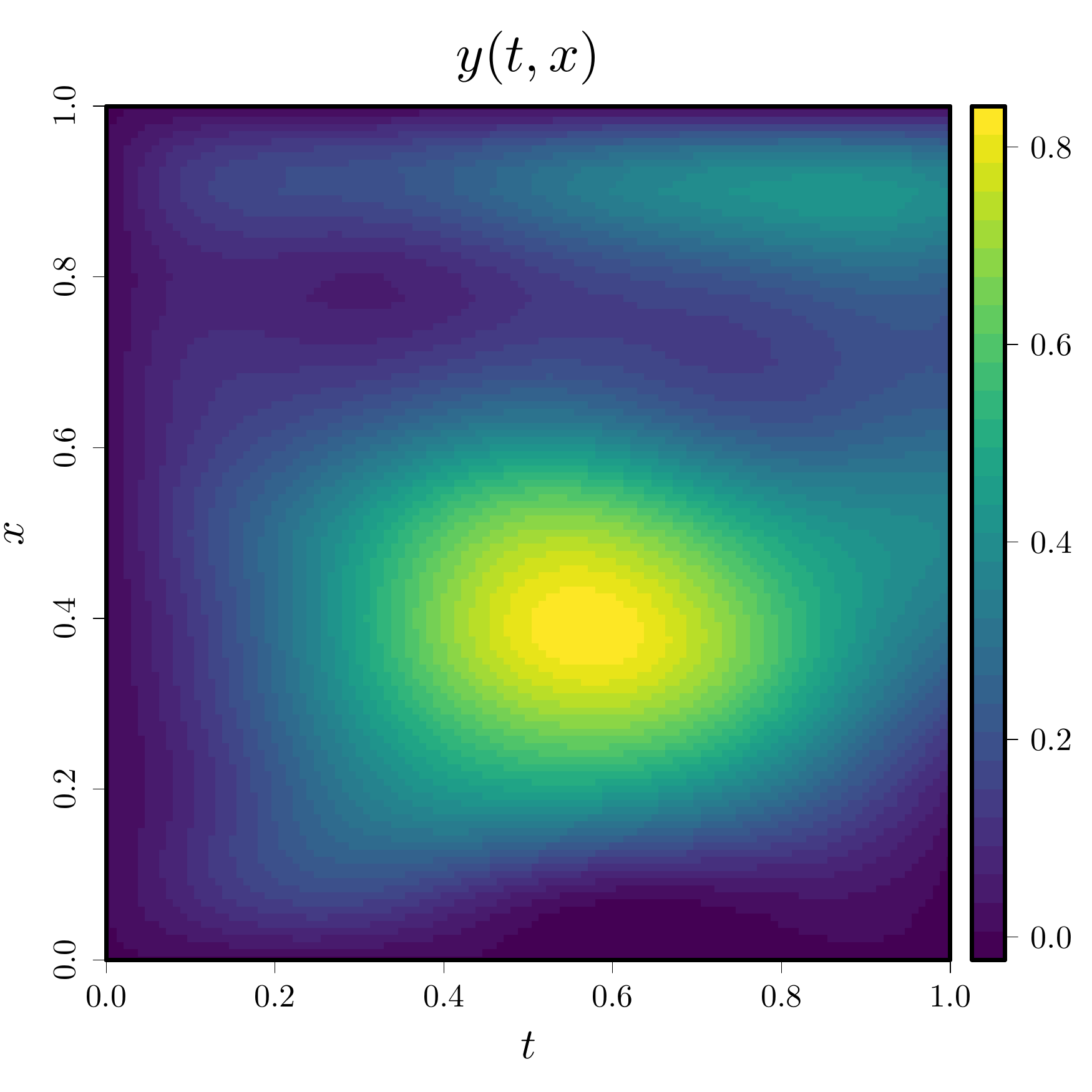}}
			\subfigure[$Q^2 = 1.000$, $\operatorname{CA}_{\pm\sigma} = 0.557$]{\includegraphics[width=0.325\columnwidth]{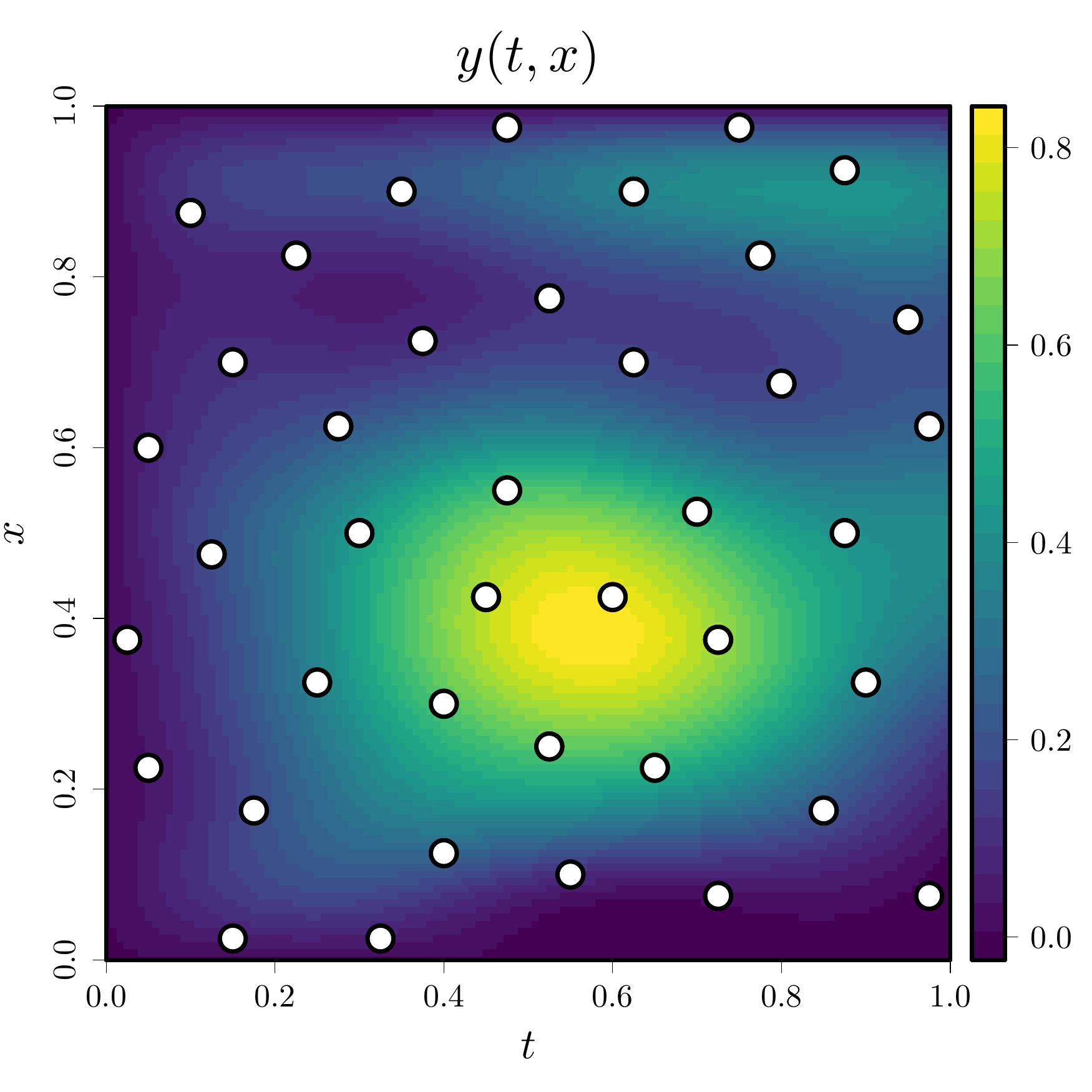}}
			\caption{GP-mRNA prediction results. Panel description is the same as in Figure \ref{fig:hGPmRNAinf1}. Conditioning data were chosen using a LHD with 40 points.}
			\label{fig:hGPmRNAinf2}
		\end{figure}

		In Figure \ref{fig:hGPmRNAinf2}, we show that if the number of conditioning points increases, the performance of the GP-mRNA model improves, obtaining $Q^2$ results above $0.98$ (and equal to one when data from both sides are used), and $\operatorname{CA}_{\pm\sigma}$ values closer to $68\%$ with maximum departures of $20\%$.
		
		\subsection{GP-Protein model}
		\label{subsec:hGPGapGene}
		As shown in Section \ref{subsec:hGPmRNA}, the establishment of the GP-mRNA models requires the explicit solution of the reaction-diffusion PDE in \eqref{eq:PDE} and the multiple-integration of kernel functions. Both calculations commonly require the evaluation of cumbersome terms which are not always feasible to compute (see discussion in Section \ref{subsec:hGPmRNA}). In this paper, we suggest placing the GP prior over the protein $y$ rather than over the mRNA $u$. This leads to a novel alternative where building up the resulting GP model is simpler since the solution of the PDE is not required.
		
		The reaction-diffusion dynamics from \eqref{eq:PDE} can be written in terms of the protein $y$, obtaining
		\begin{align}
		u(x,t) = \frac{1}{S} \bigg[ \partialdiff{y (x,t)}{t} + \lambda y(x,t) - D \partialdiff{^2 y (x,t)}{x^2} \bigg].
		\label{eq:hGPGapGeneEq}
		\end{align}
		As for the GP-mRNA model, we use the same SE kernel structure of \eqref{eq:hGPmRNAKuu} for the covariance function of the GP prior over $y$, i.e.
		\begin{equation}
		k_{y,y}(x,t,x',t')
		= \sigma^2 k(x,x') k(t,t'),
		\label{eq:hGPGapGeneKyy}
		\end{equation}
		where $k(z,z') = \exp\{-(z-z')^2/\theta_z^2\}$ with length-scale $\theta_z$.
		
		One can note that the establishment of the GP-Protein model is not restricted to the use of SE kernel functions, and that other classes of differentiable kernels can be used instead (e.g. Mat\'ern family of covariance functions) \citep{Rasmussen2005GP}.
		
		Now, we need to compute the covariance function for the driving-force $k_{u,u}$, and the cross-covariance function between the output and the driving-force $k_{y,u}$. For ease of readability, next we summarise the expressions required for the computation of $k_{u,u}$ and $k_{y,u}$. We refer to Appendix \ref{alg:hGPmodels} for further details.
		
		\subsubsection{Covariance function for the driving-force}
		Since \eqref{eq:hGPGapGeneEq} involves only the differentiation of the output process $y$, and due to the symmetry of SE kernel functions, the covariance function for the mRNA is given by
		\begin{align}
		k_{u,u}(x,t,x',t')
		= \frac{\sigma^2 D}{S^2}\left[D k^{iv}(x,x') - 2 \lambda k^{ii}(x,x') \right] k(t,t')
		- \frac{\sigma^2}{S^2} \left[k^{ii}(t,t') - \lambda^2 k(t,t')\right] k(x,x'),
		\label{eq:hGPGapGeneKuu}
		\end{align}
		where $k^{j}(z,z')$ is the $j$-th derivative of the SE kernel $k(z,z')$ w.r.t. the input $z$. Then, the complexity of the problem is in the computation of the corresponding derivatives of the SE kernel function, and they follow
		\begin{align}
		&k^{i}(z,z') = \left[-\frac{2(z-z')}{\theta^2_z}\right] k(z,z'), \nonumber
		\\
		&k^{ii}(z,z') = \left[-\frac{2}{\theta^2_z} + \frac{4(z-z')^2}{\theta^4_z}\right] k(z,z'), \label{eq:hGPGapGeneSEdiff}
		\\
		&k^{iv}(z,z') = \left[\frac{12}{\theta^4_z} - \frac{48(z-z')^2}{\theta^6_z}+ \frac{16(z-z')^4}{\theta^8_z} \right] k(z,z'). \nonumber
		\end{align}
		Notice that, since we have to differentiate partially $k_{y,y}$ four times, the GP-Protein model is limited to the use of differentiable kernels (e.g. Mat\'ern family of covariance functions with regularity parameter $\nu>\frac{5}{2}$) \citep{Rasmussen2005GP,stein1999interpolation}.
		
		\begin{figure}[t!]
			\centering
			\subfigure[mRNA]{\includegraphics[width=0.325\columnwidth]{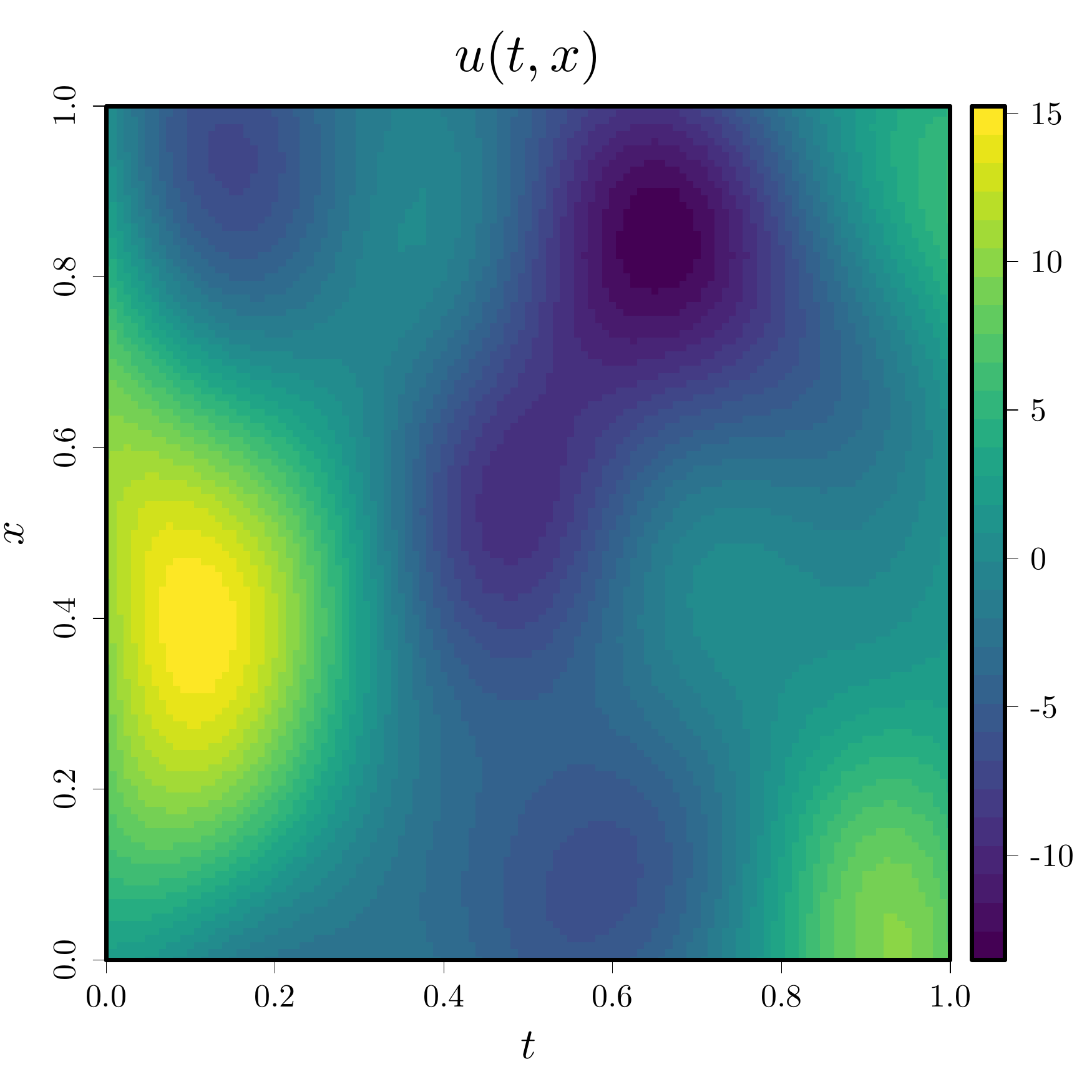}}
			\subfigure[Protein]{\includegraphics[width=0.325\columnwidth]{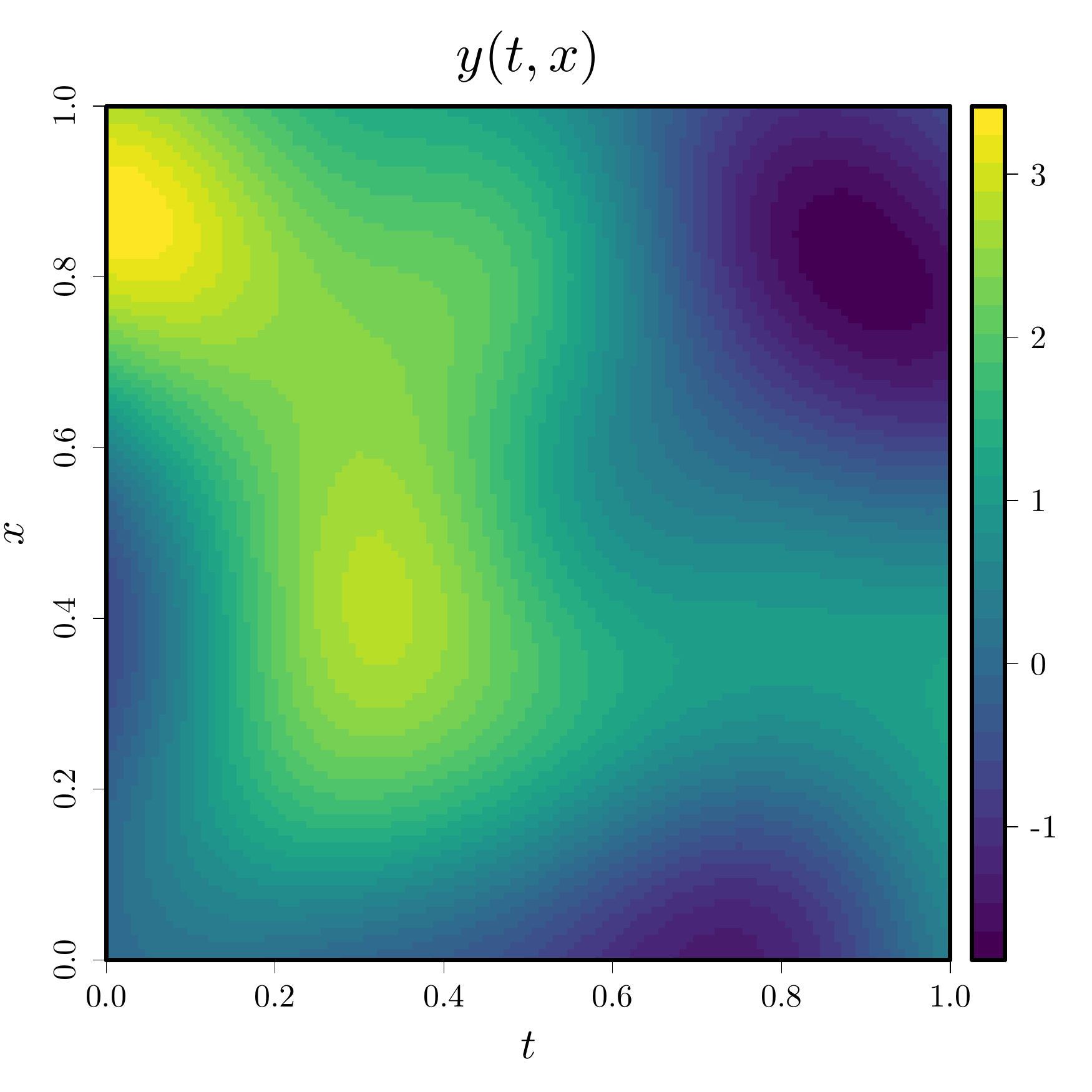}}
			\caption{Synthetic example generated by the GP-Protein model.}
			\label{fig:hGPGapGenesample}
			\subfigure[ $Q^2 = 0.373$, $\operatorname{CA}_{\pm\sigma} = 0.504$]{\includegraphics[width=0.325\columnwidth]{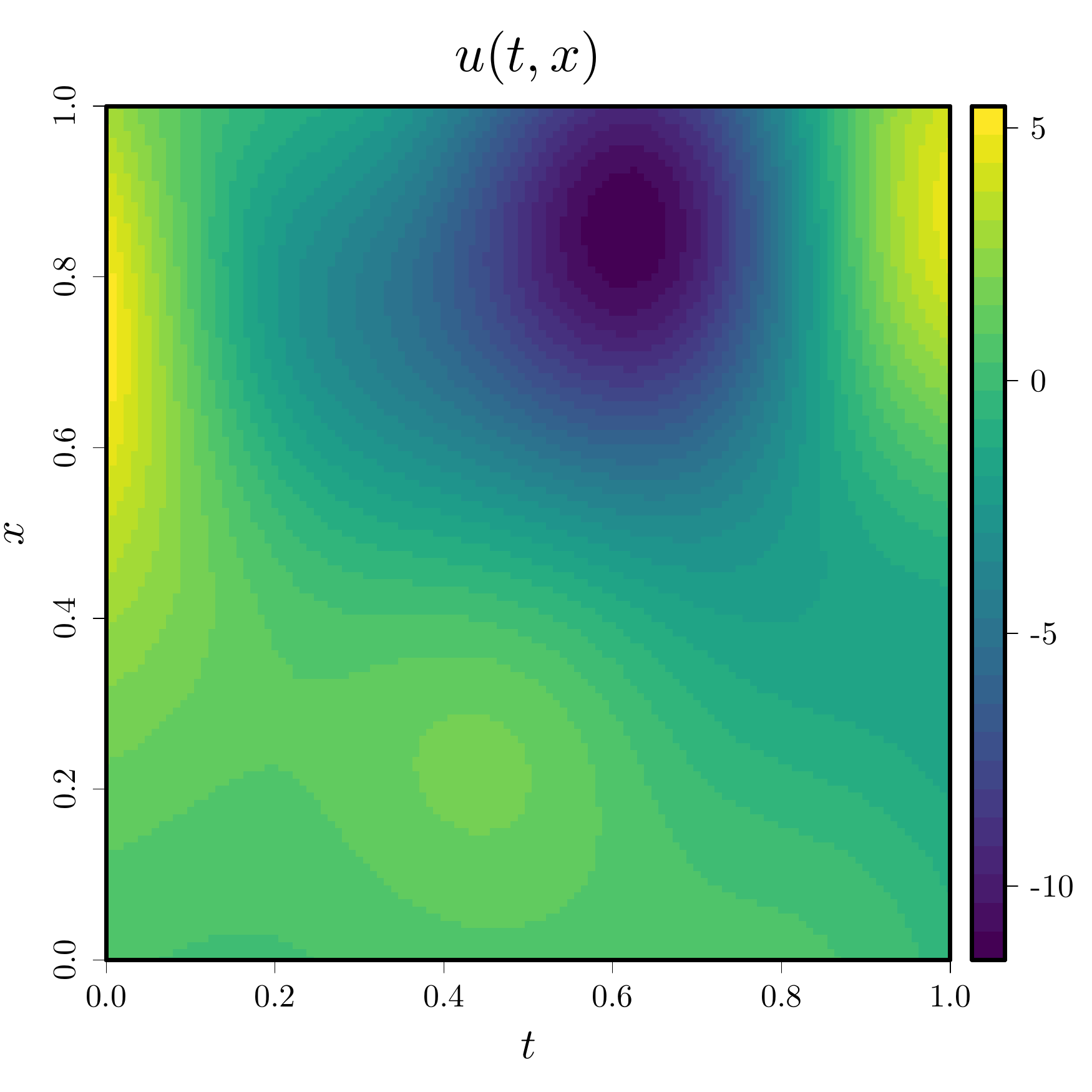}}
			\subfigure[ $Q^2 = 0.756$, $\operatorname{CA}_{\pm\sigma} = 0.685$]{\includegraphics[width=0.325\columnwidth]{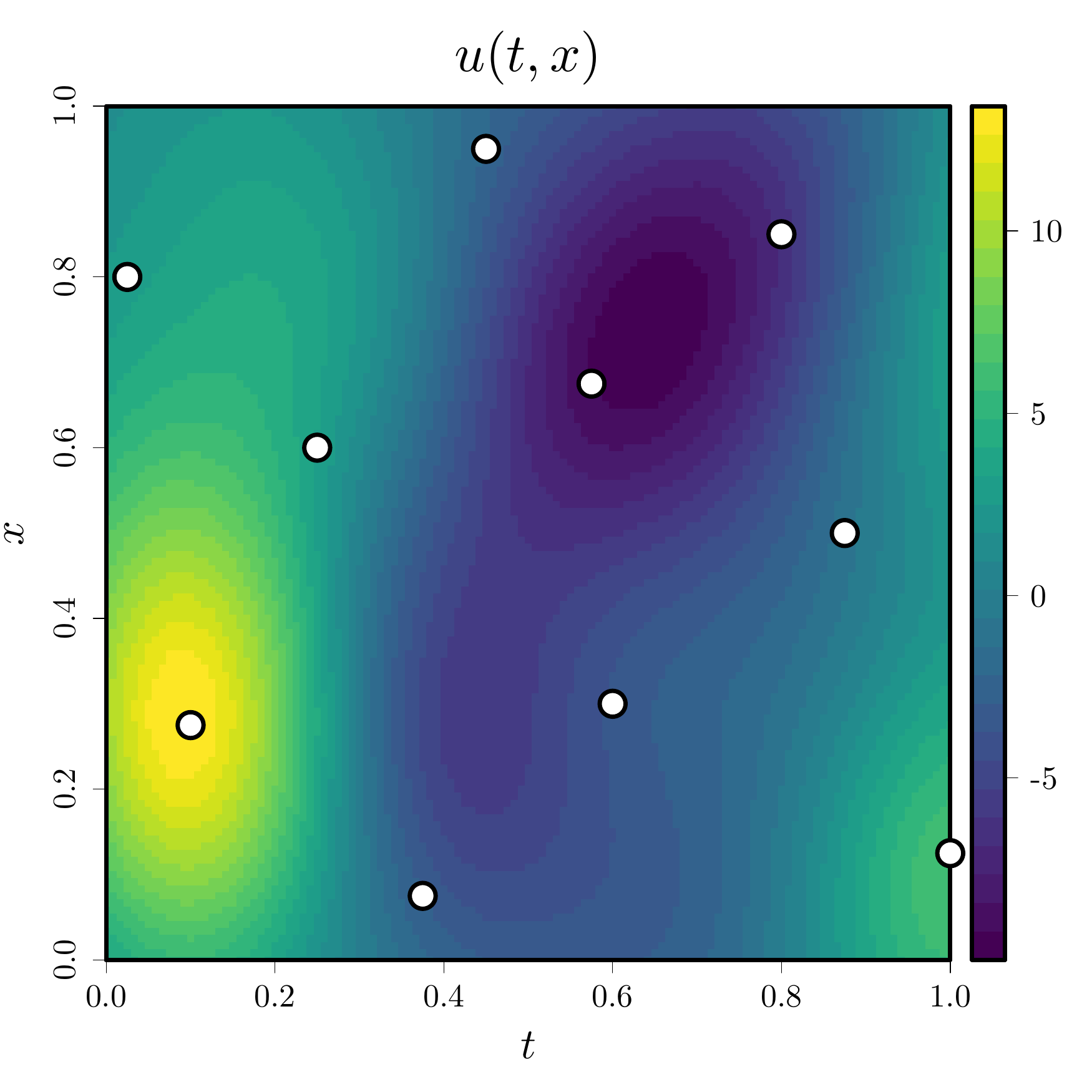}}
			\subfigure[ $Q^2 = 0.893$, $\operatorname{CA}_{\pm\sigma} = 0.571$]{\includegraphics[width=0.325\columnwidth]{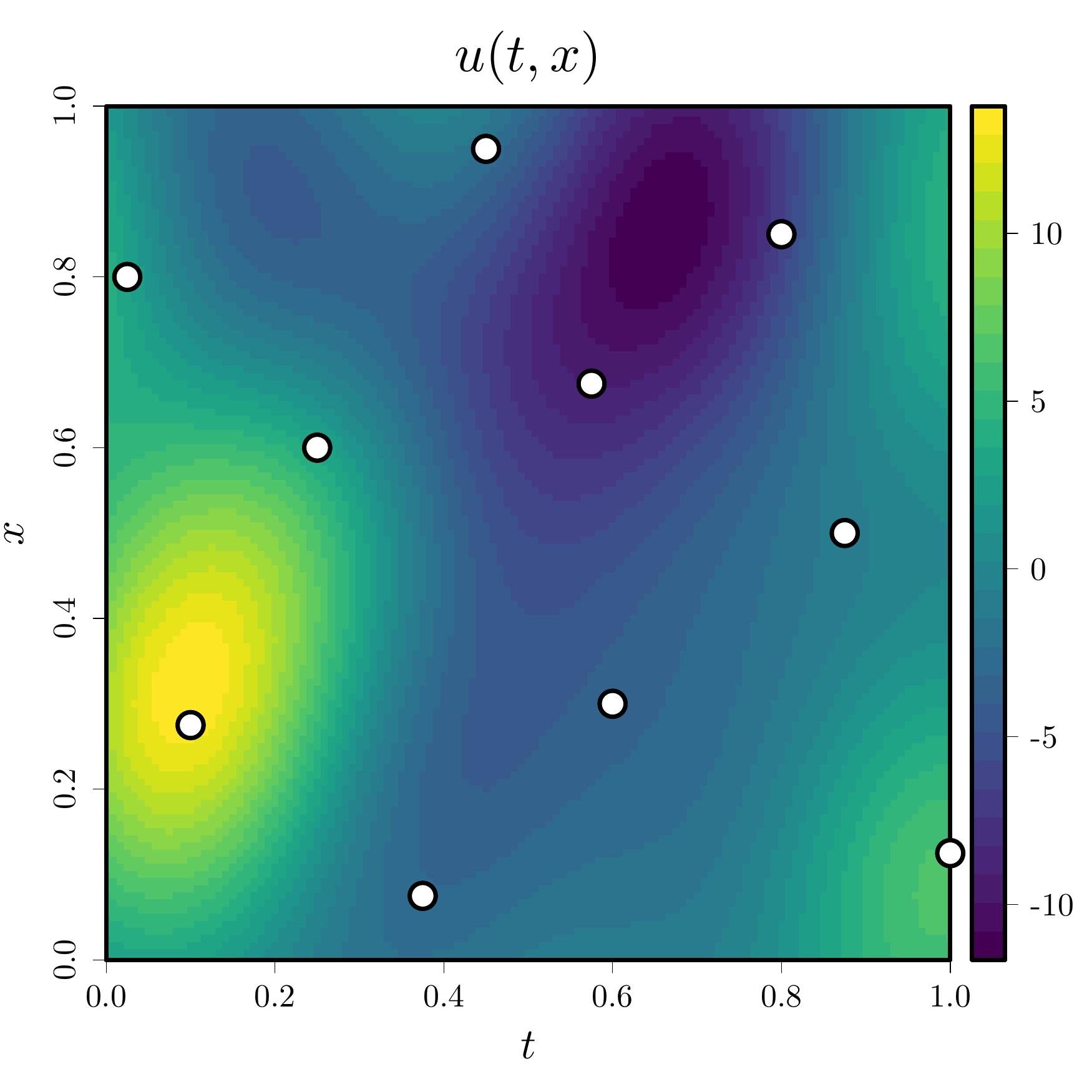}}
			
			\subfigure[ $Q^2 = 0.714$, $\operatorname{CA}_{\pm\sigma} = 0.558$]{\includegraphics[width=0.325\columnwidth]{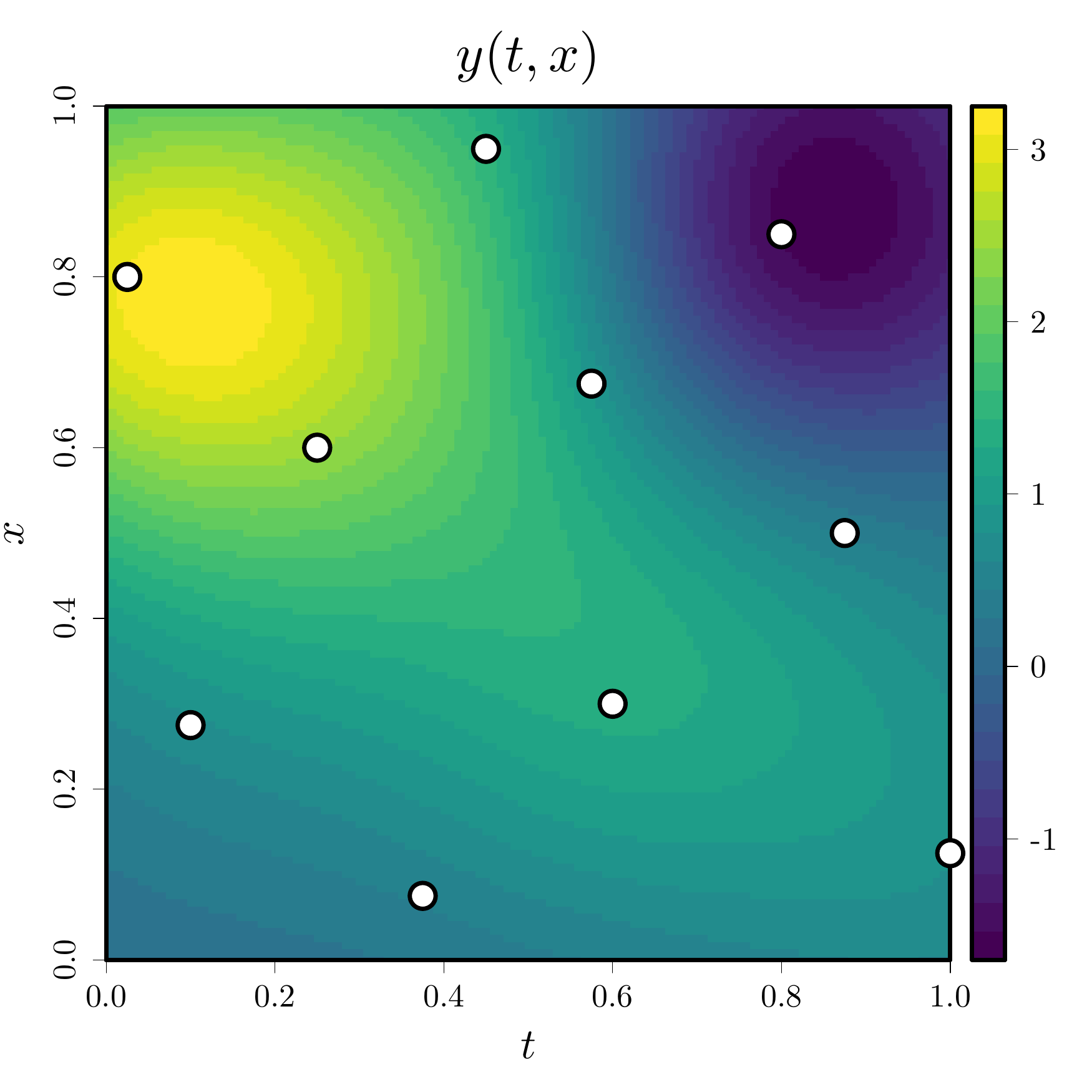}}
			\subfigure[ $Q^2 = 0.467$, $\operatorname{CA}_{\pm\sigma} = 0.555$]{\includegraphics[width=0.325\columnwidth]{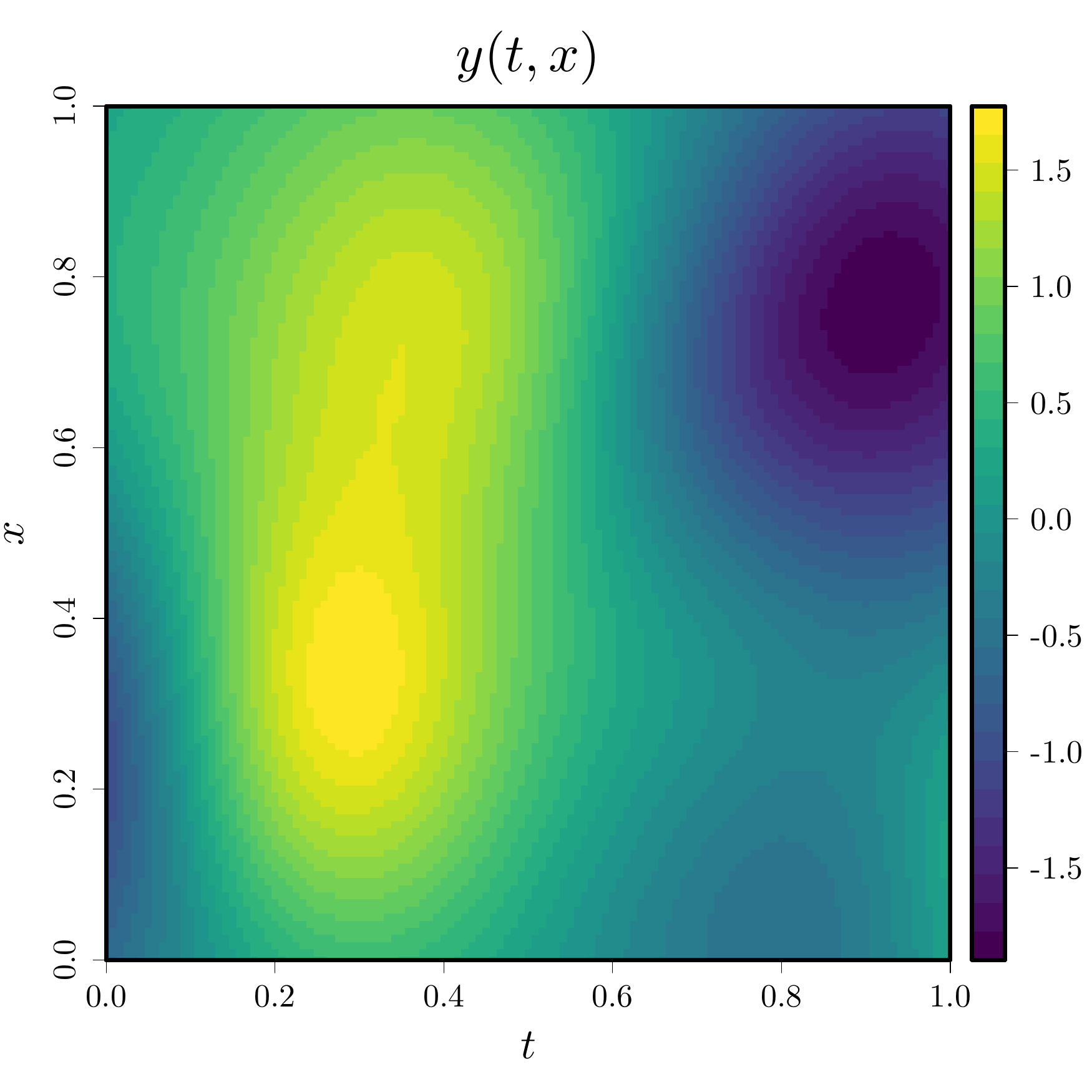}}
			\subfigure[ $Q^2 = 0.948$, $\operatorname{CA}_{\pm\sigma} = 0.683$]{\includegraphics[width=0.325\columnwidth]{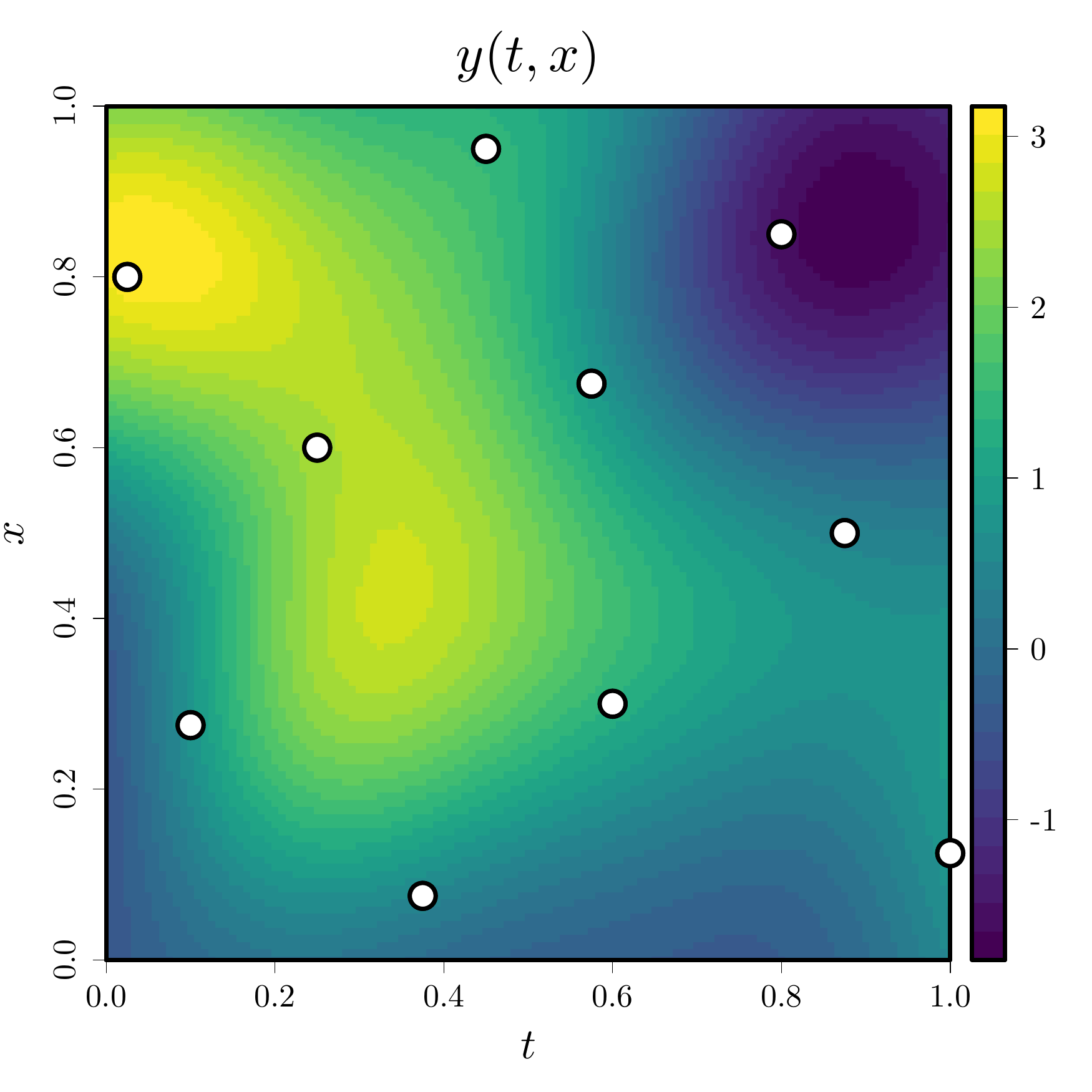}}
			\caption{GP-Protein prediction results using conditioning data either only from the mRNA (left), or from the protein (centre), or from both of them (right). Conditioning points (white dots) were chosen using a maximin LHD with 10 points, and the quality of predictions is assessed using the $Q^2$ and $\operatorname{CA}_{\pm\sigma}$ criteria.}
			\label{fig:hGPGapGeneinf1}
		\end{figure}
		
		\subsubsection{Covariance function between the driving-force and the output}
		The cross-covariance function between the output $y$ and the force $u$, $k_{y,u}(x,t,x',t') = \cov{y(x,t),u(x',t')}$ is given by
		\begin{align}
		k_{y,u}(x,t,x',t') =
		\frac{\sigma^2}{S} [ \lambda k(x,x') k(t,t') 
		- k(x,x') k^{i}(t,t') - D k^{ii}(x,x') k(t,t')],
		\label{eq:hGPGapGeneKyu}
		\end{align}
		with derivatives of the SE kernel given in \eqref{eq:hGPGapGeneSEdiff}. 
		\begin{figure}
			\centering
			\subfigure[ $Q^2 = 0.990$, $\operatorname{CA}_{\pm\sigma} = 0.857$]{\includegraphics[width=0.325\columnwidth]{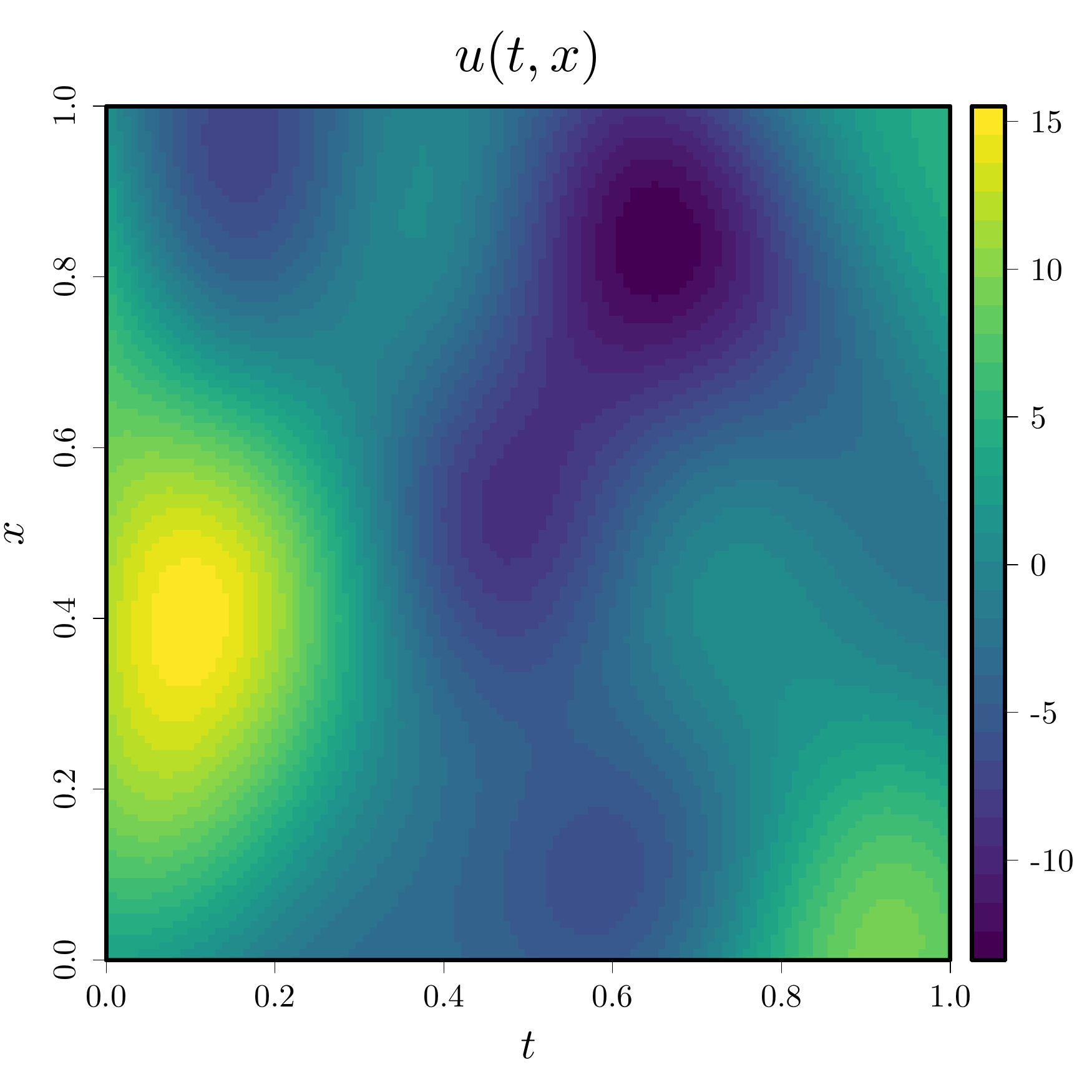}}
			\subfigure[ $Q^2 = 0.997$, $\operatorname{CA}_{\pm\sigma} = 0.807$]{\includegraphics[width=0.325\columnwidth]{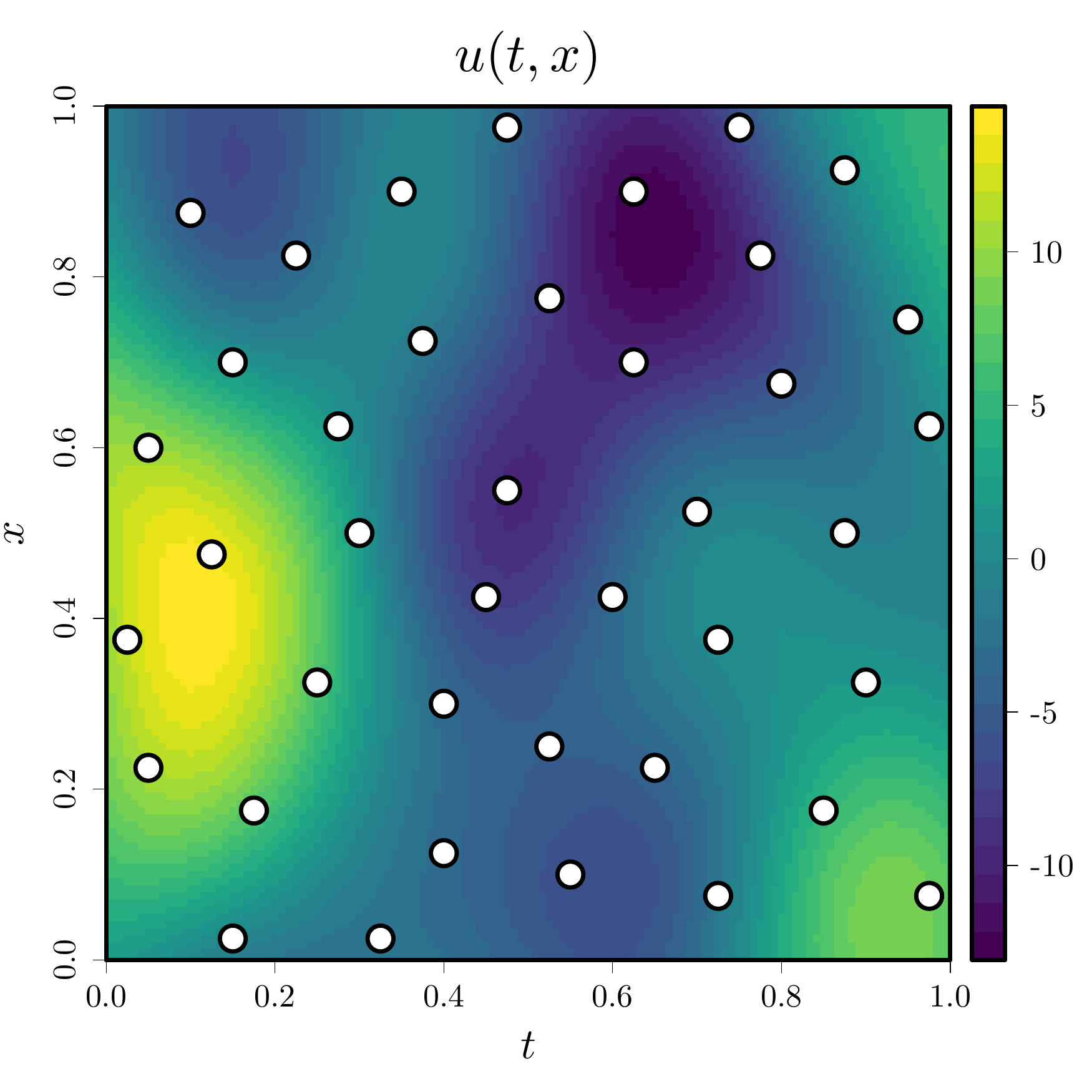}}
			\subfigure[ $Q^2 = 0.999$, $\operatorname{CA}_{\pm\sigma} = 0.678$]{\includegraphics[width=0.325\columnwidth]{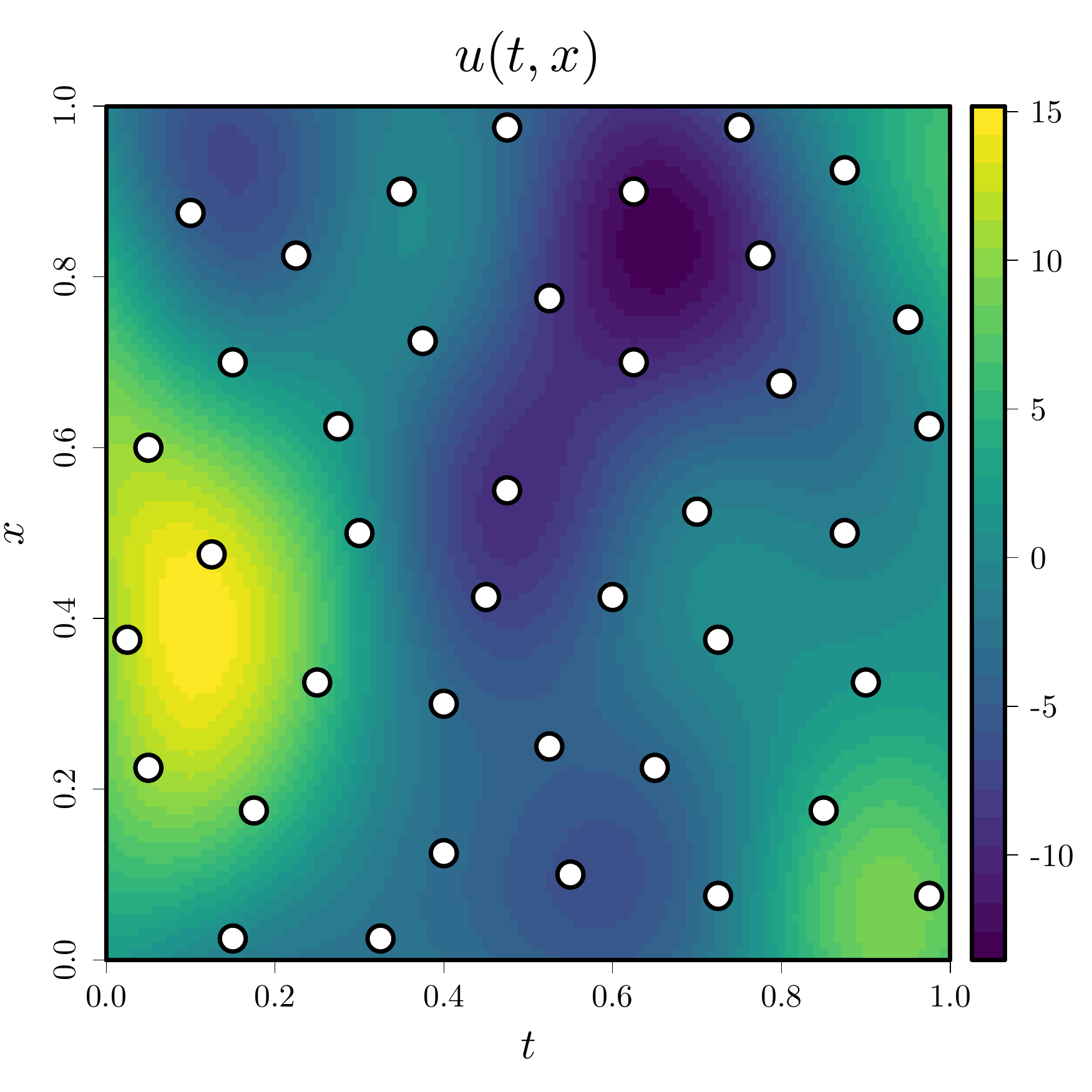}}			
			
			\subfigure[ $Q^2 = 0.999$, $\operatorname{CA}_{\pm\sigma} = 0.871$]{\includegraphics[width=0.325\columnwidth]{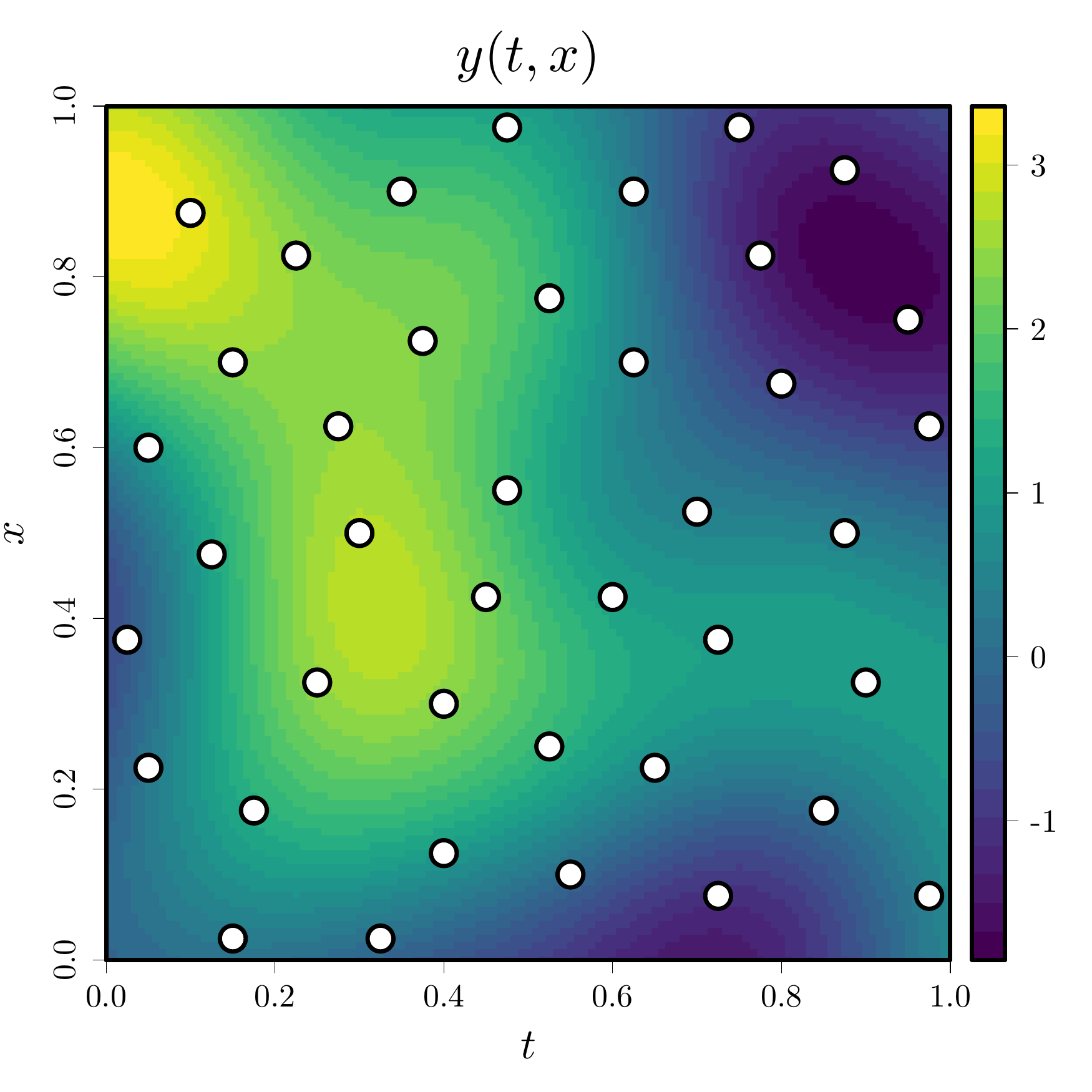}}			
			\subfigure[ $Q^2 = 0.656$, $\operatorname{CA}_{\pm\sigma} = 0.439$]{\includegraphics[width=0.325\columnwidth]{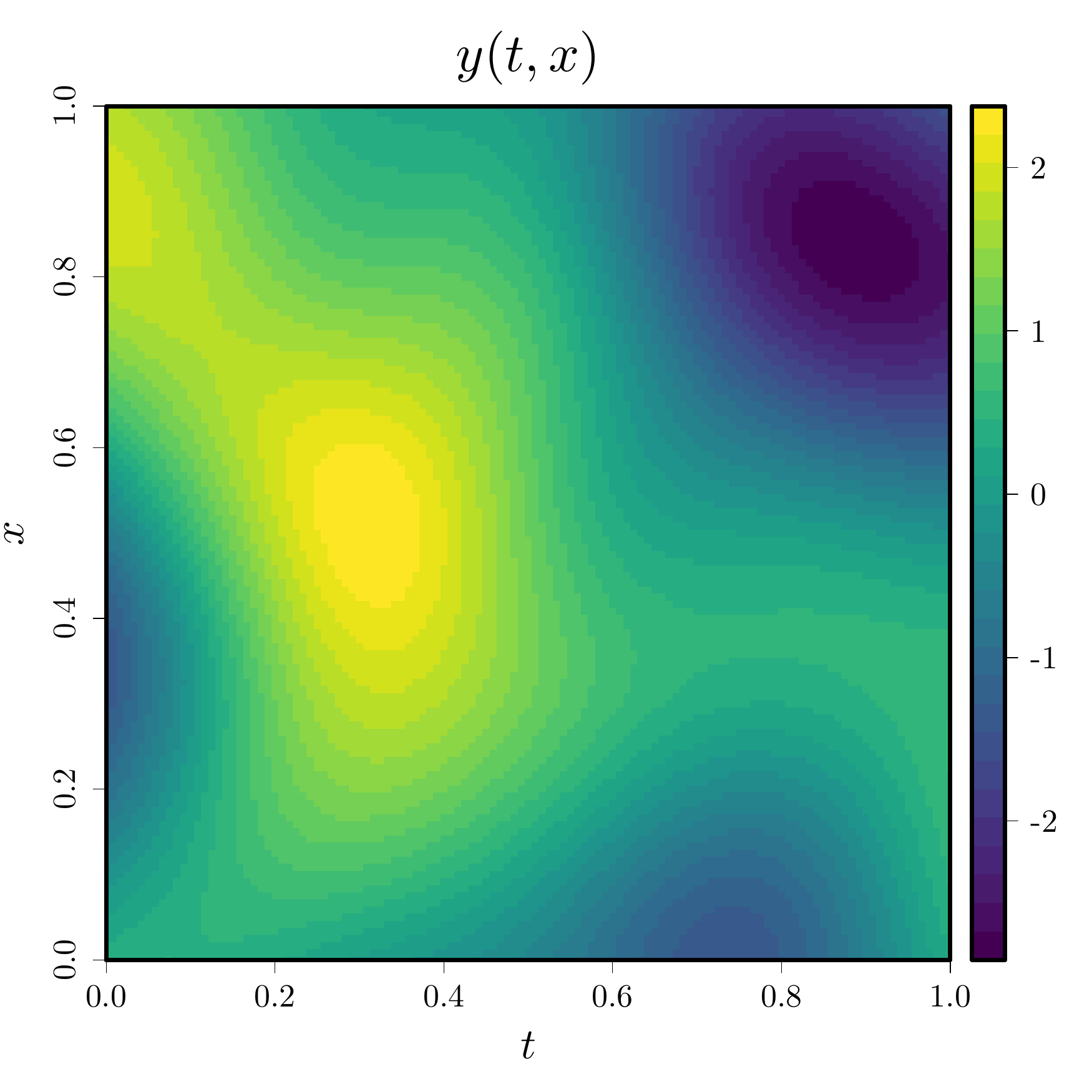}}
			\subfigure[ $Q^2 = 1.000$, $\operatorname{CA}_{\pm\sigma} = 0.690$]{\includegraphics[width=0.325\columnwidth]{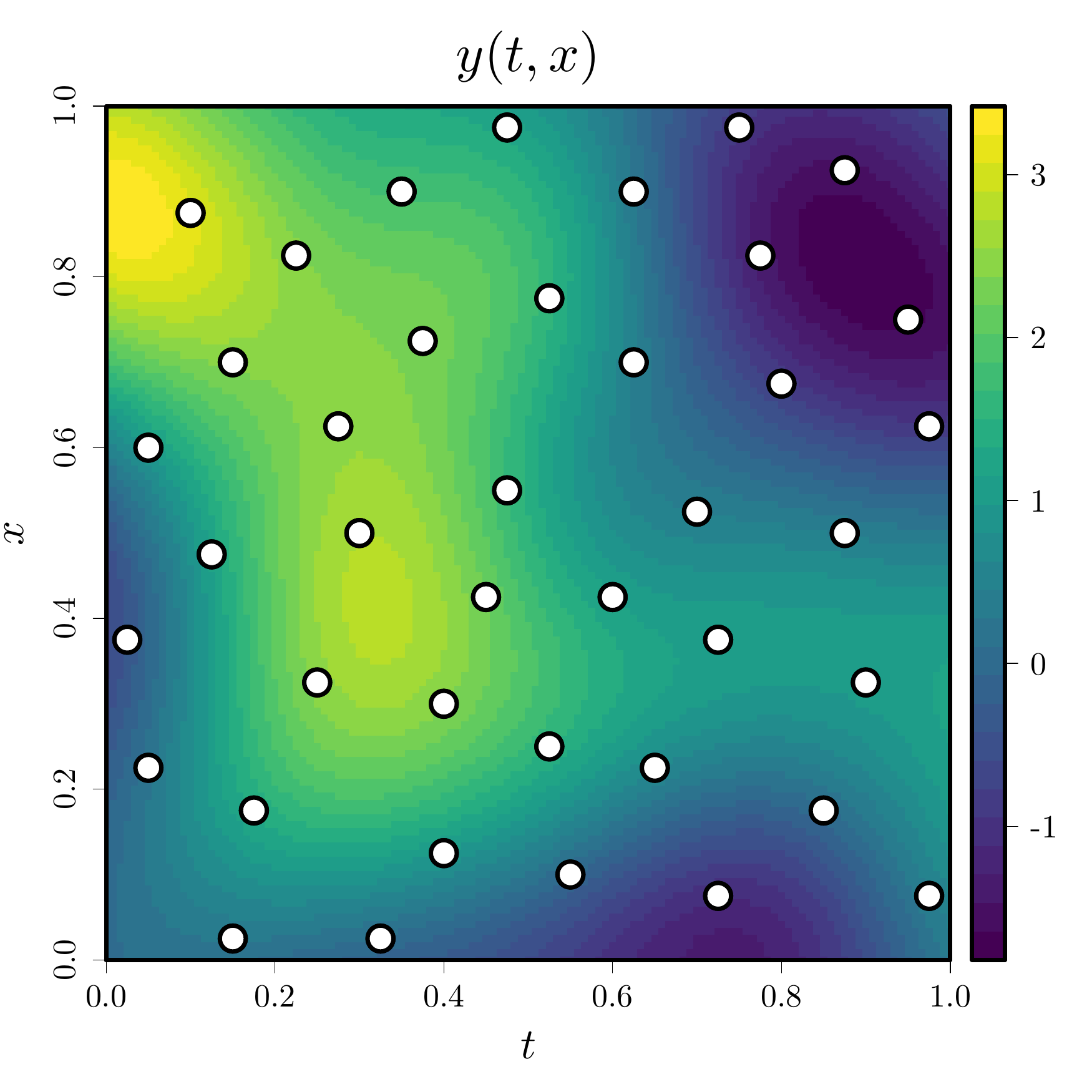}}
			\caption{GP-Protein prediction results. Panel description is the same as in Figure \ref{fig:hGPGapGeneinf1}. Conditioning data were chosen using a LHD with 40 points.}
			\label{fig:hGPGapGeneinf2}	
		\end{figure}
		\subsubsection{Toy example: inference of simulated data}
		\label{subsubsec:hGPGapGenetoy}
		As in Section \ref{subsubsec:hGPmRNAtoy}, we generate a synthetic example by sampling from the GP in \eqref{eq:jointGP} using the kernel functions \eqref{eq:hGPGapGeneKyy}, \eqref{eq:hGPGapGeneKuu} and \eqref{eq:hGPGapGeneKyu}. We assume the same parametrisation used for GP-mRNA. Figure \ref{fig:hGPGapGenesample} shows the generated mRNA and protein. One can observe that since we did not enforce the initial and boundary conditions, homogeneous conditions are not necessarily ensured by the GP-Protein model.
		
		Now, we test the performance of the GP-Protein model under the three conditions studied in Section \ref{subsubsec:hGPmRNAtoy}. Figure \ref{fig:hGPGapGeneinf1} shows the performance of GP-Protein using a fixed maximin LHD at ten locations. As observed in Section \ref{subsubsec:hGPmRNAtoy}, if only conditioning points are used from the mRNA or protein, predictions over the unobserved quantity are less reliable; and they improve when data are available from both sides. In the latter case, we obtain $Q^2$ values above $0.87$ and $\operatorname{CA}_{\pm\sigma}$ values around $57$-$68\%$. In figure \ref{fig:hGPGapGeneinf2}, one can observe that if the number of conditioning points increases, the performance of the GP-Protein also improves with $Q^2$ values close to one in almost all the cases, and $\operatorname{CA}_{\pm\sigma}$ values almost equals to $68\%$ when using data from both quantities.
		
		\section{Results and Discussions}
		\label{sec:results}
		\subsection{Numerical setup}
		\label{subsec:setup}
		Both physically-inspired GP approaches for modelling the post-transcriptional regulation of the early embryo of Drosophila melanogaster were implemented in R, and the codes are available on Github: \url{https://github.com/anfelopera/PhysicallyGPDrosophila}. They are based on the R package \texttt{kergp}.\footnote{The \texttt{kergp} project is an open source R package available in CRAN for Gaussian Process models with customised covariance kernels \citep{kergp2015}.} For the computation of the $\operatorname{erf}$ and Faddeeva functions, after testing the numerical stability of various R packages, we chose the \textit{pracma} \citep{pracma2018} and \texttt{NORMT3} \citep{NORMT32015} packages. The hyperparameters $\Btheta = (S, \lambda, D, \sigma^2, \theta_x, \theta_t)$ are estimated by maximising the joint marginal log-likelihood $p_\Btheta(\Bu, \By)$ using gradient-descent methods \citep{Rasmussen2005GP}, i.e. $\widehat{\Btheta} = \arg \max_{\Btheta} \log \{p_\Btheta(\Bu, \By)\}$. Algorithm \ref{alg:hGPmodels} shows the pseudo-code of both GP-mRNA and GP-Protein for the post-transcriptional regulation of Drosophila.
		
		\begin{algorithm}[t!]
			\begin{minipage}{\columnwidth}
				\small	
				\caption{Physically-inspired GPs for transcriptional regulation of the early embryo of Drosophila melanogaster.}\label{alg:hGPmodels}
				\begin{algorithmic}[1]
					\Procedure{Prediction of mRNA $u$ and Gap Protein $y$}{}
					\BState Input: training set $\mathcal{D} = (\textbf{x}, \textbf{t}, \Bu, \By)$,\footnote{One may note that is not necessary to have access to conditioning data from both $y$ and $u$ simultaneously.} initial set of hyperparameters $\Btheta = \{S, \lambda, D, \sigma^2, \theta_x, \theta_y\}$
					\BState Compute the covariance matrices $\BK_{\By,\By}$, $\BK_{\Bu,\Bu}$, and $\BK_{\By,\Bu}$ according to Section \ref{sec:hGP}.
					\BState Estimate the hyperparameters $\widehat{\Btheta} = \arg \max_{\Btheta} \log \{p_\Btheta(\Bu, \By)\}$.
					\BState According to \citep{Rasmussen2005GP}, compute the conditional distribution for the test set $\mathcal{D}^\ast = (\textbf{x}^\ast, \textbf{t}^\ast)$, i.e. $p(\Bu^\ast, \By^\ast|\mathcal{D})$.
					\EndProcedure
				\end{algorithmic}
			\end{minipage}
		\end{algorithm}
		
		\subsection{Quantitative gap gene mRNA expression data}
		\label{subsec:dataset}
		Here we aim at testing the performance of both physically-inspired GP models from Section \ref{sec:hGP} on the high spatial and temporal resolution dataset used in \citep{Becker2013DrosMel}, describing the entire duration of the blastoderm stage for the early embryo of Drosophila melanogaster. This dataset exhibits homogeneous conditions, and it contains quantified independent time-series of gap gene mRNA expressions for the trunk gap genes \textit{Kr\"uppel} ($kr$), \textit{knirps} ($kni$) and \textit{giant} ($gt$). We ignored the observations in the range $t \in [0, 50]$ due to the poor quality of the data, and we focused on the late part where the data is equispaced: after the first 53 min and around A-P positions of $x = 25.5\%$ for $kr$, and $x = 32.5\%$ for $kni$ and $gt$. This leads to a total of 512, 456 and 512 measurements for the trunk gap genes $kr$, $kni$ and $gt$, respectively.
		
\begin{figure}[t!]
			\centering
			\rotv{\hskip 8ex  mRNA Expressions}
			\subfigure{\includegraphics[width=0.32\columnwidth]{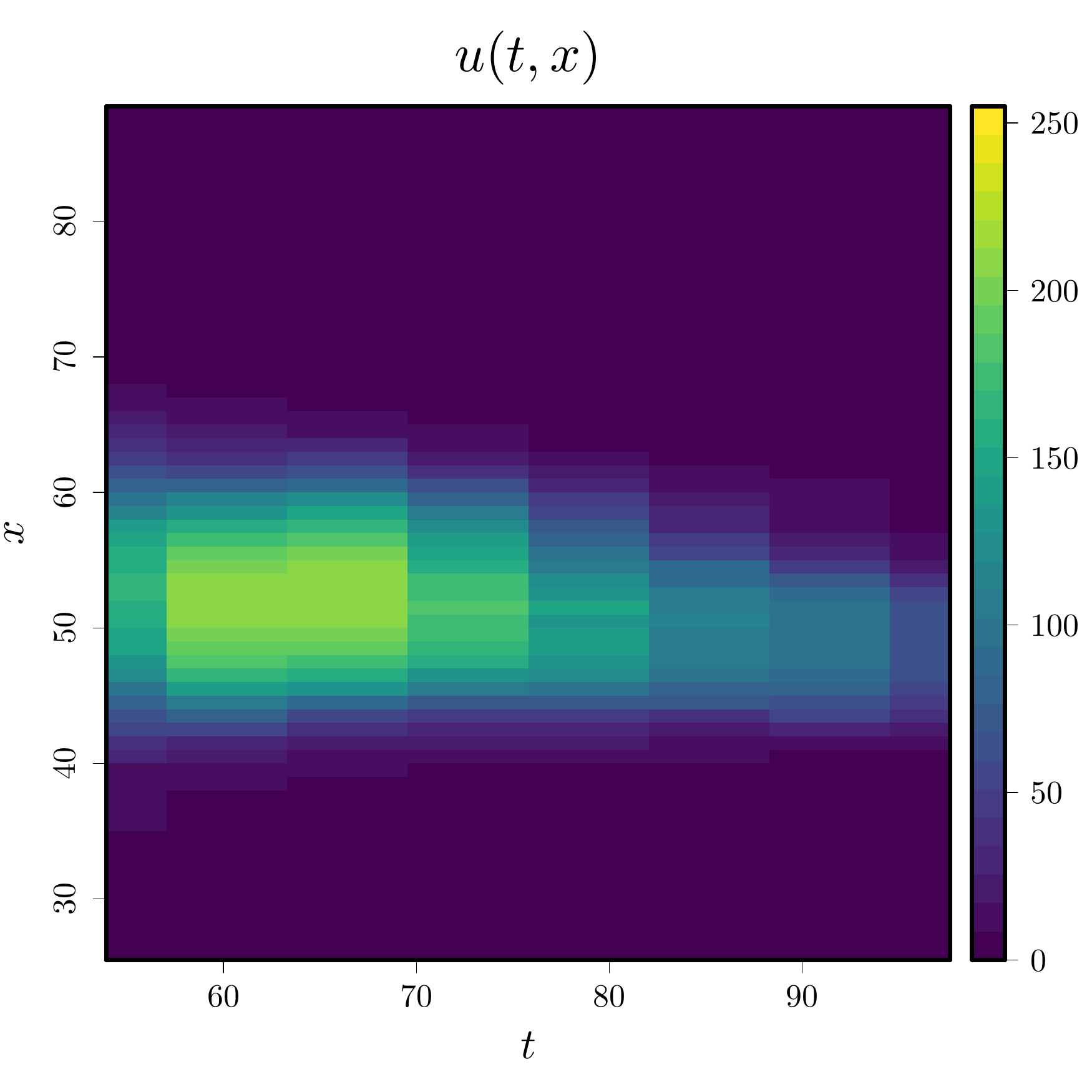}}
			\subfigure{\includegraphics[width=0.32\columnwidth]{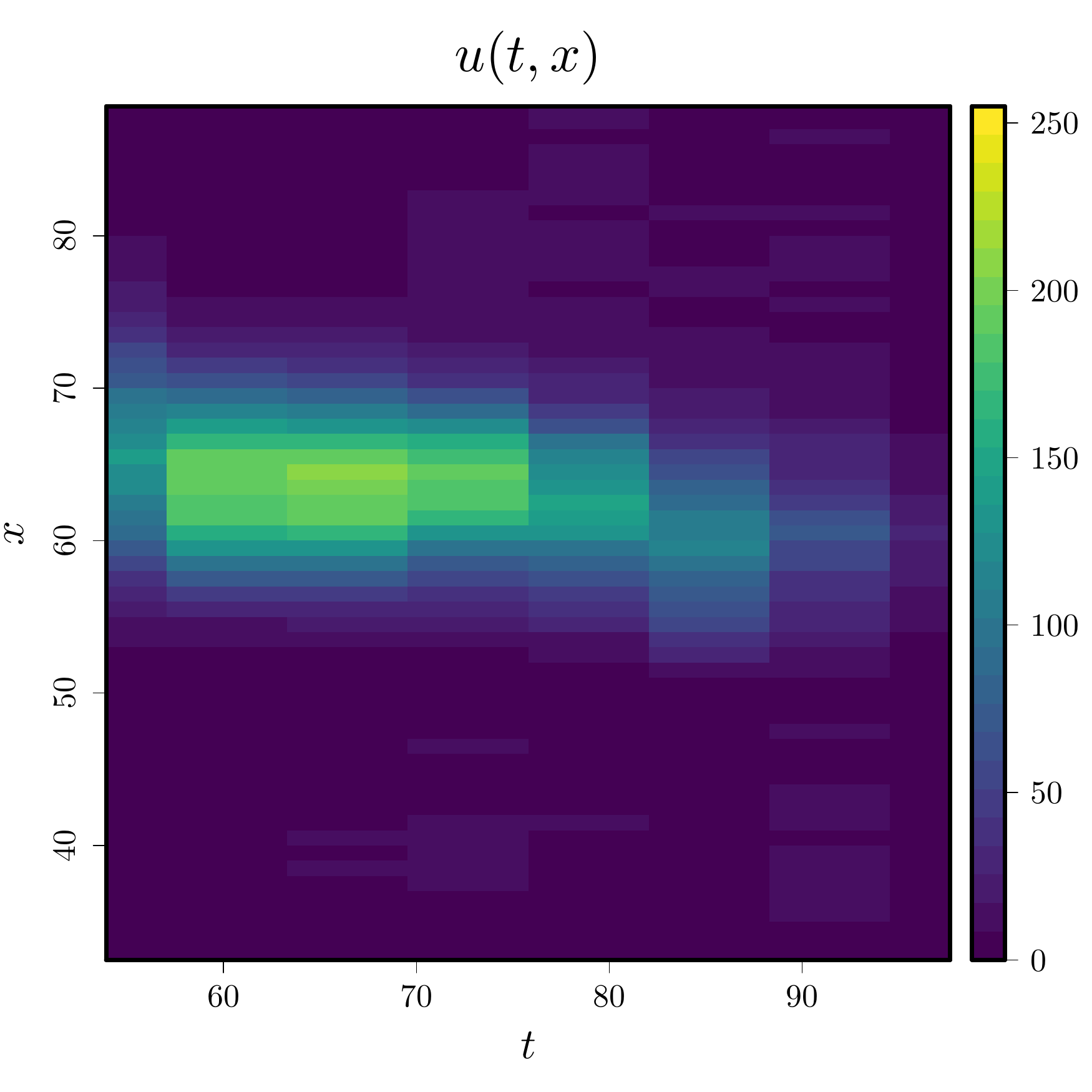}}
			\subfigure{\includegraphics[width=0.32\columnwidth]{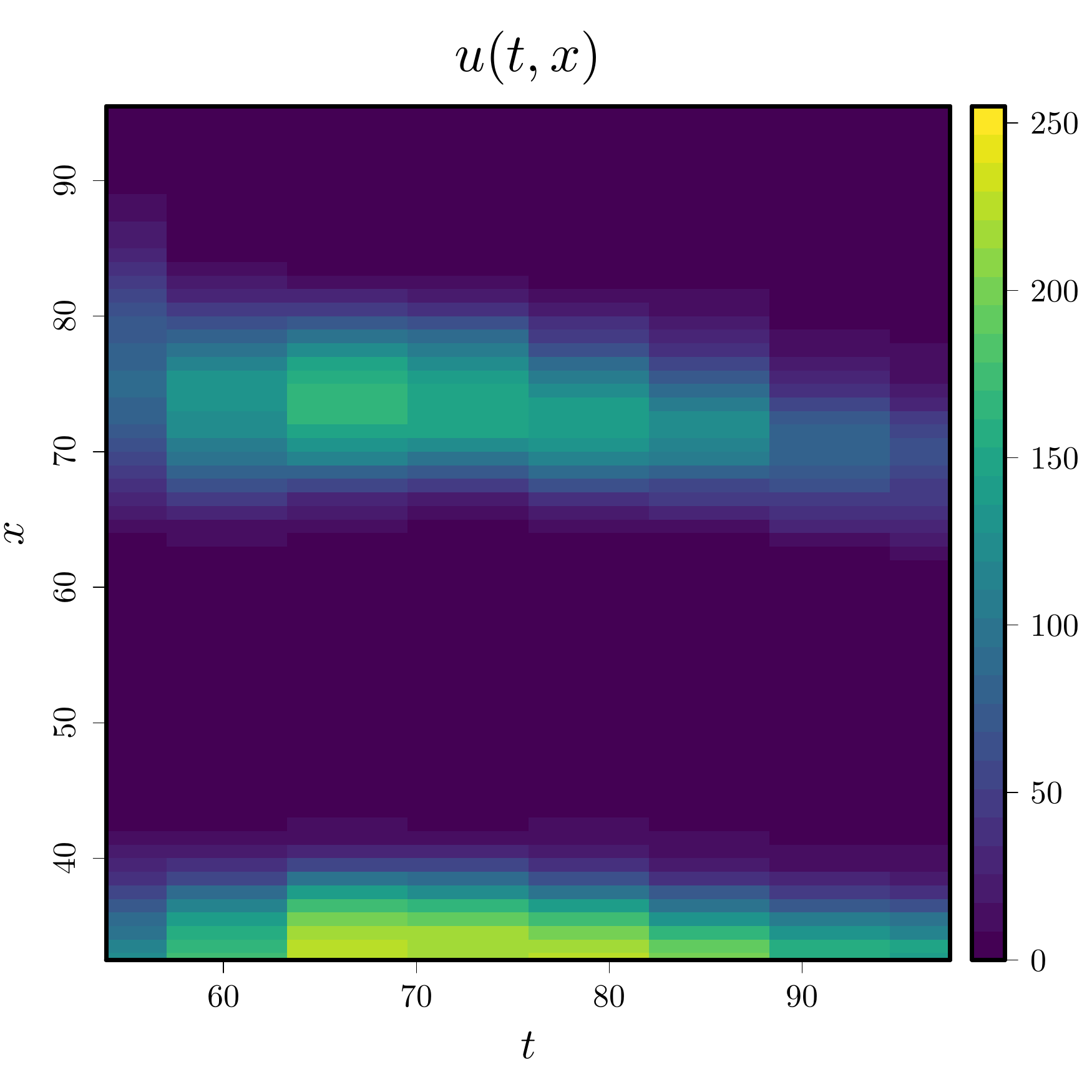}}
						
			\rotv{\hskip 12ex Gap Proteins}					
			\subfigure{\includegraphics[width=0.32\columnwidth]{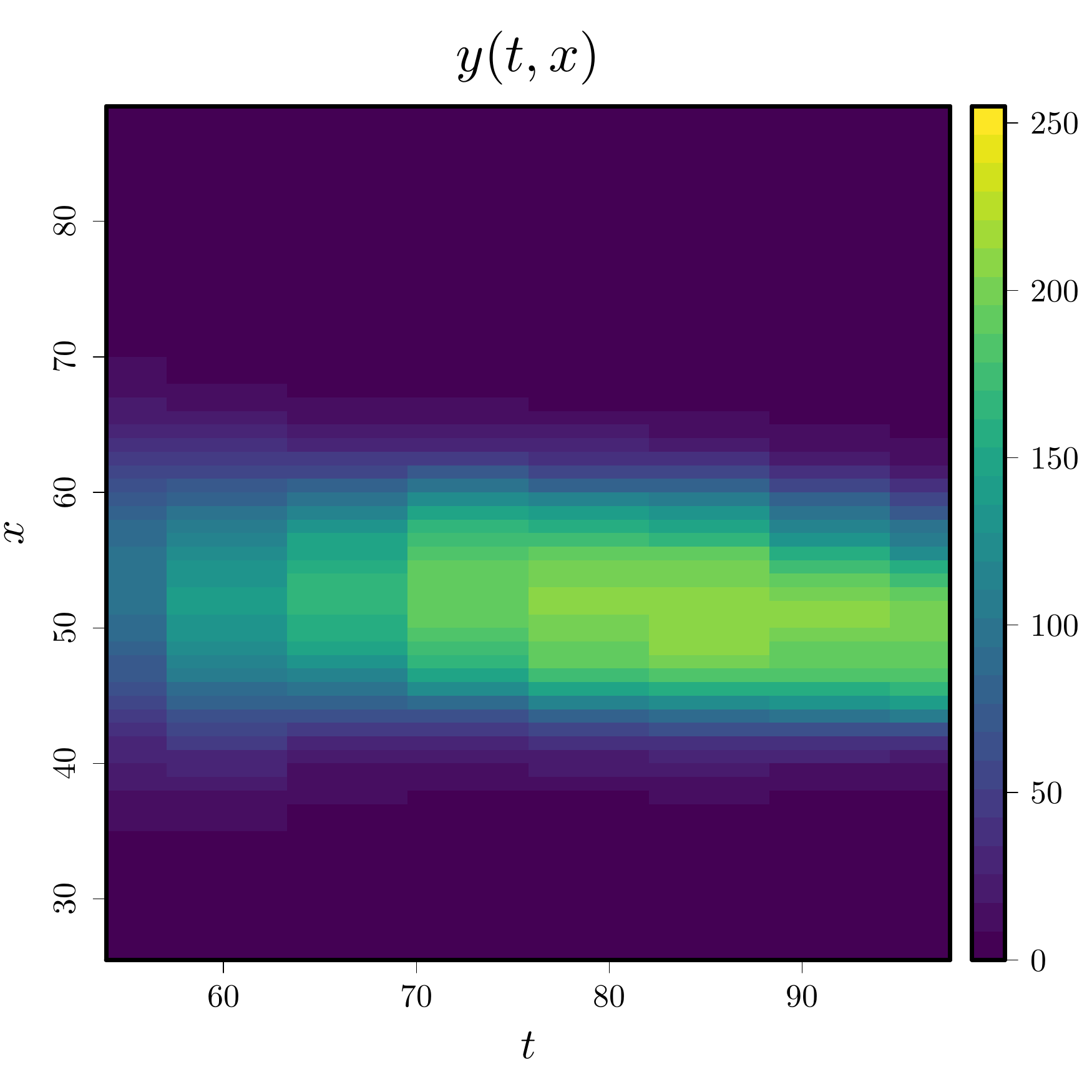}}
			\subfigure{\includegraphics[width=0.32\columnwidth]{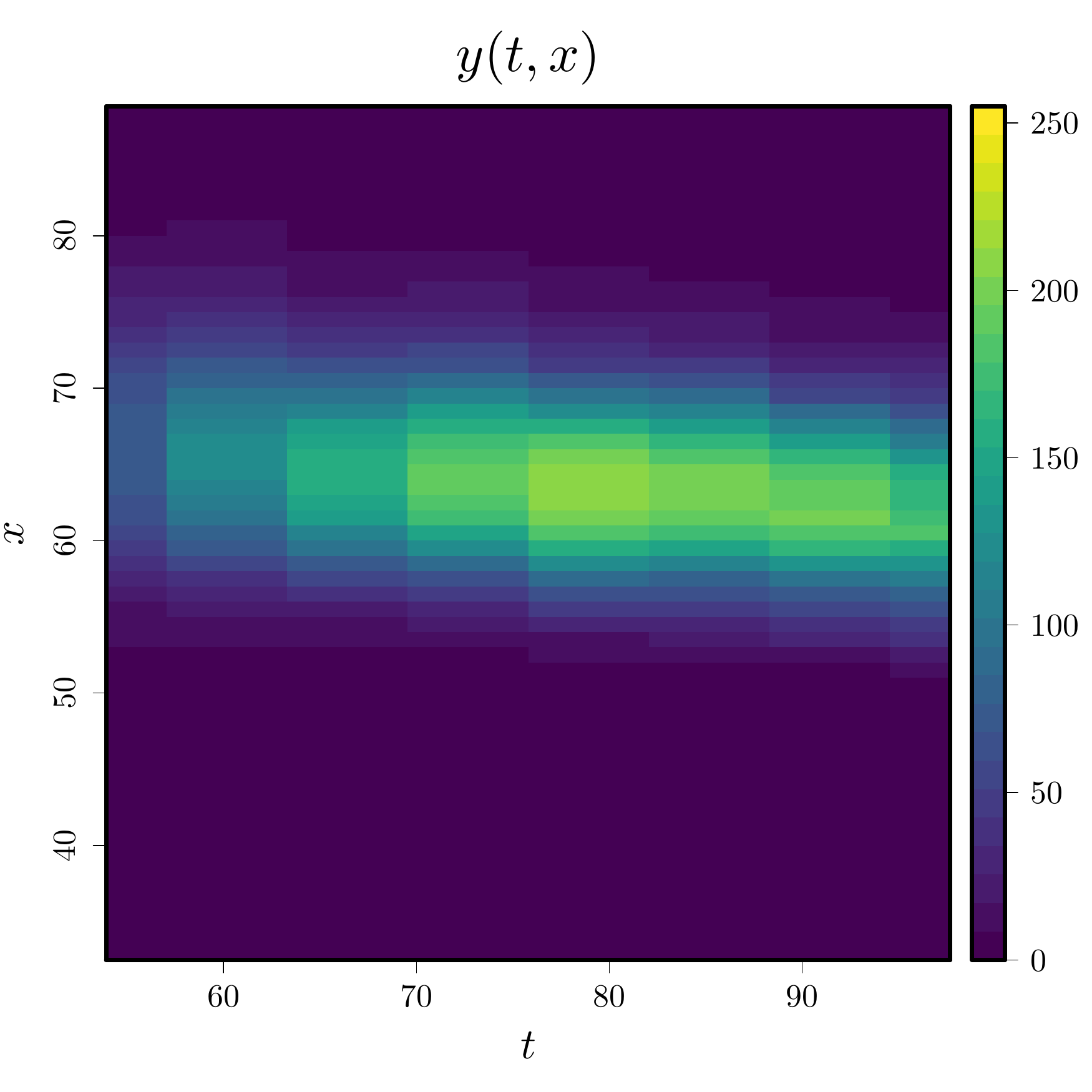}}
			\subfigure{\includegraphics[width=0.32\columnwidth]{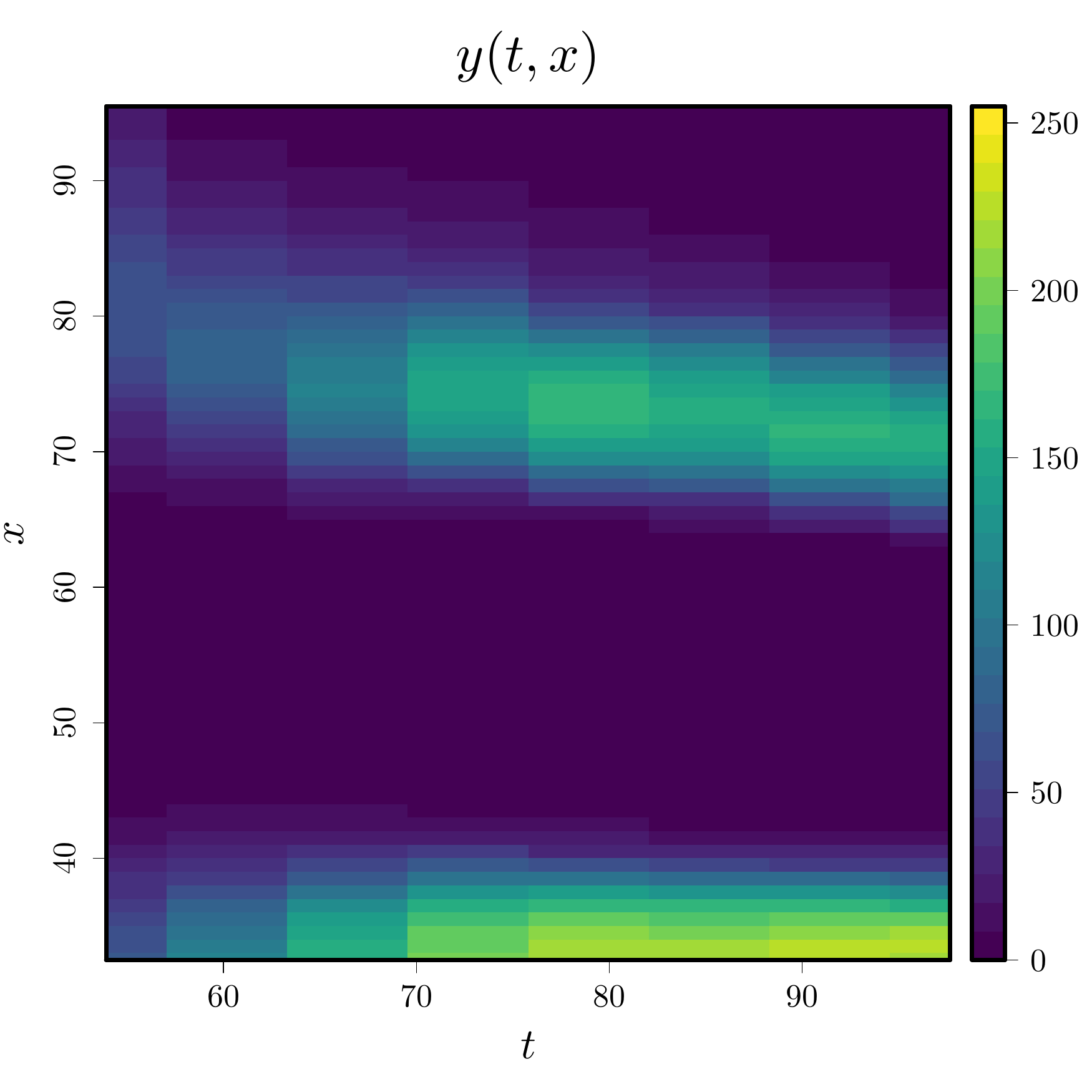}}
			\caption{Gap gene mRNA expression data from \citep{Becker2013DrosMel} for \textit{Kr\"uppel} (left), \textit{knirps} (centre), and \textit{giant} (right) trunk gap genes.
			}
			\label{fig:BeckerData}
		\end{figure}

		Figure \ref{fig:BeckerData} shows both mRNA and gap protein concentrations for each trunk gap gene of the dataset. We observe that both profiles exhibit similar patterns for each trunk gap gene with an active synthesis of proteins in the A-P intervals between 50-80\%, except for the trunk gap gene $gt$ where there is also synthesis between 30-40\%. We can also note that the synthesis of the three gap proteins remains almost equal along the time axis. 
		
		\begin{table}
			\centering
			\caption{Estimated parameters by \citep{Becker2013DrosMel} via LSA-based global optimisation.}
			\label{tab:BeckerParams}
			\begin{tabular}{cccc}
				\toprule
				\multirow{2}{*}{Parameter} & \multicolumn{3}{c}{Trunk Gap Gene} \\
				& \textit{Kr\"uppel} ($kr$) &  \textit{knirps} ($kni$) & \textit{giant} ($gt$)\\
				\midrule
				Translation rate ($S$) & 0.0970 & 0.0783 & 0.1107 \\
				Decay rate ($\lambda$) & 0.0764 & 0.0770 & 0.1110 \\ 
				Diffusion rate ($D$) & 0.0015 & 0.0125 & 0.0159 \\
				\bottomrule
			\end{tabular}
		\end{table}
		\begin{table}[t!]
			\scriptsize
			\centering
			\caption{Prediction assessment of the physically-inspired GP models using the dataset from \citep{Becker2013DrosMel}. Predictive accuracies of ten repetitions with different training sets are evaluated using the $Q^2$ (left) and $\operatorname{CA}_{\pm\sigma}$ (right) criteria. The mean $\mu$ and the standard deviations $\sigma$ of the results are shown using GP-mRNA and GP-Protein models for: \textit{Kr\"uppel} ($kr$), \textit{knirps} ($kni$) and \textit{giant} ($gt$). Best $Q^2$ results for each trunk gap gene (rows) are shown in bold.}
			\label{tab:BeckerDataQ2}
			\begin{minipage}{0.49\textwidth}
				\begin{tabular}{ccccc}
					\multicolumn{5}{c}{$Q^2 [\%]$ Results} \\
					\toprule
					Trunk & \multicolumn{2}{c}{GP-mRNA} & \multicolumn{2}{c}{GP-Protein} \\
					Gap & mRNA & Gap Protein & mRNA & Gap Protein \\
					Gene & $\mu \pm \sigma$ & $\mu \pm \sigma$ & $\mu \pm \sigma$ & $\mu \pm \sigma$ \\
					\midrule
					\multicolumn{5}{c}{Training data only from the gap protein concentration} \\
					\midrule		
					$kr$ & \boldsymbol{$90.5 \pm 2.0$} & \boldsymbol{$90.8 \pm 0.6$} & \boldsymbol{$92.0 \pm 0.6$} & \boldsymbol{$90.6 \pm 0.5$} \\
					$kni$& \boldsymbol{$81.1 \pm 2.5$} & \boldsymbol{$88.7 \pm 0.8$} & $77.6 \pm 4.7$ & $88.6 \pm 0.7$ \\
					$gt$ & $91.2 \pm 1.9$ & $92.3 \pm 0.6$ & \boldsymbol{$93.2 \pm 1.3$} & \boldsymbol{$92.8 \pm 0.5$} \\
					\midrule
					\multicolumn{5}{c}{Training data only from the mRNA concentration} \\
					\midrule		
					$kr$ & \boldsymbol{$86.7 \pm 1.4$} & \boldsymbol{$97.5 \pm 0.7$} & $84.0 \pm 2.1$ & $60.6 \pm  1.2$ \\
					$kni$& \boldsymbol{$82.9 \pm 2.1$} & \boldsymbol{$86.7 \pm 1.3$} & $80.7 \pm 3.2$ & $55.2 \pm 12.7$ \\
					$gt$ & \boldsymbol{$91.2 \pm 0.7$} & \boldsymbol{$93.9 \pm 0.3$} & $88.2 \pm 3.1$ & $84.3 \pm 1.7$ \\		
					\midrule
					\multicolumn{5}{c}{Training data from both biological quantities} \\
					\midrule		
					$kr$ & $96.8 \pm 0.5$ & $97.9 \pm 0.3$ & \boldsymbol{$98.6 \pm 0.6$} & \boldsymbol{$99.6 \pm 0.2$} \\
					$kni$& $91.2 \pm 2.9$ & $95.0 \pm 0.7$ & \boldsymbol{$94.5 \pm 3.5$} & \boldsymbol{$99.4 \pm 0.3$} \\
					$gt$ & $95.2 \pm 1.4$ & $96.2 \pm 0.6$ & \boldsymbol{$97.7 \pm 1.7$} & \boldsymbol{$99.3 \pm 0.2$} \\
					\bottomrule
				\end{tabular}
			\end{minipage}
			\begin{minipage}{0.49\textwidth}
				\hskip 3ex
				\begin{tabular}{ccccc}
					\multicolumn{5}{c}{$\operatorname{CA}_{\pm\sigma} [\%]$ Results} \\		
					\toprule
					Trunk & \multicolumn{2}{c}{GP-mRNA} & \multicolumn{2}{c}{GP-Protein} \\
					Gap & mRNA & Gap Protein & mRNA & Gap Protein \\
					Gene & $\mu \pm \sigma$ & $\mu \pm \sigma$ & $\mu \pm \sigma$ & $\mu \pm \sigma$ \\
					\midrule
					\multicolumn{5}{c}{Training data only from the gap protein concentration} \\
					\midrule		
					$kr$ & $71.7 \pm 6.7$ & $46.2 \pm 4.0$ & $69.2 \pm 1.3$ & $57.2 \pm 1.3$ \\
					$kni$& $83.0 \pm 5.3$ & $37.1 \pm 4.4$ & $49.7 \pm 6.3$ & $32.7 \pm 5.3$ \\
					$gt$ & $85.6 \pm 6.9$ & $31.8 \pm 2.7$ & $68.2 \pm 3.3$ & $42.2 \pm 1.8$ \\
					\midrule
					\multicolumn{5}{c}{Training data only from the mRNA concentration} \\
					\midrule		
					$kr$ & $57.3 \pm 1.5$ & $80.7 \pm 2.8$ & $58.2 \pm 2.3$ & $78.0 \pm 1.3$ \\
					$kni$& $48.4 \pm 5.4$ & $64.7 \pm 4.2$ & $54.9 \pm 7.8$ & $74.3 \pm 3.5$ \\
					$gt$ & $40.0 \pm 3.1$ & $63.0 \pm 2.4$ & $53.1 \pm 4.2$ & $66.8 \pm 1.4$ \\
					\midrule
					\multicolumn{5}{c}{Training data from both biological quantities} \\
					\midrule		
					$kr$ & $34.5 \pm 2.8$ & $21.9 \pm 4.6$ & $81.9 \pm 2.8$ & $87.7 \pm 2.0$ \\
					$kni$& $20.5 \pm 2.4$ & $13.7 \pm 2.2$ & $74.4 \pm 7.1$ & $86.6 \pm 1.9$ \\
					$gt$ & $19.6 \pm 4.5$ & $14.6 \pm 3.1$ & $79.8 \pm 5.6$ & $83.5 \pm 2.6$ \\
					\bottomrule
				\end{tabular}		
			\end{minipage}
		\end{table}
		
		\subsection{Inference results}
		\label{subsec:inference}
		For each numerical experiment, we randomly selected the 30\% of the available data from both biological profiles to train each of the GP model, and the remaining 70\% of the data is used to test the quality of the models. For the hyperparameter parameter estimation, the mechanistic parameters $(S, \lambda, D)$ were fixed to be equal to the ones estimated by \citep{Becker2013DrosMel} via Lam simulated annealing (LSA)-based global optimisation (see Table \ref{tab:BeckerParams}), and the covariance parameters $(\sigma^2, \theta_x, \theta_t)$ are estimated via maximum likelihood. We repeat this procedure for ten different random training sets. For the GP-mRNA model, we choose the number of terms of the Green's function according to two criteria: the quality of the predictions and the computational cost. We gradually increased the number of terms, starting with the first five terms, and we checked the quality of the resulting model in terms of the $Q^2$ and $\operatorname{CA}_{\pm\sigma}$ criteria. We observed that the results became stable and accurate after the first twenty terms. Finally, we tested the same three conditions of data availability discussed in Sections \ref{subsubsec:hGPmRNAtoy} and \ref{subsubsec:hGPGapGenetoy}.
		
		Table \ref{tab:BeckerDataQ2} shows the performance of both physically-inspired GP models for ten repetitions using different training sets. The mean $\mu$ and the standard deviations $\sigma$ of the $Q^2$ and $\operatorname{CA}_{\pm\sigma}$ results are shown. One can note that when only mRNA or gap protein data were used to train the models, better $Q^2$ values were commonly obtained when the GP prior was placed over the process where data were available. Although both models yielded similar departures of $\operatorname{CA}_{\pm\sigma}$ percentages. This result agrees with the log-likelihood performances of both GP models (see Table \ref{tab:loglikePerformance}). This comparison is valid since the biological parameters $(S, \lambda, D)$ were assumed to be known. However, these parameters are commonly unknown, and they have to be estimated in real applications. Since $(S, \lambda, D)$ are not encoded in the covariance function of the GP prior, they cannot be learned if training data are available only from the prior. In this sense, in real applications, GP priors should be placed over the unobserved processes as suggested in \citep{Lawrence2006modelling,Alvarez2011LLFM}.
		
		Another interesting result can be pointed out by the performance of GP-mRNA model. This model was previously studied in \citep{Alvarez2011LLFM,Vasquez2014LFMDrosMel} for the inference of mRNA using gap protein data only. However, it could not be further tested due to the lack of mRNA data, and inference results were justified according to qualitative criteria. Here, one can observe from Table \ref{tab:BeckerDataQ2} that GP-mRNA yielded accurate quantitative results, with $Q^2$ values over the 80\%, on both biological quantities independently on the training data availability. However, we must note that the GP-mRNA model led to costly procedures due to the evaluation of more expensive kernel structures (e.g. depending on the number of terms from the Green's function, and the computation of the $\operatorname{erf}$ and Faddeeva functions). More precisely, while the parameter estimation using the GP-Protein model takes a couple of minutes, the running time for GP-mRNA is in the order of hours.
		\begin{table}
			\centering
			\caption{Log-likelihood performance of GP-mRNA and GP-Protein for one repetition. 
				First and second best results are shown in bold and grey.}
			\label{tab:loglikePerformance}
			\begin{tabular}{ccccc}
				\toprule
				\multirow{2}{*}{Model} & Training & \textit{Kr\"uppel} & \textit{knirps} & \textit{giant} \\
				& Data Usage & ($kr$) & ($kni$) & ($gt$) \\
				\midrule
				\multirow{2}{*}{GP} & Protein & -53518.6 & -50542.3 & -42692.8 \\ 
				\multirow{2}{*}{mRNA} & mRNA & \color{gray} \textbf{-53502.0} & \color{gray} \textbf{-50537.0} & \color{gray} \textbf{-42654.9} \\ 
				& Protein \& mRNA & \textbf{-19603.9} & \textbf{-36868.5} & \textbf{-31246.4} \\ 
				\midrule
				\multirow{2}{*}{GP} & Protein & \color{gray} \textbf{-53502.0} & \color{gray} \textbf{-50537.0} & \color{gray} \textbf{-42654.9} \\ 
				\multirow{2}{*}{Protein} & mRNA & -53696.7 & -50742.9 & -42905.7 \\ 
				& Protein \& mRNA & \textbf{-1238.7}  & \textbf{-1181.0}  & \textbf{-1269.7}  \\ 
				\bottomrule
			\end{tabular}
		\end{table}
		
		Finally, when both mRNA and gap protein concentration data are used to train the models, we can note that the GP-Protein model outperformed the results provided by GP-mRNA with $Q^2$ improvements around 3-5\% in all the cases. Here, the hyperparameters $\Btheta = (S, \lambda, D, \sigma^2, \theta_x, \theta_t)$ were estimated via maximum likelihood with an initial set of biological parameters $(S, \lambda, D)$ given by Table \ref{tab:BeckerParams}. The choice of using the estimated values of Table \ref{tab:BeckerParams} as starting point is due to, according to numerical experiments, it seems that $(S, \lambda, D)$ cannot be estimated consistently. As some covariance parameters from certain GP models cannot be estimated consistently due to their non-microergodicity \citep{stein1999interpolation,Zhang2004InconsistentEst}, we believe that both GP-mRNA and GP-Protein models suffer from the same downside. Finally, after convergence of the maximum likelihood estimation, we observed that the estimated values of $(S, \lambda, D)$ remained around the ones from Table \ref{tab:BeckerParams}.
		
		According to the $\operatorname{CA}_{\pm\sigma}$ results, one can observe that the GP-Protein model provide a more reasonable predictive variances than GP-mRNA. The harsh underestimation of the predictive intervals by GP-mRNA is produced by numerical instabilities in the computation of their covariance matrices and the gradients of the joint process. In practice, one possible solution to avoid this overfitting is the early stopping of the maximum likelihood optimisation. In terms of the likelihood performance, we also noticed that results using GP-Protein are of a lesser order of magnitude to those obtained by GP-mRNA (see Table \ref{tab:loglikePerformance}). This suggests that the GP-Protein model better describes the behaviour of the three trunk gap genes. Since homogeneous conditions are enforced in GP-mRNA, we believe those constraints may affect the likelihood performance of the model.
		\begin{figure*}[t!]
			\centering
			\hskip 2ex \textit{Kr\"uppel} ($kr$) \hskip 15ex \textit{knirps} ($kni$) \hskip 17ex \textit{giant} ($gt$)\\
			
			\rotv{\hskip 4ex  GP-mRNA -- mRNA} \hskip 0.8ex
			\subfigure{\includegraphics[width=0.245\textwidth]{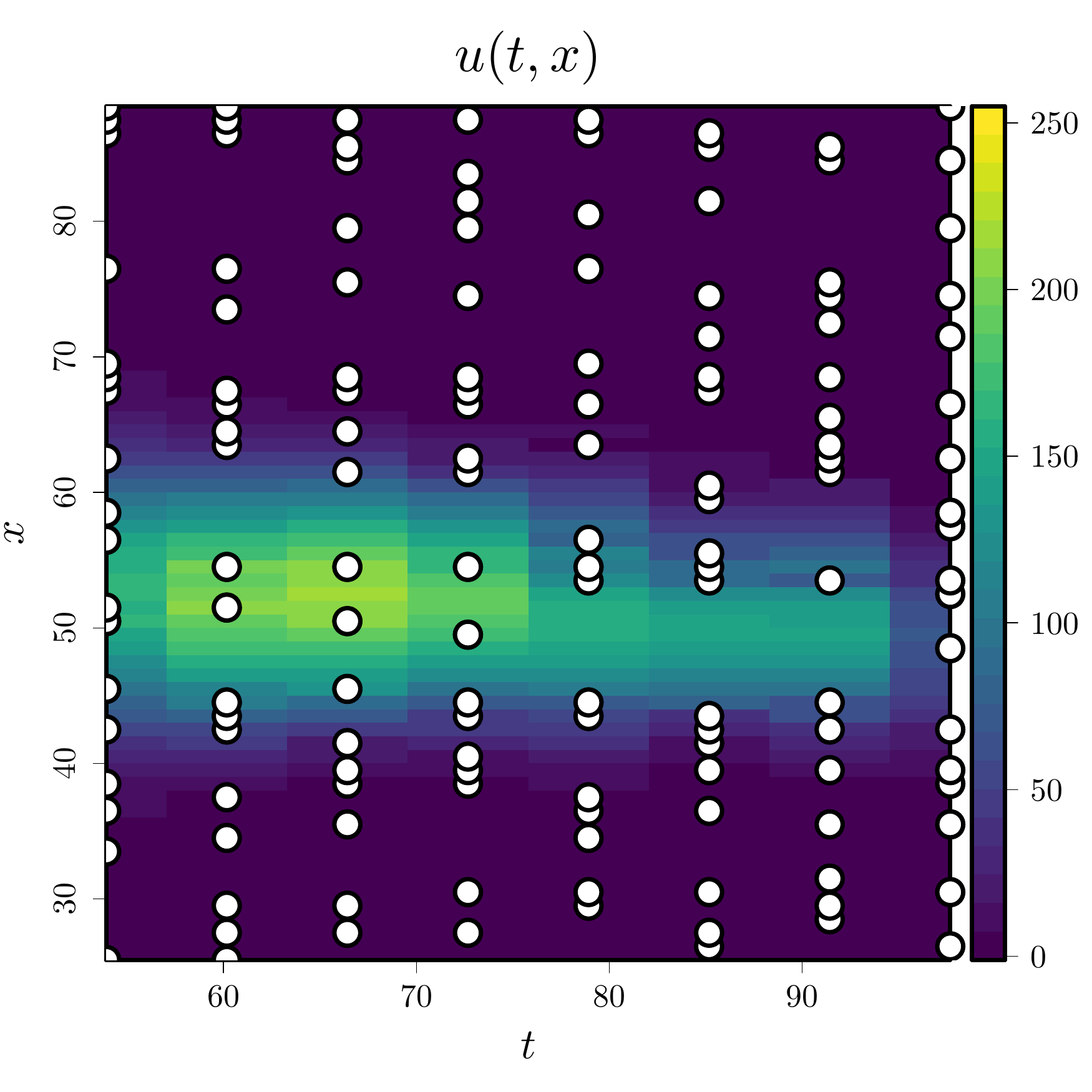}}
			\subfigure{\includegraphics[width=0.245\textwidth]{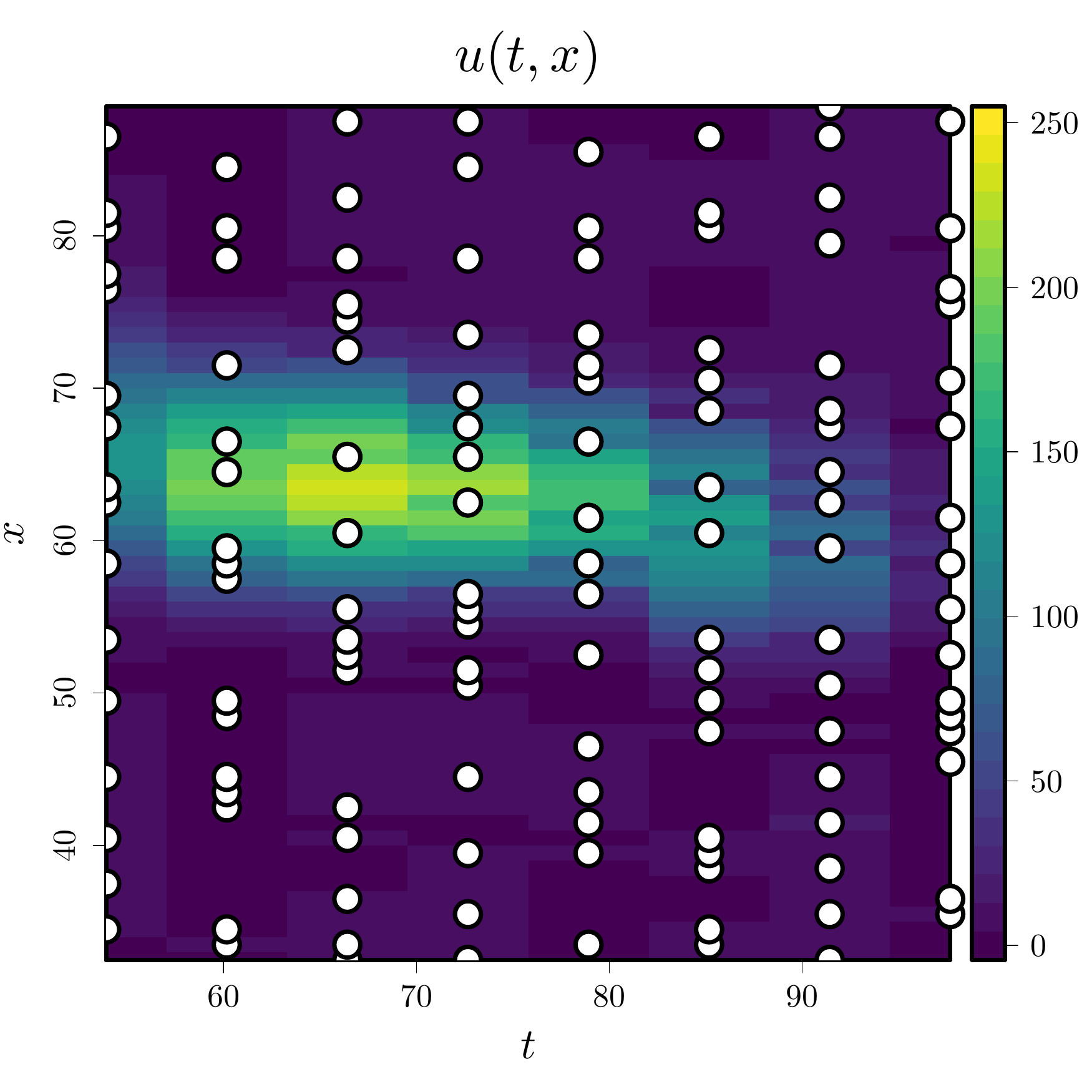}}
			\subfigure{\includegraphics[width=0.245\textwidth]{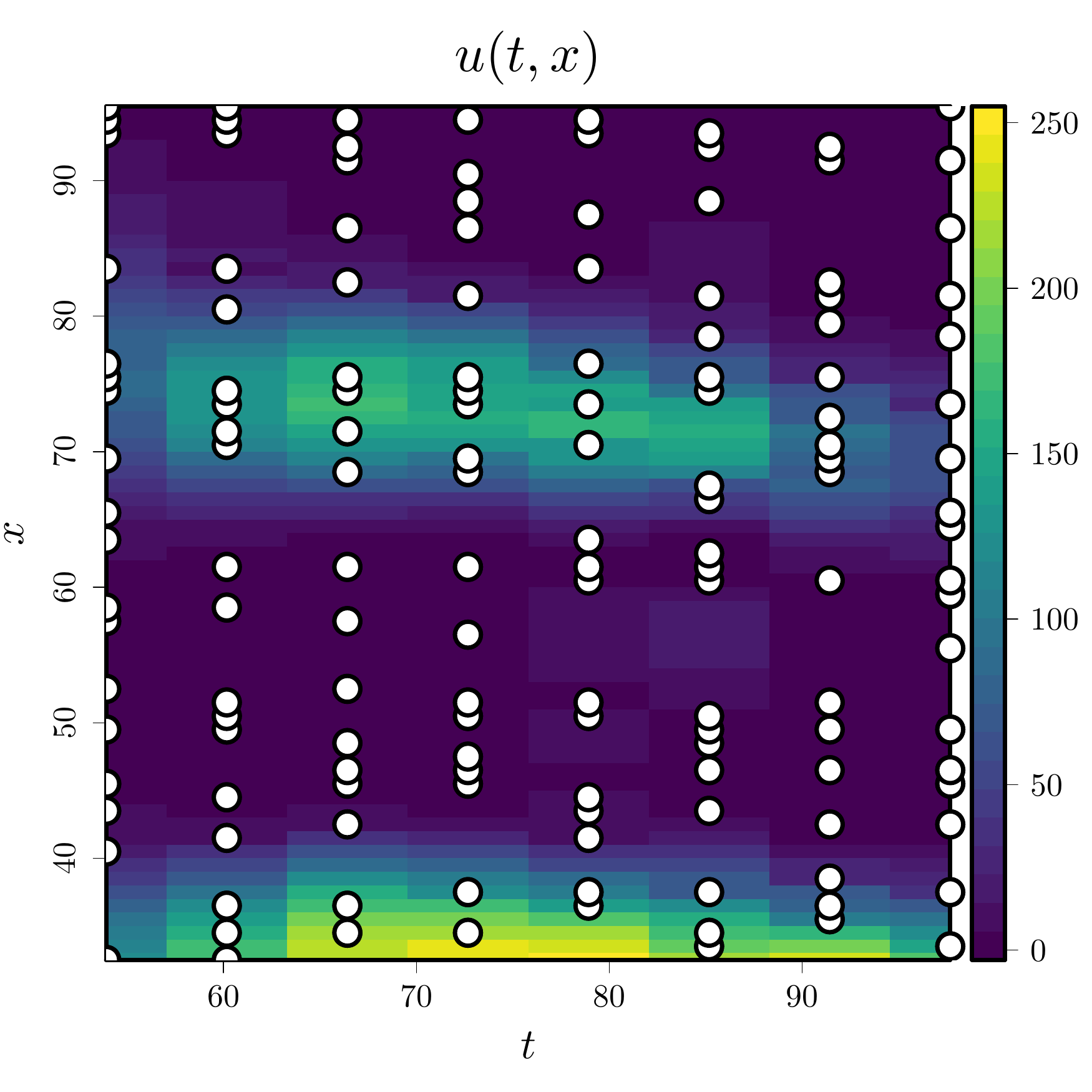}}
			
			\rotv{\hskip 2ex GP-mRNA -- Gap Protein} \hskip 0.8ex
			\subfigure{\includegraphics[width=0.245\textwidth]{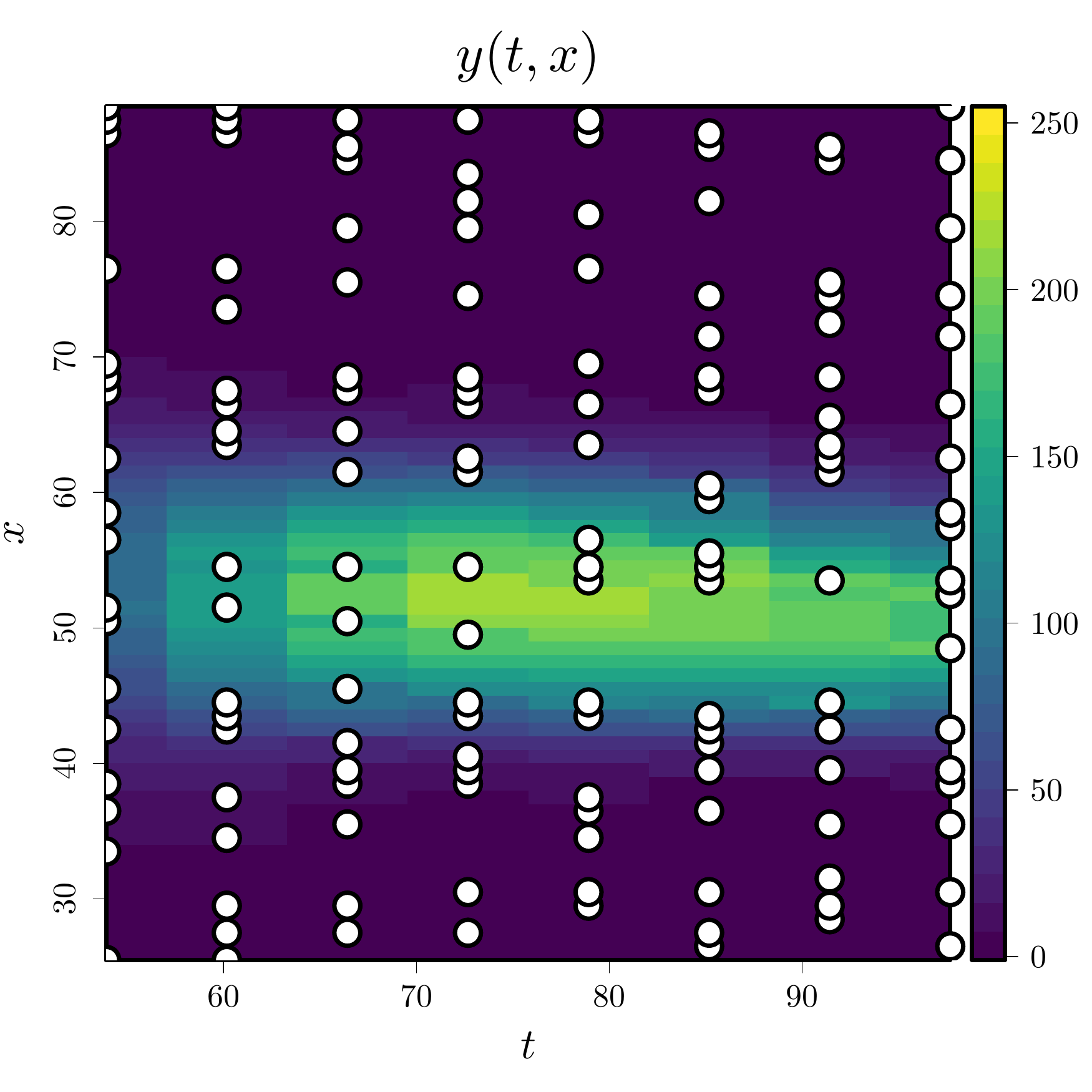}}
			\subfigure{\includegraphics[width=0.245\textwidth]{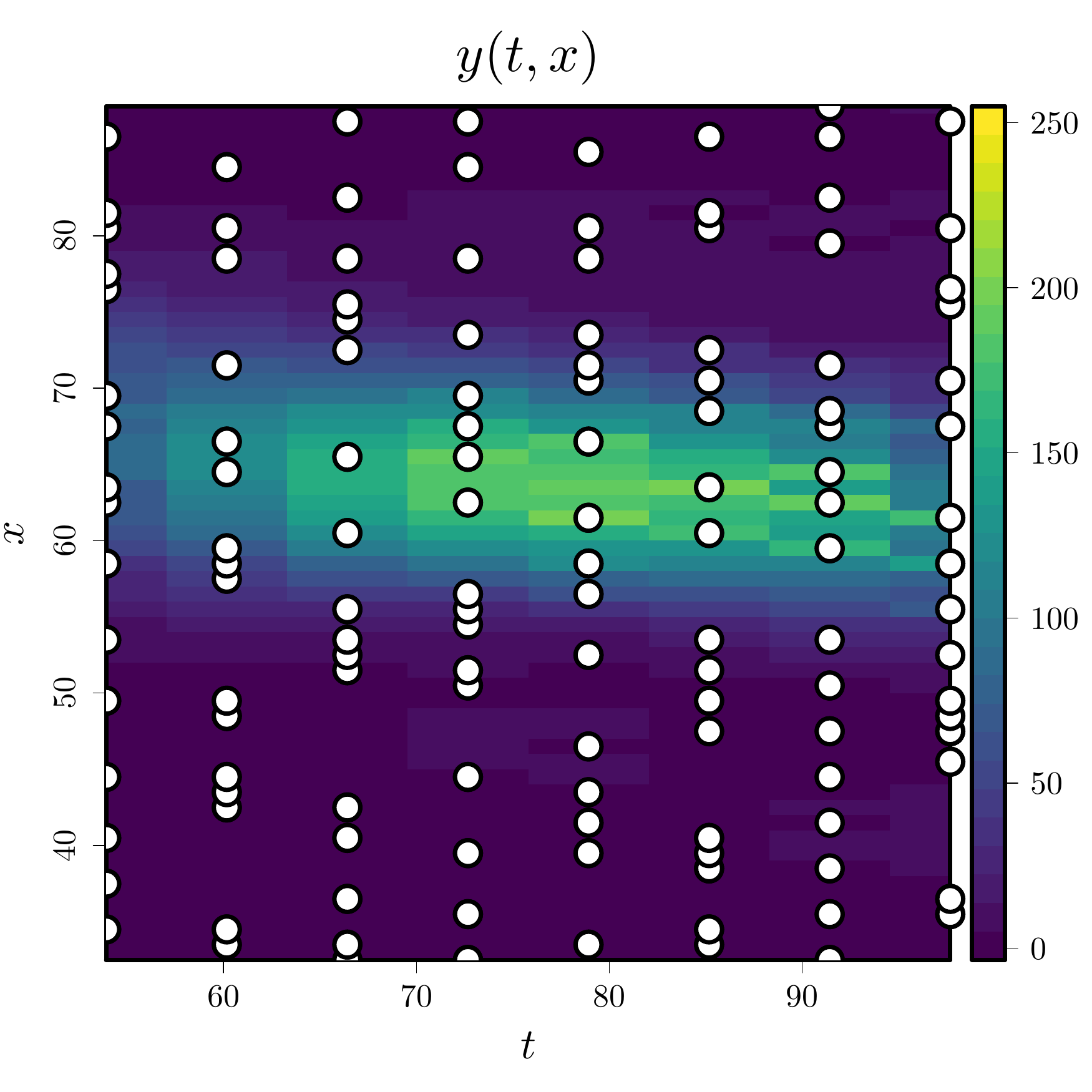}}
			\subfigure{\includegraphics[width=0.245\textwidth]{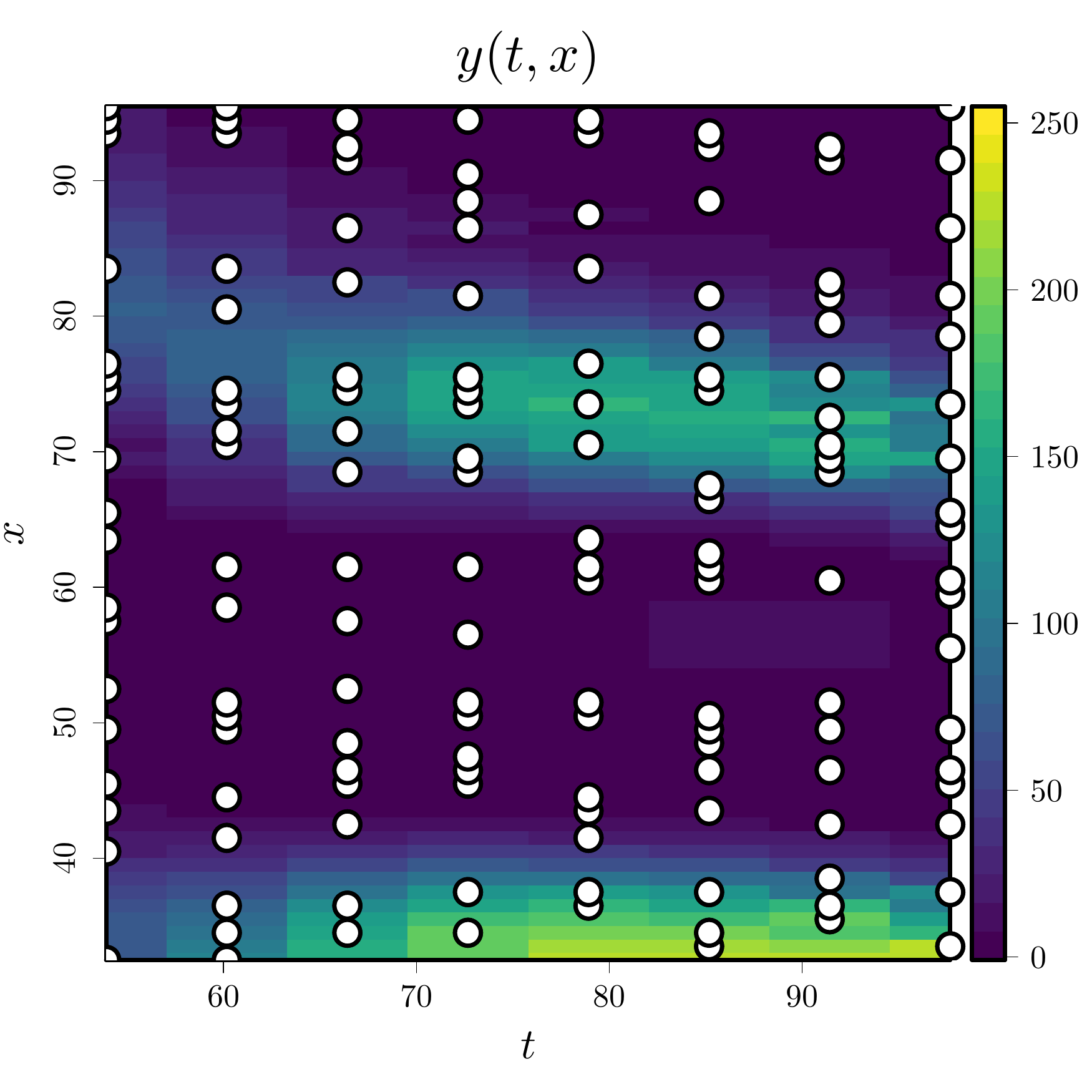}}
			
			\rotv{\hskip 4ex GP-Protein -- mRNA}  \hskip 0.8ex	
			\subfigure{\includegraphics[width=0.245\textwidth]{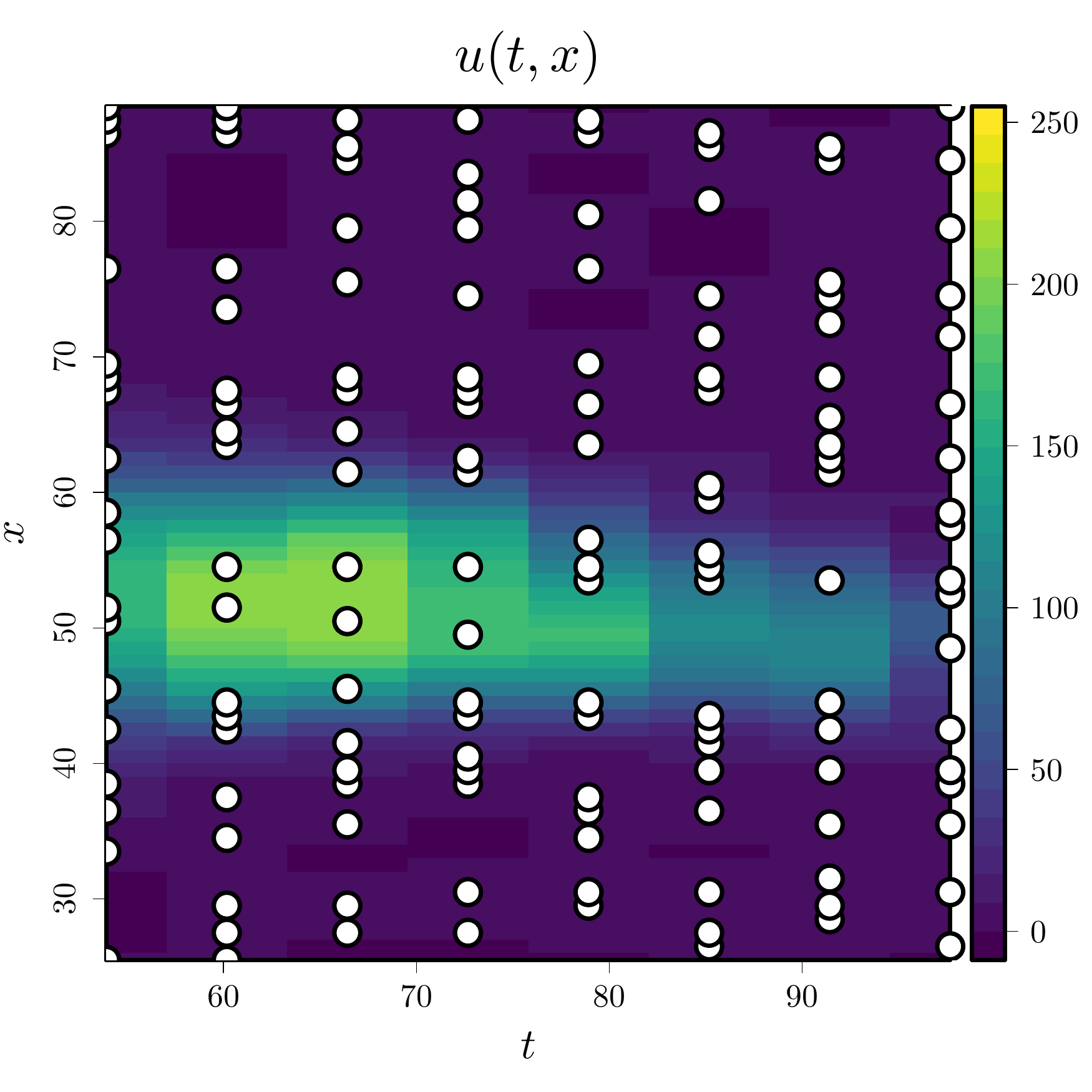}}
			\subfigure{\includegraphics[width=0.245\textwidth]{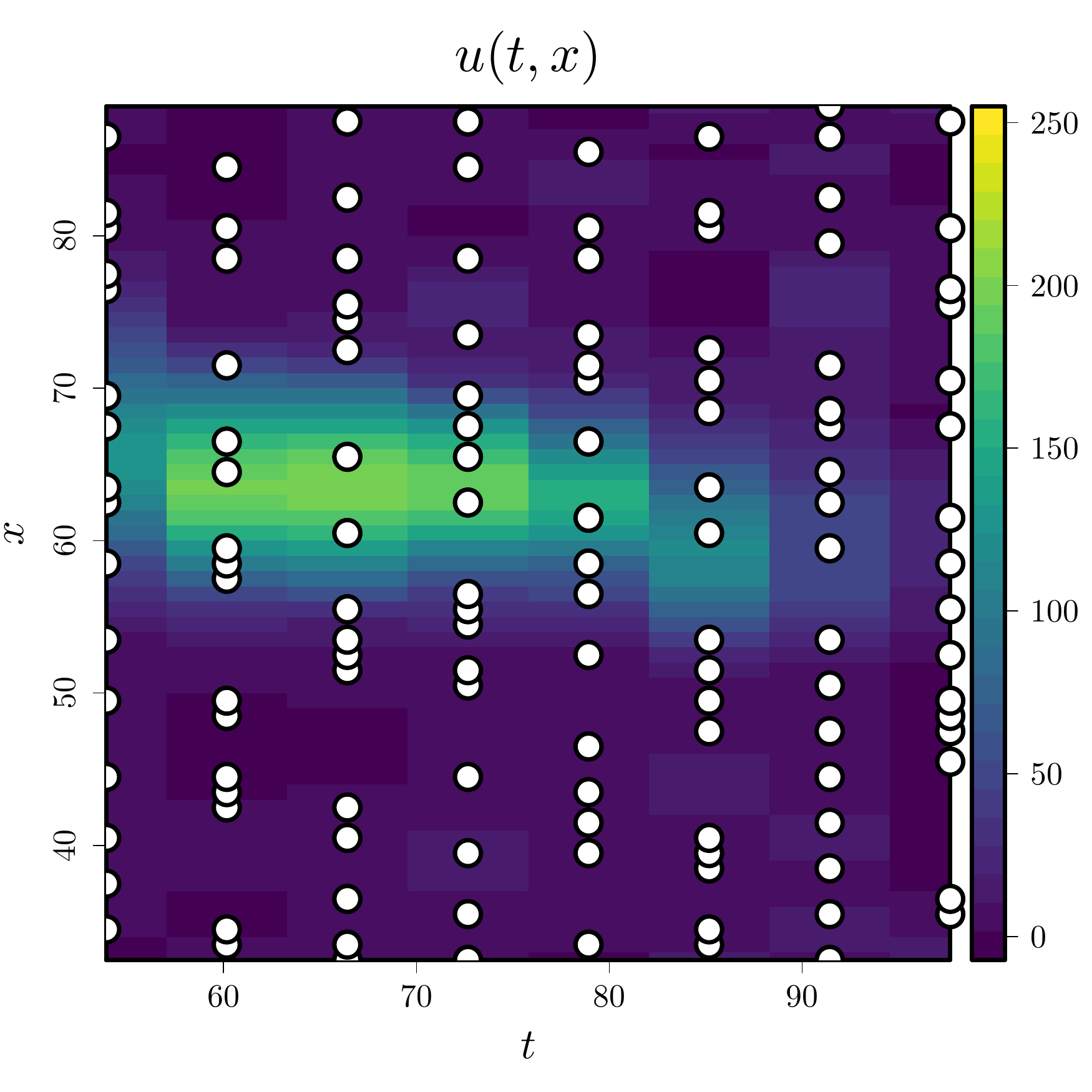}}
			\subfigure{\includegraphics[width=0.245\textwidth]{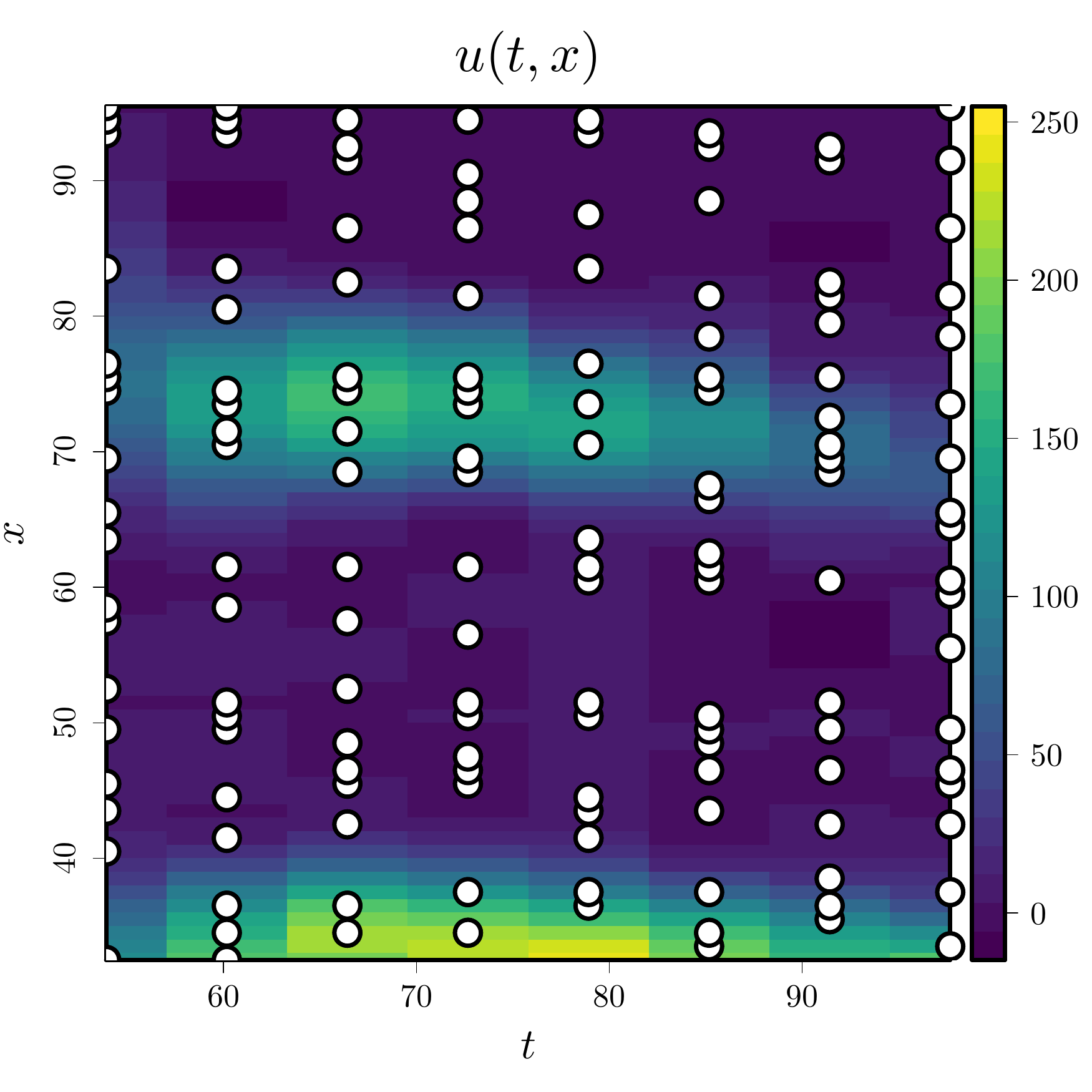}}
			
			\rotv{\hskip 2ex GP-Protein -- Gap Protein} \hskip 0.8ex	
			\subfigure{\includegraphics[width=0.245\textwidth]{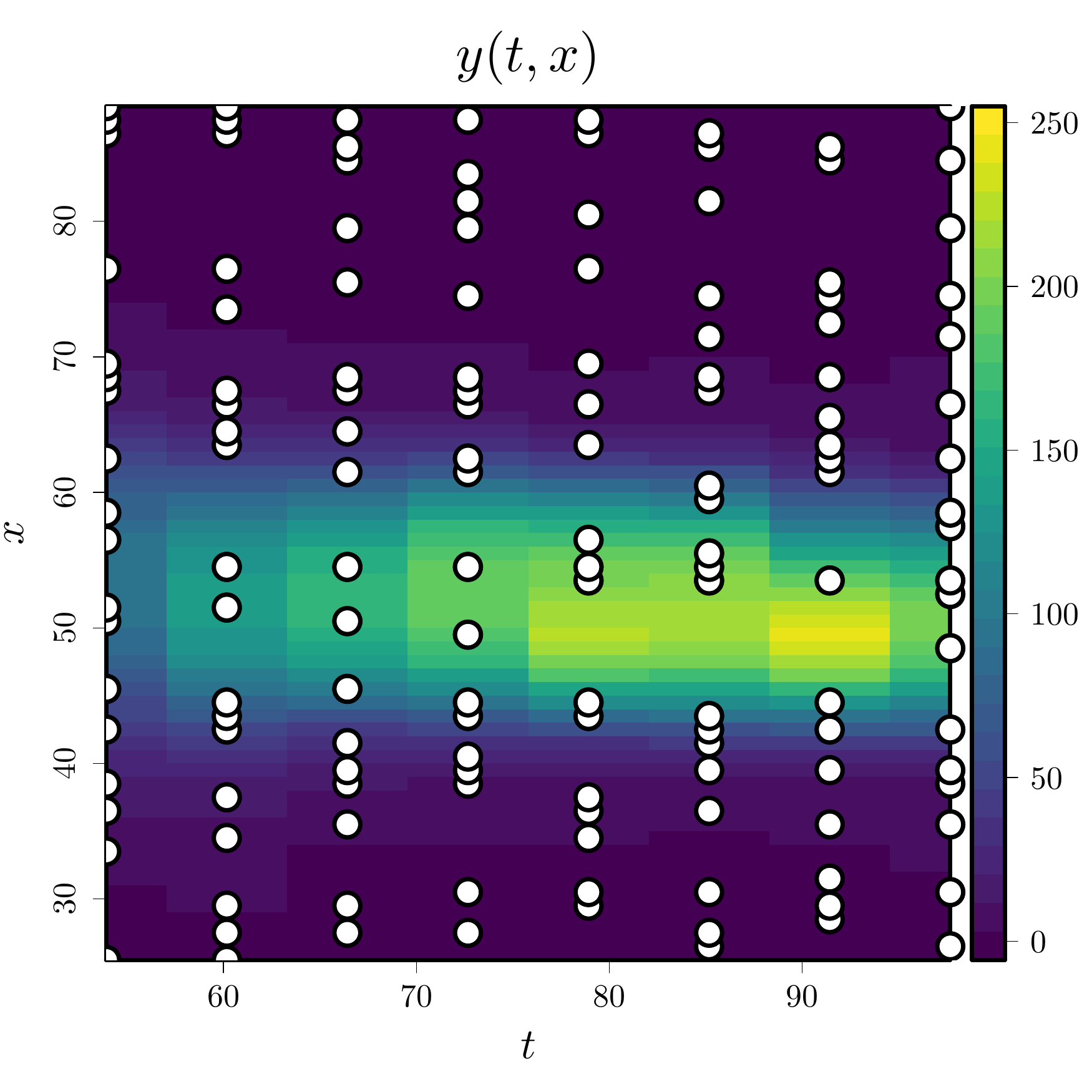}}	
			\subfigure{\includegraphics[width=0.245\textwidth]{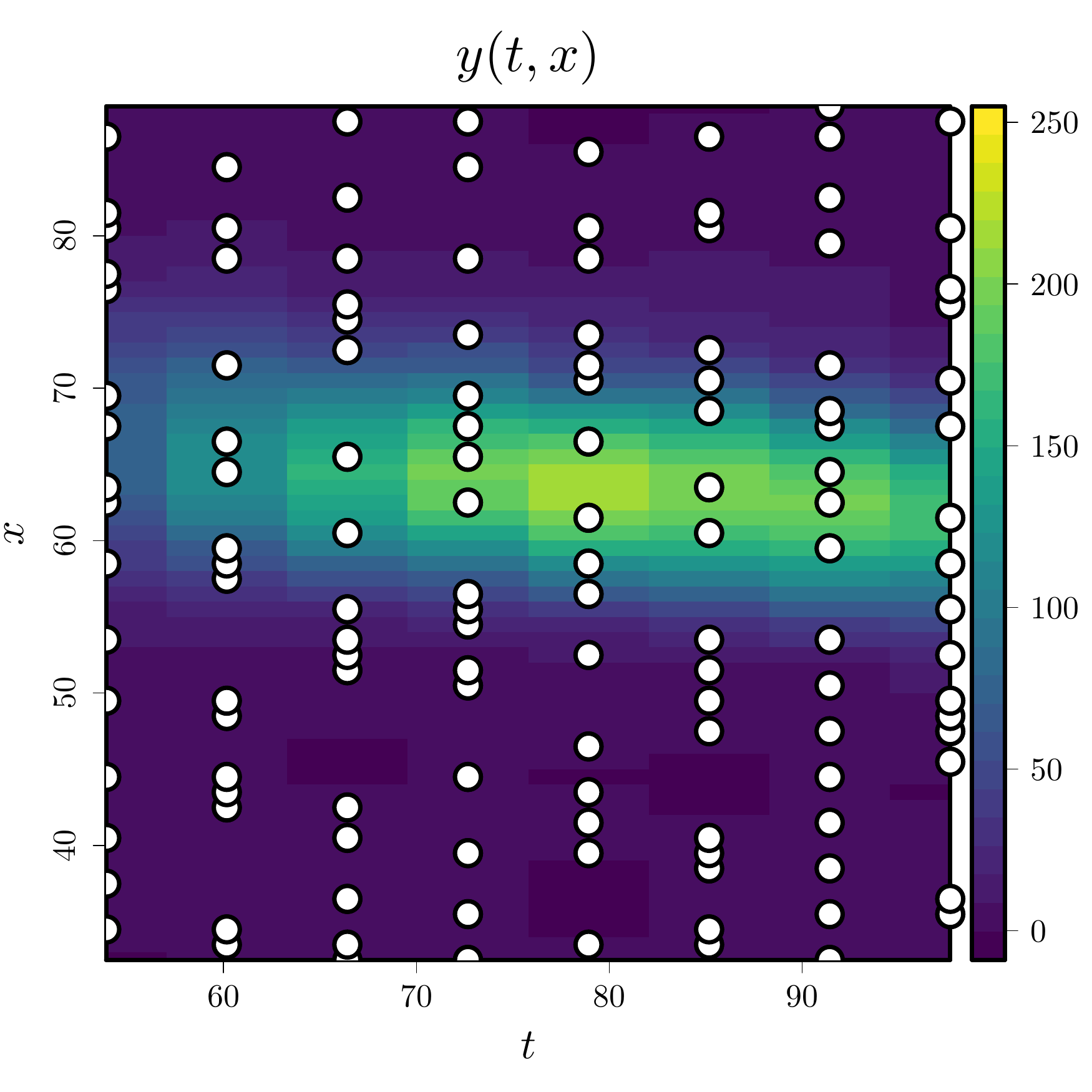}}
			\subfigure{\includegraphics[width=0.245\textwidth]{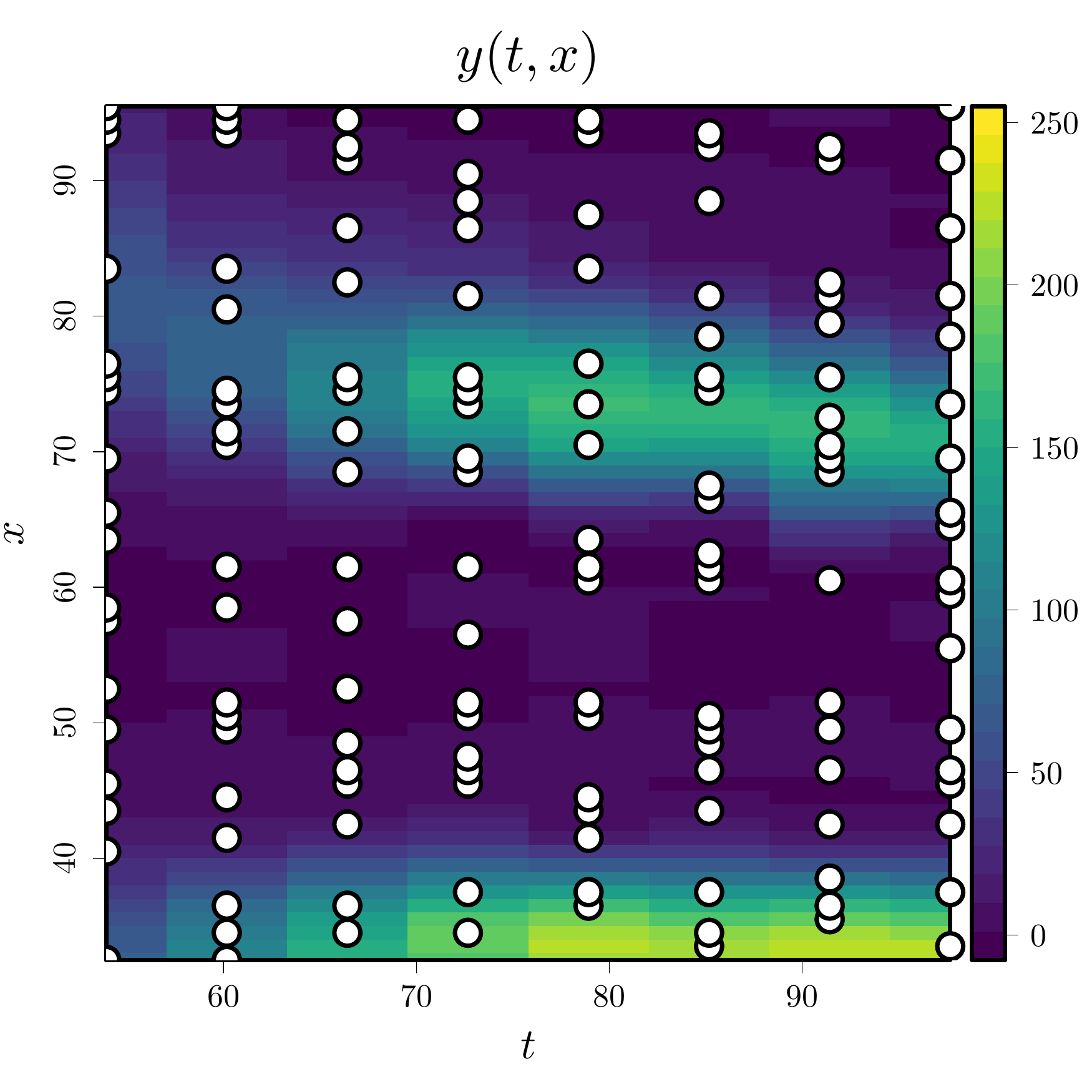}}
			
			\caption{Prediction results on the three trunk gap proteins $kr$, $kni$ and $gt$ (columns) using GP-mRNA (first and third row) and GP-Protein (second and fourth row). Training points (white dots) correspond to the 30\% of the dataset.}
			\label{fig:hGPBeckerInfForward}
		\end{figure*}
		
		Figure \ref{fig:hGPBeckerInfForward} shows the obtained predictive means for one of the repetitions of the three trunk gap genes for both the GP-mRNA and GP-Protein models. One can observe that both models are able to capture the correlation between the mRNA and gap protein concentrations. In particular, we note that both models are capable of precisely recovering the time lag between the peaks in the mRNAs and the ones in the proteins for the three trunk gap genes: about 15-20 min. We also observe that the GP-Protein model provides smoother profiles than the ones obtained by GP-mRNA.
		
		As pointed out in Section \ref{sec:hGP}, one must note that both GP-mRNA and GP-Protein models do not necessarily guarantee that the mRNA and gap proteins are strictly positive quantities. In practice, positiveness assumptions can be fulfilled by positive non-linear transformations \citep[e.g. exponential of GPs][]{Vanhatalo2007SparseLogGPs}. However, those transformations do not yield analytical solutions of the resulting joint GP as we provided in Section \ref{sec:hGP}. Another possible approach to guarantee positiveness conditions is based on finite-dimensional approximations of GPs \citep{Maatouk2016GPineqconst,LopezLopera2017FiniteGPlinear}. This approach could be potentially investigated in future implementations.
		
		\section{Conclusion}
		\label{sec:conclusion}
		We have studied two types of physically-inspired Gaussian process (GP) regression approaches to model the post-transcriptional regulation of the early embryo of Drosophila. Both approaches are based on a continuous version of the linear reaction-diffusion differential equation. The main difference between both GP models lies on whether the GP prior is placed: either over the mRNA (GP-mRNA) or gap protein (GP-Protein). First, for the GP-mRNA model (framework known as \textit{latent force model}), previous studies have been restricted to the use of gap protein data due to the lack of mRNA data \citep{Alvarez2011LLFM,Vasquez2014LFMDrosMel}. In this paper, we tested it when the information from both mRNA and protein concentrations are available, and we analysed their performance under different situations depending on the availability of data. Second, we introduced the GP-Protein model as a novel alternative where the complexity of computations is reduced to the differentiation of kernels.
		
		We studied three conditions depending on the availability of data: whether from the mRNA, the gap protein or both quantities. We concluded that both models provide promising predictions when the number of training data is large enough. We also tested them in a real-world biological problem to model the early embryo of Drosophila. One interesting result we could pointed out from the GP-mRNA model is referred by its reliable inference results. We observed it yielded accurate results, with $Q^2$ values over the 80\%, independently of the training data availability. Finally, we also observed that GP-Protein model slightly outperformed the prediction results provided by GP-mRNA when training data from both quantities were used. 
		
		According to numerical experiments proposed in this paper, we have different recommendations depending on the data availability. First, we recommend the use of the GP-Protein model when data are available from both mRNA and gap protein concentrations. This proposition stands due to its numerical stability, computational cost and accurate performance. Second, if data are available from only one of the biological quantities, we suggest placing the GP prior assumption over the unobserved profile in order to obtain more accurate inference results and to be able to learn both the mechanistic and covariance parameters via maximum likelihood estimation. Finally, we may prefer one of the GP models over the other depending on the nature of the biological data (e.g. initial and boundary conditions, smoothness of observations).  As an example, when data exhibit homogeneous conditions (both initial and boundary conditions equal to zero), we may recommend the use of GP-mRNA rather than GP-Protein as these conditions are explicitly encoded in the structure of the GP-mRNA model.
		
		Both frameworks discussed in this paper could be improved in different ways. First, it is known that some regulatory processes exhibit delays during the translation step. We consider that taking into account those delays would be interesting as a future work \citep[see, e.g. ][]{Honkela2015GPBioDelays}. Second, the GP-Protein model has been introduced for single-input single-output schemes. Other possible potential future work is to derive the GP-Protein framework with multiple mRNA profiles (driving-forces) and multiple gap proteins (outputs). Finally, one may consider accounting for positiveness constraints into the GP framework using finite-dimensional Gaussian approximations \citep[see, e.g.,][]{Maatouk2016GPineqconst,LopezLopera2017FiniteGPlinear}.
	
\appendix
\section{Single-input single-output GP-Gene model}
\label{app:SISOfGPGene}
Let the reaction-diffusion model in \eqref{eq:PDE}. Next, we compute the covariance function for the driving-force $k_{u,u}$, and the cross-covariance function between the output and the driving-force $k_{y,u}$

\subsection*{Covariance function for the driving-force}
For the computation of the covariance function of the driving-force, we assume that $y$ is a zero-mean GP with covariance function given by Equation \eqref{eq:hGPmRNAKuu}. Then, the mRNA expression $u$ is also a zero-mean GP with covariance function $k_{u,u}(x,t,x',t') = \cov{u(x,t),u(x',t')}$ given by
\begin{align*}
k_{u,u}(x,t,x',t')
=& \mathbb{E} \bigg\{ \frac{1}{S} \bigg[ \partialdiff{y (x,t)}{t} + \lambda y(x,t) - D \partialdiff{^2 y (x,t)}{x^2} \bigg] \times \frac{1}{S} \bigg[ \partialdiff{y (x',t')}{t'} + \lambda y(x',t') - D \partialdiff{^2 y (x',t')}{{x'}^2} \bigg] \bigg\} \\
=& \frac{1}{S^2} [D^2 k^{xxx'x'}(x,t,x',t') - D k^{xxt'}(x,t,x',t')
- D k^{x'x't}(x,x',t,x',t') - D \lambda k^{xx}(x,t,x',t') \\
&- \lambda D k^{x'x'}(x,t,x',t') + k^{tt'}(x,t,x',t')
+ \lambda k^{t}(x,t,x',t') + \lambda k^{t'}(x,t,x',t') + \lambda^2 k(x,t,x',t')],	
\end{align*}
where $k^{x}$ is the derivative of $k$ w.r.t. the space variable $x$, and $k^{xt}$ is the derivative of $k^{x}$ w.r.t. the time variable $t$. The other derivatives follow the same structure. Due to the symmetry of the derivatives of the SE kernel, then we obtain
\begin{align*}
k_{u,u}(x,t,x',t')
=\frac{1}{S^2} \bigg[D^2 k^{xxxx}(x,t,x',t') - 2 D \lambda k^{xx}(x,t,x',t') 
- k^{tt}(x,t,x',t') + \lambda^2 k(x,t,x',t') \bigg],
\end{align*}
with the derivatives of $k(x,t,x',t')$ w.r.t. $x$ and $t$ given by
\begin{align*}
&k^{t}(x,x',t,t') = \sigma^2 k(x,x') k^{i}(t,t'), \\
&k^{tt}(x,x',t,t') = \sigma^2 k(x,x') k^{ii}(t,t'), \\	
&k^{xx}(x,x',t,t') = \sigma^2 k^{ii}(x,x') k(t,t'), \\
&k^{xxt}(x,x',t,t') = \sigma^2 k^{ii}(x,x') k^{i}(t,t'), \\
&k^{xxxx}(x,x',t,t') = \sigma^2 k^{iv}(x,x') k(t,t').
\end{align*}
The derivatives of the SE kernel function are given in \eqref{eq:hGPGapGeneSEdiff}.

\subsection*{Covariance function between the driving-force and the output}
The covariance function between the output $y$ and the force $u$, $k_{y,u} (x,t,x',t') = \cov{y(x,t),u(x',t')} $, is given by
\begin{align*}
k_{y,u}(x,t,x',t') 
&= \frac{1}{S} \mathds{E} \bigg\{ y(x,t) \bigg[ \partialdiff{y (x',t')}{t'} + \lambda y(x',t') - D \partialdiff{^2 y (x',t')}{{x'}^2} \bigg] \bigg\} \\
&= \frac{1}{S} \bigg[ \lambda k (x,t,x',t') - k^{t}(x,t,x',t') - D k^{xx}(x,t,x',t') \bigg].
\end{align*}

	\section*{Acknowledgements}	
	This work was funded by the project ``Probabilistic spatio-temporal models based on partial differential equations for the description of the regulatory dynamics for the Bicoid protein in the Drosophila Melanogaster body segmentation'' (by Colciencias, Colombia and  ECOS-NORD, France) with grant number C15M02. MAA has been partially financed by the EPSRC Research Projects EP/N014162/1 and EP/R034303/1.

\bibliographystyle{apa}  
\bibliography{arXiv2018_GPDrosM}  

\end{document}